\documentclass{article} 
\usepackage{iclr2025_conference,times}

\usepackage{amsmath,amsfonts,bm}

\def\eqref#1{equation~\ref{#1}}

\def\1{\bm{1}}

\DeclareMathAlphabet{\mathsfit}{\encodingdefault}{\sfdefault}{m}{sl}
\SetMathAlphabet{\mathsfit}{bold}{\encodingdefault}{\sfdefault}{bx}{n}

\usepackage{hyperref}
\usepackage{url}

\usepackage{bm}
\usepackage{amsmath}
\usepackage{multirow}
\usepackage{graphicx}
\usepackage{comment}

\def\zhiwu{\textcolor{black}}

\usepackage[normalem]{ulem}
\useunder{\uline}{\ul}{}

\newcommand{\methodname}{PoundNet }
\newcommand{\methodnameN}{PoundNet}

\title{Penny-Wise and Pound-Foolish in Deepfake Detection}

\author{
  Yabin Wang\textsuperscript{\rm 1, 2}, Zhiwu Huang\textsuperscript{\rm 2}$^\dagger$, Su Zhou\textsuperscript{\rm 1}, Adam Prugel-Bennett\textsuperscript{\rm 2}, Xiaopeng Hong\textsuperscript{\rm 3, 4}\\
  \textsuperscript{\rm 1}Xi'an Jiaotong University, P. R. China, \textsuperscript{\rm 2}University of Southampton, United Kingdom, \\
  \textsuperscript{\rm 3}Harbin Institute of Technology, P. R. China, \textsuperscript{\rm 4}Pengcheng Laboratory, P. R. China, \\
  \texttt{iamwangyabin@stu.xjtu.edu.cn, zhiwu.huang@soton.ac.uk} \\ \texttt{zhousu@xjtu.edu.cn},  \texttt{A.Prugel-Bennett@soton.ac.uk}, \\
  \texttt{hongxiaopeng@ieee.org}
  \thanks{$^\dagger$Corresponding author: Zhiwu Huang.}
}

\iclrfinalcopy 
\begin{document}

\maketitle

\begin{abstract}
The diffusion of deepfake technologies has sparked serious concerns about its potential misuse across various domains, prompting the urgent need for robust detection methods. Despite advancement, many current approaches prioritize short-term gains at expense of long-term effectiveness. 
This paper critiques the overly specialized approach of fine-tuning pre-trained models solely with a penny-wise objective on a single deepfake dataset, while disregarding the pound-wise balance for generalization and knowledge retention. To address this "Penny-Wise and Pound-Foolish" issue, we propose a novel learning framework (PoundNet) for generalization of deepfake detection on a pre-trained vision-language model. PoundNet incorporates a learnable prompt design and a balanced objective to preserve broad knowledge from upstream tasks (object classification) while enhancing generalization for downstream tasks (deepfake detection). 
We train PoundNet on a standard single deepfake dataset, following common practice in the literature. We then evaluate its performance across 10 public large-scale deepfake datasets with 5 main evaluation metrics—forming the largest benchmark test set for assessing the generalization ability of deepfake detection models, to our knowledge. The comprehensive benchmark evaluation demonstrates the proposed PoundNet is significantly less "Penny-Wise and Pound-Foolish", achieving a remarkable improvement of 19\% in deepfake detection performance compared to state-of-the-art methods, while maintaining a strong performance of 63\% on object classification tasks, where other deepfake detection models tend to be ineffective. Code and data are open-sourced at \href{https://github.com/iamwangyabin/PoundNet}{github}.
\end{abstract}

\section{Introduction}

The proliferation of deepfake technologies (e.g.,~\cite{rombach2022high, saharia2022photorealistic, ramesh2022hierarchical}) has raised significant concerns regarding its potential misuse in various domains, including politics, entertainment, and cybersecurity. 
Deepfakes, which are AI-generated synthetic media that either convey entirely new content or convincingly depict individuals saying or doing things they never did, pose a serious threat to the authenticity and trustworthiness of digital information and visuals. 
Consequently, there is an urgent need for robust deepfake detection methods to protect against potential malicious use cases.

\begin{figure}[]
\centering
\centerline{\includegraphics[width=0.95\linewidth]{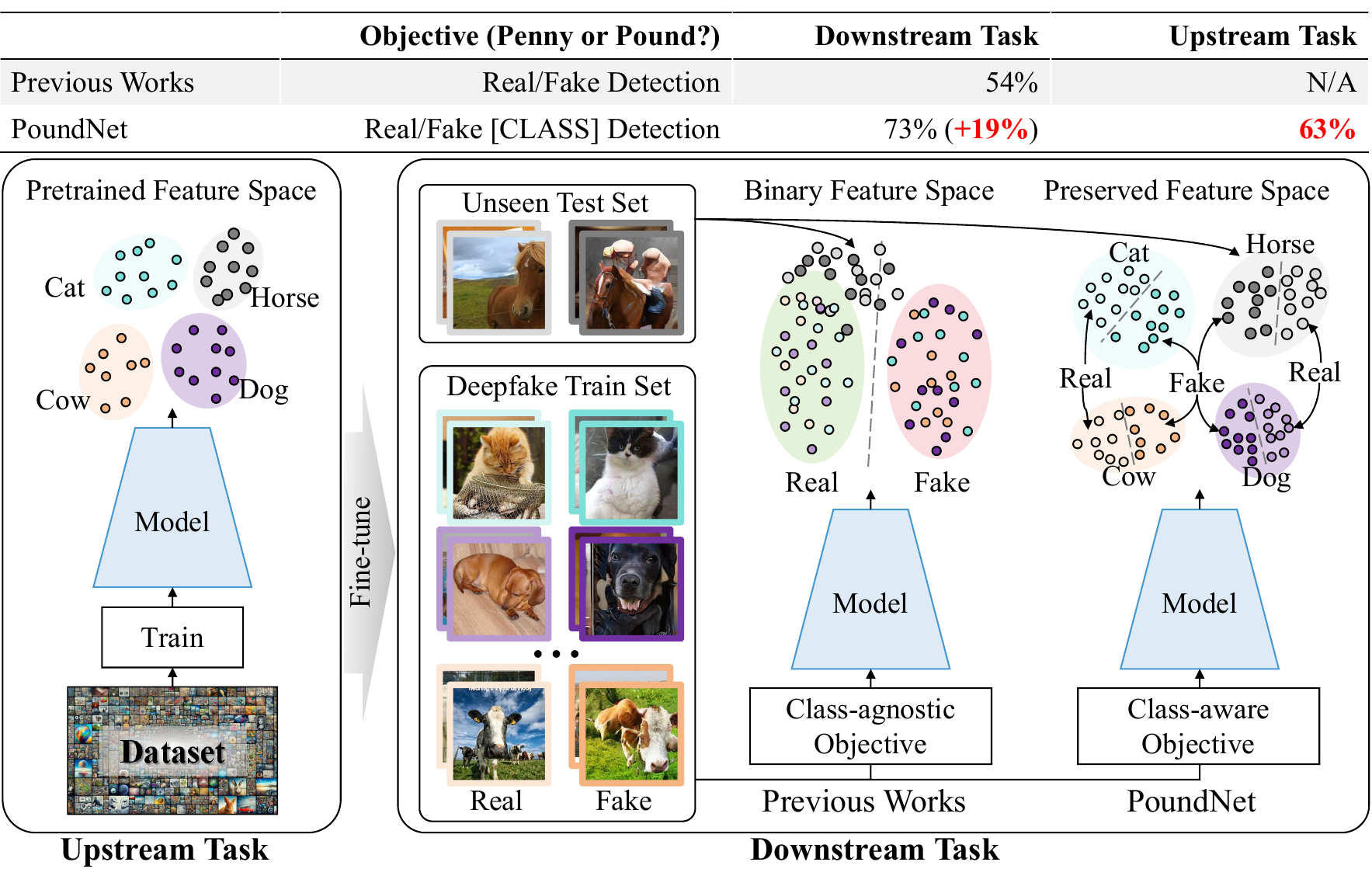}}
\caption{"Penny" and "Pound" in previous methods and proposed approach (PoundNet) for deepfake detection and beyond.  PoundNet improves deepfake detection (downstream) by 19\% across 10 datasets and maintains a strong 63\% performance in zero-shot object detection (upstream) across 5 datasets where many current detectors struggle. }
\label{fig:intro}
\vspace{-2em}
\end{figure}

In recent years, deepfake detection has made considerable progress, with numerous methods~\cite{wang2020cnn, chen2022self, ciftci2020fakecatcher} being proposed to identify and mitigate their impact. 
Typically, a deepfake detector functions as a binary classifier trained on specialized deepfake datasets.
Although a trained detector can perform well on known deepfakes, the challenge lies in generalizing to a wide range of unseen deepfake types~\cite{mirsky2021creation, wang2023dire, zhu2024genimage}.
To achieve generalization, as depicted in Fig.~\ref{fig:intro}, previous works like~\cite{wang2020cnn, ojha2023towards, tan2024frequencyaware} fine-tune pre-trained large-scale deep learning models, which excel at generalizing to new tasks, using a class-agnostic binary classification objective tailored to specific deepfake datasets. 
However, they frequently overlook the importance of balancing generalization ability with knowledge retention, which are two main essentials for effective and efficient deep learning~\cite{kirkpatrick2017overcoming, li2017learning}.
This strategy is akin to being "Penny-Wise and Pound-Foolish", where decisions made for short-term gains ultimately lead to larger losses or missed opportunities for long-term effectiveness. 
Fine-tuning large pre-trained models~\cite{radford2021learning} solely with a binary classification objective on downstream deepfake datasets is a "short-term gain" approach. 
While the trained detector can excel at identifying limited deepfake types encountered during training, it deteriorates the valuable, broad knowledge encoded in the pre-trained model. 
As a result, such compromise makes the detector hard to generalize to unseen deepfake types, as this underscores the principles of continual learning~\cite{li2017learning, Wang_Ma_Huang_Wang_Su_Hong_2023, Li_2023_WACV} to prevent catastrophic forgetting.

In light of the shortcomings associated with the greedy approach of fine-tuning pre-trained models on deepfake datasets, we propose a novel deepfake detection approach, \methodnameN, that prioritizes both generalization and knowledge preservation. 
Illustrated in Fig.~\ref{fig:intro}, our proposed approach differs from prior methods, which primarily focus on a class-agnostic binary training objective for deepfake detection. Instead, \methodname aims for class-aware deepfake detection, achieving notable improvements in generalization for downstream tasks (deepfake detection), with a substantial relative enhancement compared to state-of-the-art methods. Additionally, it demonstrates robust knowledge preservation, achieving a strong performance on upstream tasks (zero-shot object classification), a challenge for existing deepfake detection models due to their limited adaptability stemming from the greedy fine-tuning strategy. The efficacy of the class-aware deepfake detection objective in enhancing generalization is attributed to its simplification of the deepfake detection task by distinguishing between individual classes. 
Notably, large vision-language pre-trained models like CLIP~\cite{radford2021learning}, typically trained on extensive datasets (e.g., LAION-5B~\cite{schuhmann2022laion}), often encounter a broader range of classes during pre-training. Consequently, preserving class-based semantics during fine-tuning aids in generalizing to deepfake detection tasks. As shown in Fig.~\ref{fig:intro}, the proposed \methodname surpasses previous approaches by recognizing specific classes like the horse class. This focus reduces the primary task to distinguishing between real and fake instances within these classes, a more straightforward task compared to detecting fakes or reals without class information. Moreover, the class-preserved objective not only enhances deepfake detection but also maintains the integrity of class semantics, thereby ensuring promising performance on upstream tasks like object classification.

In our experimental evaluation, we assess the performance of state-of-the-art deepfake detection models as well as our proposed \methodnameN. 
The models are trained on a standard deepfake dataset, and their performance is evaluated on a wide range of unseen deepfakes. 
The evaluation is conducted on a large set of $10$ public deepfake datasets, containing over $50$ different deepfake types\footnote{Although some generative models overlap, variations in creation lead to diverse deepfakes, including different image classes and styles, which are highlighted in the experiments.}. 
Our results demonstrate the efficacy of our approach in achieving better generalization, showing around $19\%$ improvement averaged across 5 standard metrics, compared to existing methods. 
Furthermore, our approach preserves prior knowledge for zero-shot classification on general datasets, achieving an accuracy of $63\%$ on the ImageNet series benchmarks. 
In contrast, most state-of-the-art deepfake detection models are unable to perform upstream tasks. 
Overall, our proposed approach offers a more robust and reliable solution to the challenge of deepfake detection by balancing generalization capabilities and knowledge preservation. 
In summary, this paper’s main contributions include:

\begin{itemize}
    \item Identifying the Penny-Wise and Pound-Foolish phenomenon within many current deepfake detection models. 

    \item Proposing an Anti-Pennywise and Pound-Foolish learning framework (PoundNet) to improve generalization for detection (downstream) and classification (upstream) tasks in the context of fine-tuning pre-trained deep learning models. 

    \item Evaluating the proposed PoundNet and 10 state-of-the-art deepfake detection methods using $10$ large test datasets and 5 standard metrics, plus 5 datasets for object classification. To our knowledge, this benchmark is the largest for assessing deepfake detection generalization, as existing studies usually use one or two of these datasets.
\end{itemize}

\section{Related Work}




\noindent\textbf{Deepfake Detection and Approaches for Generalization.} 
In recent years, many deepfake detectors~\cite{wang2020cnn, afchar2018mesonet, nguyen2019capsule, li2020face, dang2020detection, ni2022core, cao2022end} have been proposed for ensuring the authenticity of visual content.
CNNDet~\cite{wang2020cnn} uses a plain binary cross-entropy loss to fine-tune a pre-trained CNN model, ResNet50~\cite{he2016deep}, for deepfake detection. 
The training dataset, ForenSynths, consists of real images from $20$ categories in the LSUN dataset~\cite{yu2015lsun} with synthetic images generated by ProGAN~\cite{karras2017progressive}.
They also propose using data augmentation, random JPEG compression, and Gaussian blurring to improve the model's cross-generator performance and robustness.
Patchfor~\cite{chai2020makes} utilizes truncation at different layers of neural networks (e.g., ResNet~\cite{he2016deep}, Xception~\cite{chollet2017xception}) to obtain predictions based on the collective information from all patches. 
Recently, advancements in large pre-trained models, such as ViT~\cite{dosovitskiy2020image} and CLIP~\cite{radford2021learning}, have demonstrated powerful zero-shot generalization across various computer vision tasks.
UnivFD~\cite{ojha2023towards} utilizes the image features extracted by the pre-trained CLIP vision encoder to train a linear classifier for binary classification. 
Since CLIP already has extensive prior knowledge about the world, even training a detector exclusively on its frozen vision features can achieve good generalization.
SPrompts~\cite{wang2022s} and CLIPping~\cite{khan2024clipping} employ prompt tuning strategies to fine-tune the CLIP model specifically for deepfake detection.
When deepfake detection is treated as a downstream task for pre-trained models, previous methods that fine-tune these models with a binary classification objective on deepfake datasets adopt a "short-term gain" strategy. While effective for specific deepfake types encountered during training, this approach compromises the general knowledge in pre-trained models, hindering their ability to generalize to new deepfakes and perform upstream tasks like object detection.


As synthetic images often lack high-frequency details compared to real images, ~\cite{frank2020leveraging, tan2024frequencyaware, tan2023rethinking, liu2020global} learn to extract frequency information from images to achieve generalizability. 
LGrad~\cite{tan2023learning} utilizes a pre-trained CNN model to convert images to gradients, which can extract generalizable features in GAN-generated images. 
Fusing~\cite{ju2022fusing} is a two-branch model that integrates global and local features of AI-synthesized images.
Other methods~\cite{liu2022detecting, wang2023dire} use the predicted noise or reconstruction error of the image to learn consistent generalizable features. However, these methods are computationally expensive for inference, as they require a denoising model to preprocess test samples. 
All these methods can be regarded as using different feature extractors to obtain better generalizable features, but they still use the greedy binary classification objective for training. 
Moreover, these methods heavily rely on low-level, imperceptible features within the image, which may not be robust in real-world scenarios. 
Recent works~\cite {grommelt2024fake, zhu2024genimage, wang2020cnn} have found that the effectiveness of low-level features can be easily compromised by common image processing techniques, such as JPEG compression.



Some recent studies~\cite{DBLP:conf/visigrapp/Khan24, Corvi_2023_CVPR, zhu2024genimage} show that existing deepfake detectors trained solely on GAN-generated images do not effectively generalize to identify synthetic images produced by diffusion models. 
One reason is that the frequency spectrum of diffusion-generated images differs from that of GAN-generated images. This issue, along with the emergence of text-to-image generative models, underscores the need for more robust and adaptable detection methods.
As the task becomes increasingly complex with the advent of more generative models, it is crucial to fully evaluate detection methods. To address this, several benchmarks~\cite{zhu2024genimage, ojha2023towards, wang2023dire, chang2023antifakeprompt, Sinitsa_2024_WACV} that contain mixed deepfake types for evaluation have been proposed.
HiFi~\cite{guo2023hierarchical} utilizes a multi-level classification system to enhance the detection and localization of various image forgeries, from GANs to Diffusions.
Some methods~\cite{Li_2023_WACV, wang2022s} have found that continual learning is an effective way to enhance the generalization ability for detecting unseen deepfakes by learning more deepfake types.
However, none of the previous works identify the underlying issue, the "Penny-Wise and Pound-Foolish" phenomenon, that prevents deepfake detectors from generalizing well to new, unseen data.

\noindent\textbf{Prompt Tuning for Adapting Pre-trained Models.} 
Large vision-language models (VLMs), such as CLIP~\cite{radford2021learning}, have exhibited remarkable zero-shot performance across a wide range of computer vision tasks. 
Prompt tuning~\cite{liu2023gpt, li-liang-2021-prefix, lester-etal-2021-power} is a popular technique used to transfer pre-trained VLMs to specific downstream tasks without modifying the entire model.
Instead of using hand-crafted prompts, 
CoOp~\cite{zhou2022learning} and CoCoOp~\cite{zhou2022conditional} optimize a set of continuous learnable vectors for downstream tasks and concatenated with the input text prompts to adapt the VLMs to the specific task. 
VPT~\cite{jia2022visual} extends the concept of prompt tuning to the visual domain, which uses learnable vectors as part of input images for vision encoder.
MaPLe~\cite{khattak2023maple} enhances the performance and generalization capabilities of VLMs by leveraging multi-modal prompts of both textual and visual components.
Recent works~\cite{ren2024prompt, Khattak_2023_ICCV} explore effective learning prompts from base datasets to capture rich, transferable knowledge across various domains.
Similarly, CLIPping~\cite{khan2024clipping} and SPrompts~\cite{wang2022s} find that prompt tuning is an effective way to enhance generalization for detection.
However, none of the existing study the connection between generalization and forgetting when tuning pre-trained models for deepfake detection.
In contrast, we propose a prompt-tuning strategy to train a robust deepfake detector by preserving learned knowledge to enhance generalization capabilities in deepfake detection.

\section{Proposed Approach}

\begin{figure}[]
\centering
\centerline{\includegraphics[width=0.9\linewidth]{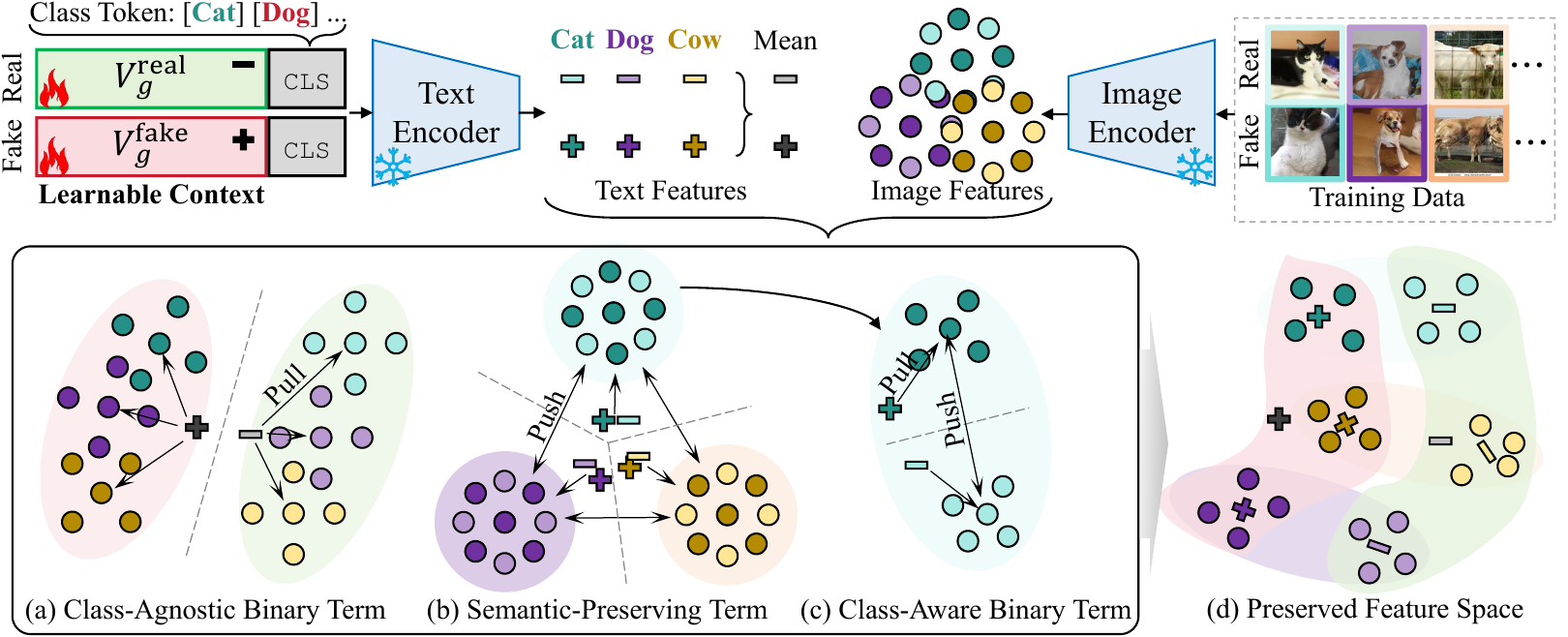}}
\vspace{-0.5em}
\caption{Proposed framework (PoundNet) on a pre-trained vision-language model with a learnable prompt pair (Top Left) and a balanced object (Bottom Left) with three basic loss components: (a) class-agnostic binary, (b) semantic-preserving, and (c) class-aware binary terms. } 
\label{fig:framework}
\vspace{-1em}
\end{figure}

The proposed learning approach, \methodnameN, aims to achieve class-aware deepfake detection while balancing generalization and knowledge preservation. 
\methodname builds upon the pre-trained Contrastive Language-Image Pre-Training (CLIP)~\cite{radford2021learning} and fine-tunes it using our designed prompt pair and suggested balanced objective. 
CLIP leverages contrastive learning to create a joint feature space for images and text, where corresponding image and text caption pairs are closer together, and non-corresponding pairs are pushed farther apart.
For zero-shot object classification, the typical text prompt format is ``\texttt{a photo of a [CLASS]}'', where \texttt{[CLASS]} should be replaced with the specific object or category name for classification. 
Ideally, we can use the prompt format of ``\texttt{a [real/fake] photo of a [CLASS]}'' to guide the CLIP model in performing downstream deepfake detection. 
However, it is challenging for pre-trained CLIP to understand the abstract concept of deepfakes in natural language. 
As shown in Fig. \ref{fig:framework} (Top Left), to better parameterize the ``\texttt{a [real/fake] photo}'' context for the deepfake detection scenario, we introduced learnable paired prompts for reals and fakes respectively. The prompts are fed into the text encoder, which produces corresponding text feature representations. 
On the other end, a training set of real and fake images is processed through the image encoder to generate associated image features. 
To align the text features and image features for deepfake detection and form a preserved feature space (Fig.\ref{fig:framework} (d)), we propose a balanced objective comprising three components: Class-Agnostic Binary term (Fig.\ref{fig:framework} (a)), which is a high-level, abstract concept, focuses on distinguishing real and fake instances without considering detailed semantic context. It represents a general binary classification of real versus fake.
Semantic-Preserving term (Fig.\ref{fig:framework} (b)) ensures the retention of broader knowledge encoded within the pre-trained model, which is often neglected by existing approaches that excessively fine-tune using only class-agnostic binary term. 
Class-Aware Binary term (Fig.\ref{fig:framework} (c)) aims at discerning between individual classes (e.g., detecting fake images of cats within a set of cat images), making the deepfake detection process more streamlined and effective.

\textbf{Prompt Design.}
The CLIP~\cite{radford2021learning} model consists of a text encoder $\mathcal{G}$ and an image encoder $\mathcal{V}$. 
It is widely used to perform zero-shot object classification by measuring the similarity between visual features and class-specific text features. 
Generally, the class-specific text features are computed by the text encoder accepting hand-crafted prompts with class names (e.g., ``\texttt{a photo of a [CLASS]}'').
The probability for the class $c$  is given by
\begin{equation}
    \label{eq:zs}
    p(c|x)=\frac{\exp(<T_c, I>)}{\sum_j^K\exp(<T_{c_j}, I>)}, 
\end{equation}
where $I = \mathcal{V}(x)$ is the feature of the input image $x$, $T_i=\mathcal{G}(t_i)$ is the class-specific features of the textual input $t_i$ for the class $c_i \in \mathcal{C}$, $K$ is the class number, and $<\cdot,\cdot>$ denotes the cosine similarity.

Recent methods~\cite{khattak2023maple, wang2022s, jia2022visual} use prompt tuning as an effective strategy for fine-tuning CLIP on downstream tasks to better preserve prior knowledge.
For the language side, \cite{khattak2023maple} suggests optimizing a set of $M$ learnable continuous context vectors $\bm{V_\mathcal{G}}=[\bm{v}^1_\mathcal{G},\bm{v}^2_\mathcal{G},...,\bm{v}^M_\mathcal{G}]$ for downstream tasks. 
The input prompts for the text encoder follow the form $\bm{t}_c=[\bm{V}_\mathcal{G}; \bm{c}]$, where $\bm{c}$ represents the class token embedding corresponding to the class $c$. 
For deepfake detection, SPrompt~\cite{wang2022s} and CLIPping~\cite{khan2024clipping} propose appending the class names "real" and "fake" to a set of learnable prompts $\bm{V_\mathcal{G}}$ and then doing optimization with a binary cross-entropy loss. 
Their prompts resemble hand-crafted versions like ``\texttt{a photo of [real/fake]}''. 
However, similar to other prompting approaches, they cannot fully leverage the pre-trained knowledge of CLIP, since the concepts of "real" and "deepfake" are abstract and distant from the semantic concepts of general objects.

Instead of directly using "real" and "fake" as class names, we maintain the term\texttt{[CLASS]} in the prompts to represent general semantic classes and introduce a pair of learnable contexts, $\bm{V}^{\text{fake}}_\mathcal{G}$ and $\bm{V}^{\text{real}}_\mathcal{G}$, to capture the abstract real and fake concepts, respectively.
This design is analogous to using the phrases  ``\texttt{a [fake/real] photo of a [CLASS].}'' in natural language, but instead of using hand-crafted textual descriptions, we optimize these context vectors through gradient backpropagation. 
As illustrated in Fig. \ref{fig:framework} (Top Left), by concatenating the class tokens $\bm{c}$ with the context vectors ($\bm{V}^{\text{fake}}_\mathcal{G}$ and $\bm{V}^{\text{real}}_\mathcal{G}$), and feeding these concatenated prompts into the text encoder, we can generate class-specific text features ($T_c^{\text{fake}}$ and $T_c^{\text{real}}$) for both the fake and real versions of a given class $c$.

As discovered by~\cite{khattak2023maple, wang2022s}, it's crucial to prompt both the vision and language branches simultaneously to enhance the efficacy of adaptation to downstream tasks. Thus, we follow them to additionally introduce a set of $M$ learnable tokens as visual prompts $\bm{V_\mathcal{V}} = [\bm{v}^1_\mathcal{V},\bm{v}^2_\mathcal{V},...,\bm{v}^M_\mathcal{V}]$ into the vision branch, similar to language prompts but combined with the image tokens. 
Both learnable text and vision prompts can be inserted into each transformer block of the encoder up to a depth $J$, allowing for more tuning parameters. 
During fine-tuning, we merely optimize prompts while keeping other network components frozen, which reduces the risk of overfitting and allows the model to adapt efficiently to new tasks without extensive retraining.

\textbf{Balanced Objective.}
The training dataset for deepfake detection can be denoted as $\mathcal{D} = \{(x_i, y_i, c_i)\}_{i=1}^{N}$, where $x_i \in \mathcal{X}$ represents an image, $y_i \in \{0, 1\}$ is the binary label ($0$ for real, $1$ for fake), and $c_i \in \mathcal{C}$ denotes a semantic category label (e.g., "dog", "cat") for image $x_i$.
Semantic category labels are typically predefined in widely used deepfake datasets~\cite{wang2020cnn, zhu2024genimage} because they are created using generative models trained on datasets like ImageNet~\cite{deng2009imagenet}, LSUN~\cite{yu2015lsun}, and CelebA~\cite{liu2015deep}.


The proposed balanced objective consists of three components: Class-Agnostic Binary term (Fig.\ref{fig:framework} (a)),
Semantic-Preserving term (Fig.\ref{fig:framework} (b)), and
Class-Aware Binary term (Fig.\ref{fig:framework} (c)). The Class-Agnostic Binary term is based on commonly-used training objective of many previous works. It is the binary cross-entropy (BCE) loss function, as defined in Eq.~\ref{eq:bce}. 
This loss function is class-agnostic as it only considers the binary labels $y_i$, disregarding the semantic category labels $\mathcal{C}$.
\begin{equation}
\mathcal{L}_{bce} = - \frac{1}{N} \sum_{i=1}^{N} \left[ y_i \log(\hat{y}_i) + (1 - y_i) \log(1 - \hat{y}_i) \right], 
\label{eq:bce}
\end{equation}
where $\hat{y}_i$ is the predicted probability that the image $x_i$ is fake, and $N$ is the sample number.

To adapt $\mathcal{L}_{bce}$ to CLIP, which requires class tokens appended to the learnable contexts to make predictions as Eq.~\ref{eq:zs}, we propose an approach to achieve class-agnostic binary classification with semantic awareness.
As shown in Fig.~\ref{fig:framework}, we gather a set of class names $\mathcal{C}$ (e.g., "dog", "cat", "cow", etc.) that adequately represent the diversity of objects or scenes the model might encounter. 
Then, we concatenate the class tokens with the real and fake learnable contexts to obtain the corresponding text embeddings for the real and fake features of each class. 
We take the mean of all class embeddings to represent the general concept of real and fake, as $T^{\text{real}} = \frac{1}{C} \sum_{j}^{C} T_{j}^{\text{real}}$ and $T^{\text{fake}} = \frac{1}{C} \sum_{j}^{C} T_{j}^{\text{fake}}$. 
We can modify binary cross-entropy loss (Eq.~\ref{eq:bce}) by replaceing $\hat{y}_i$ with $p(T^{\text{fake}} | x_i)$ as defined in Eq.~\ref{eq:p_fake}.
\begin{equation}
p(T^{\text{fake}} | x_i) = \frac{\exp(\langle T^{\text{fake}}, I_i \rangle)}{\exp(\langle T^{\text{real}}, I_i \rangle) + \exp(\langle T^{\text{fake}}, I_i \rangle)}.
\label{eq:p_fake}
\end{equation}

However, solely optimizing the binary cross-entropy loss $\mathcal{L}_{bce}$ for the downstream deepfake detection task can lead to a myopic focus on short-term performance gains. This approach fails to strike a balance between the downstream task and the valuable knowledge acquired from upstream pre-training tasks, resulting in a penny-wise and pound-foolish phenomenon.
Thus we further introduce a regularization loss called Semantic-Preserving Term $\mathcal{L}_{spm}$, which optimizes both context vectors, $\bm{V}^{\text{fake}}_\mathcal{G}$ and $\bm{V}^{\text{real}}_\mathcal{G}$, to perform general object classification, as shown in Eq.~\ref{eq:spml}.
\begin{equation}
    \mathcal{L}_{spm} = - \frac{1}{N} \sum_{i}^N c_i \left( \log \left( \frac{\exp(\langle T_{c_i}^{\text{fake}}, I_i \rangle)}{\sum_j^K \exp(\langle T_{c_j}^{\text{fake}}, I_i \rangle)} \right) + \log \left( \frac{\exp(\langle T_{c_i}^{\text{real}}, I_i \rangle)}{\sum_{j}^K \exp(\langle T_{c_j}^{\text{real}}, I_i \rangle)} \right) \right),
    \label{eq:spml}
\end{equation}
\zhiwu{where $T_{c_i}^{\text{X}}$ is generated by the text encoder $\mathcal{G}$ with the input being $t_{c_i}=[\bm{V}_\mathcal{G}^{\text{X}}; \bm{c}_i]$, $X$ is either for "real" or "fake", and $K$ is the total class number.} $\mathcal{L}_{spm}$ only takes into account of the semantic labels $\mathcal{C}$ and ignores the binary label $y_i$. 
It ensures that the fine-tuning process maintains the semantic relationships and generalizable knowledge inherent in the original CLIP model.
By doing so, we can leverage the rich contextual understanding of CLIP to enhance deepfake detection without compromising its ability to recognize and understand a wide range of objects and scenes.

In addition, we propose Class-Aware Binary Term, $\mathcal{L}_{cab}$, to smoothly transition CLIP from general object classification tasks to deepfake detection.
$\mathcal{L}_{cab}$ is a binary cross-entropy loss applied within each category space, ensuring that the model can differentiate between real and fake images under the conditions of each class.
$\mathcal{L}_{cab}$ can be formulated as
\begin{equation}
\mathcal{L}_{cab} = - \frac{1}{N} \sum_{i}^{N} \sum_{c}^{\mathcal{C}} \left( y_{i} \log p(c, y_i|x_i) + (1 - y_{i}) \log (1 - p(c, y_i|x_i)) \right),
\label{eq:cab}
\end{equation}
where $p(c, y_i | x_i)$ is the probability that the sample $x_i$ belongs to class $c$ and has label $y_i$. $\mathcal{L}_{cab}$ ensures that the model learns to differentiate between real and fake images within the context of each semantic class, or in other words, in a class-wise manner. 
It preserves the original semantic space of CLIP, making only minimal adjustments necessary for deepfake detection.

During training, we optimize the \methodname in an end-to-end manner with the balanced objective
\begin{equation}
\mathcal{L} = {\mathcal{L}}_{bce} + \lambda_{1} \mathcal{L}_{spm} + \lambda_{2} \mathcal{L}_{cab},
\label{eq:total_loss}
\end{equation}
where $\lambda_{1}$ and $\lambda_{2}$ are hyperparameters that balance the impact of each term. 
We simply set both $\lambda_{1}$ and $\lambda_{2}$ to $1$ in our main experiments. 
Their impact is explored in the ablation study. 

For deepfake detection inference on any given images, PoundNet concatenates them with vision prompts $\bm{V_\mathcal{V}}$ and computes the textual prompt-based embeddings $T^{\text{real}}$ and $T^{\text{fake}}$ through CLIP. 
Then, it uses Eq. \ref{eq:p_fake} to make deepfake predication. 
If the semantic class of the given samples is known, it is advantageous to use the specific class embeddings, denoted as $T_{c_i}^{\text{X}}$. 
For object classification inference, we can either use $\bm{V}^{\text{fake}}_\mathcal{G}$ or $\bm{V}^{\text{real}}_\mathcal{G}$ with preset class names as class-specific features and make predictions using Eq.~\ref{eq:zs}.




\section{Experiments}


\textbf{Benchmark Datasets.}
For general AI-generated image detection, we select $7$ large public datasets for testing include ForenSynths~\cite{wang2020cnn}, GenImage~\cite{zhu2024genimage}, GANGen-Detection~\cite{chuangchuangtan-GANGen-Detection}, DiffusionForensics~\cite{wang2023dire}, Ojha~\cite{ojha2023towards}, AntiFake~\cite{chang2023antifakeprompt} and DIF~\cite{Sinitsa_2024_WACV}.
For classical deepfake detection, we select $3$ large public datasets UADFV~\cite{yang2019exposing}, Celeb-DF v1 and V2~\cite{Celeb_DF_cvpr20}. 
To evaluate the zero-shot performance on general classification tasks, we employ the widely-used ImageNet dataset~\cite{deng2009imagenet} and four of its variants: ImageNet-V2~\cite{recht2019imagenet}, ImageNet-S~\cite{wang2019learning}, ImageNet-A~\cite{hendrycks2021natural}, and ImageNet-R~\cite{hendrycks2021many}. 
The detailed statistics of the mainly used datasets is provided in Table~\ref{tab:10datasets_stas} at the appendix.

\textbf{State-of-the-art Methods.} \zhiwu{We compare the proposed PoundNet against 10 state-of-the-art deepfake detection methods, most of which were published after 2022.} 

\textbf{Performance Metrics.} 
To thoroughly evaluate the performance of PoundNet and 10 state-of-the-art deepfake detection methods, we employ a range of widely used metrics for detection and classification, includingAverage Precision ($\mathrm{AP}$) \cite{wang2020cnn}, Accuracy ($\mathrm{ACC}$), F1 score ($\mathrm{F1}$). 
As the mainstream metrics primarily focus on ranking-based performance, we suggest using another metric called the Area Under the F1-threshold Curve ($\mathrm{AUC_{f1}}$), which captures the F1 score's variation across different decision thresholds, ranging from $0$ to $1$, as $\mathrm{AUC_{f1}} = \int_0^{1} \mathrm{F1}(\tau) \, d\tau$,
where $\tau$ is the threshold that determines the boundary between positive and negative predictions. 
Similarly, we also use the Area Under the F2-threshold Curve ($\mathrm{AUC_{f2}}$), which emphasizes recall over precision. \zhiwu{Appendix \ref{sec:benchmark_details} discusses the benefits of $\mathrm{AUC_{f1}}$ and $\mathrm{AUC_{f2}}$ over AP and others.}
Note that AntiFake~\cite{chang2023antifakeprompt} provides separate parts for real images and various deepfake types, making Accuracy the sole evaluation metric.

\textbf{Training details.} 
\zhiwu{Following ~\cite{ojha2023towards, wang2022s, khan2024clipping}, we use the same pre-trained CLIP model~\cite{radford2021learning}, ViT-L/14, for all evaluated CLIP-based models like PoundNet, SPrompts~\cite{wang2022s}, and UnivFD~\cite{ojha2023towards}}.
For PoundNet, 
we insert a pair of learnable contexts, each with a length of $16$, into the text encoder for real and fake, respectively. 
To enhance tuning for downstream tasks, we also add prompts to the vision encoder, with a length of $16$.
The model is trained for $1$ epoch using the Adam optimizer~\cite{DBLP:journals/corr/KingmaB14}, with a weight decay of $10^{-4}$ and a batch size of $128$.
\zhiwu{We set the hyperparameters of $\lambda_1$ and $\lambda_2$ to $1$ in main experiments}.
\zhiwu{Following ~\cite{wang2020cnn,wang2022s}, we train PoundNet and all other methods on the training set of ForenSynths~\cite{wang2020cnn}, where deepfakes are generated by ProGAN \cite{karras2017progressive}.}
We adopt the same data augmentment with previous works~\cite{ojha2023towards, wang2020cnn} by applying Gaussian blurring and JPEG compression with a probability of $10\%$ to the images during training.
All images are resized to $256\times256$ and then randomly cropped to $224\times224$.
We use a single NVIDIA A100 GPU for each training.

\textbf{Evaluation protocols.} 
For the $7$ general deepfake detection test datasets, we use the same process to evaluate all methods. 
We first resize the shorter side of images to $256$ pixels and then central crop to $224 \times 224$ pixels. 
As noted in previous works \cite{grommelt2024fake, zhu2024genimage}, there are some biases exist in the benchmark data. 
For example, all real images are saved in JPEG format, whereas synthetic images are in PNG format. 
In the main paper, we apply JPEG compression with $90\%$ quality~\cite{wang2020cnn} to all input images to mitigate such biases and better simulate real-world scenarios. 
More results with different compression levels are provided in Appendix~\ref{sec:jpeg_compression}. 
For the $3$ classical deepfake detection benchmarks, we adhere to the standard process outlined by DeepfakeBench~\cite{DeepfakeBench_YAN_NEURIPS2023}. 
We will make all our implementations open source to facilitate the replication of our evaluations.

\textbf{Comparisons to the state-of-the-art.} We present the results of a comprehensive evaluation of the proposed PoundNet, comparing it to 10 state-of-the-art methods for deepfake detection. 
The overall evaluation results for deepfake detection and zero-shot object classification are presented in Table~\ref{tab:10benchmarks_sum} and Table~\ref{tab:5benchmarks_sum1}.
The detailed evaluation results on $10$ deepfake test datasets, are presented in Table~\ref{tab:10benchmarks}. Except for the results related to the AP metric, which reflect performance within a much narrower and more clustered range compared to $\mathrm{AUC_{f1}}$ and $\mathrm{AUC_{f2}}$ as studied in Appendix~\ref{sec:benchmark_details}, Table~\ref{tab:10benchmarks_sum} and Table~\ref{tab:10benchmarks} shows that PoundNet exhibits exceptional performance improvement over the other methods across the rest evaluation metrics and all the 10 used deepfake datasets. Notably, our method achieves excellent results on traditional deepfake detection benchmarks (UADFV~\cite{yang2019exposing}, Celeb-DF v1 and V2~\cite{Celeb_DF_cvpr20}) without requiring any training on human faces deepfake datasets. This highlights its impressive generalization capabilities in the challenging domain of deepfake detection that requests strong generalization. Furthermore, Table~\ref{tab:5benchmarks_sum1} demonstrates that the proposed PoundNet still performs well for the zero-shot object classification tasks. In contrast, all the 10 evaluated state-of-the-art deepfake detection methods fail to handle this upstream task due to their used objectives, which limit  their generalization capabilities to the detection tasks alone.


\begin{table}[]
\centering
\caption{Average deepfake detection performance of the proposed \methodname and $10$ state-of-the-art methods on $10$ public deepfake detection benchmarks using 5 main evaluation metrics ($\mathrm{AP}$, $\mathrm{F1}$, $\mathrm{ACC}$, $\mathrm{AUC_{f1}}$, and $\mathrm{AUC_{f2}}$). Table \ref{tab:10benchmarks} presents the results for the first four evaluation metrics across each dataset, while tables in Appendix~\ref{sec:mainresults_20bench} report more detailed performance with more metrics and additional curves. \textbf{Bold} and {\ul Underlined} highlight the best and the second-best performance, respectively. \zhiwu{\emph{Note that AP merely reflects performance in a much smaller and clustered range compared to $\mathrm{AUC_{f1}}$ and $\mathrm{AUC_{f2}}$ (Appendix~\ref{sec:benchmark_details}).}}
}
\label{tab:10benchmarks_sum}
\resizebox{0.7\textheight}{!}{
\begin{tabular}{l|rrrrrrrrrr|r}
\hline
 & \multicolumn{1}{l}{CNNDet} & \multicolumn{1}{l}{FreDect} & \multicolumn{1}{l}{GramNet} & \multicolumn{1}{l}{Fusing} & \multicolumn{1}{l}{LNP} & \multicolumn{1}{l}{SPrompts} & \multicolumn{1}{l}{UnivFD} & \multicolumn{1}{l}{LGrad} & \multicolumn{1}{l}{NPR} & \multicolumn{1}{l|}{Freqnet} & \multicolumn{1}{r}{\methodname} \\ \hline
$\mathrm{AP}$ & 72.53 & 69.23 & 60.69 & 75.05 & 62.42 & 76.36 & {\ul 86.09} & 56.82 & 54.41 & 56.22 & \textbf{86.14} (\textcolor{red}{\bm{$+0.05 $}}) \\
$\mathrm{F1}$ & 17.10 & {\ul 47.50} & 9.87 & 16.87 & 10.23 & 16.26 & 42.56 & 1.38 & 14.63 & 13.36 & \textbf{70.58} (\textcolor{red}{\bm{$+23.08$}}) \\
$\mathrm{ACC}$ & 47.01 & 54.29 & 44.06 & 47.43 & 43.76 & 46.51 & {\ul 58.90} & 41.85 & 43.37 & 42.67 & \textbf{72.00} (\textcolor{red}{\bm{$+13.10$}}) \\
$\mathrm{AUC_{f1}}$ & 17.50 & {\ul 43.61} & 10.16 & 17.65 & 10.88 & 16.90 & 43.10 & 1.89 & 14.88 & 13.73 & \textbf{67.61} (\textcolor{red}{\bm{$+24.00$}}) \\ 
$\mathrm{AUC_{f2}}$ & 14.38 & {\ul 42.28} & 9.64 & 14.73 & 8.17 & 13.55 & 37.69 & 1.55 & 13.94 & 12.55 & \textbf{66.62} (\textcolor{red}{\bm{$+24.34$}}) \\
\hline
Average & 33.70 & 51.38	&26.88	&34.35	&27.09	&33.92	& {\ul 53.67}	&20.70& 28.25	&27.71	& \textbf{72.59} (\textcolor{red}{\bm{$+18.92$}})  \\
\hline
\end{tabular}
}

\end{table}

\begin{table}[]
\label{tab:5benchmarks_sum1}
\centering
\caption{Zero-shot object classification performance of the proposed \methodnameN. CLIP uses the official prompt templates specifically designed for testing ImageNet. $\mathrm{\methodname_{real}}$/$\mathrm{\methodname_{fake}}$: \methodname leverages context vectors (textual prompt) representing real/fake classes respectively, learned from the deepfake dataset, to perform inference.
\emph{Note that NONE of the 10 state-of-the-art deepfake detection methods (`OTHERS') in Table \ref{tab:10benchmarks_sum} are capable of handling this upstream task.} }
\resizebox{0.6\textheight}{!}{
\begin{tabular}{lrrrrrl}
\hline
\multicolumn{1}{l|}{Method} & ImageNet-A & ImageNet-V2 & ImageNet-R & ImageNet-S & \multicolumn{1}{r|}{ImageNet} & Average \\ \hline
\multicolumn{1}{l|}{$\mathrm{CLIP}$} & 70.92 & 69.77 & 86.91 & 58.51 & \multicolumn{1}{r|}{75.16} & \multicolumn{1}{r}{72.25} \\ \hline
\multicolumn{1}{l|}{OTHERS} & N/A & N/A & N/A & N/A & \multicolumn{1}{r|}{N/A} & \multicolumn{1}{r}{N/A} \\
\multicolumn{1}{l|}{$\mathrm{\methodname_{fake}}$} & 57.67 & 57.92 & 79.63 & 48.32 & \multicolumn{1}{r|}{63.96} & \multicolumn{1}{r}{61.50} \\
\multicolumn{1}{l|}{$\mathrm{\methodname_{real}}$} & 60.40 & 60.21 & 79.62 & 49.11 & \multicolumn{1}{r|}{66.01} & \multicolumn{1}{r}{63.07} \\ 
\hline
\end{tabular}
}
\end{table}

\begin{table}[]
\caption{Deepfake detection performance of the proposed PoundNet and $10$ state-of-the-art methods on $10$ public deepfake detection benchmarks in terms of $\mathrm{AP}$, $\mathrm{F1}$, $\mathrm{ACC}$, and $\mathrm{AUC_{f1}}$. \textbf{Bold} and {\ul Underlined} highlight the best and the second-best performance, respectively.}
\label{tab:10benchmarks}
\centering
\resizebox{0.6\textheight}{!}{
\begin{tabular}{rrrrrrrrrrrrr}
\hline
\multicolumn{1}{c|}{\multirow{2}{*}{Methods}} & \multicolumn{4}{c|}{ForenSynths} & \multicolumn{4}{c|}{GenImage} & \multicolumn{4}{c}{GANGen-Detection} \\ \cline{2-13} 
\multicolumn{1}{c|}{} & $\mathrm{AP}$ & $\mathrm{F1}$ & $\mathrm{ACC}$ & \multicolumn{1}{c|}{$\mathrm{AUC_{f1}}$} & $\mathrm{AP}$ & $\mathrm{F1}$ & $\mathrm{ACC}$ & \multicolumn{1}{c|}{$\mathrm{AUC_{f1}}$} & $\mathrm{AP}$ & $\mathrm{F1}$ & $\mathrm{ACC}$ & $\mathrm{AUC_{f1}}$ \\ \hline
\multicolumn{1}{l|}{CNNDet} & 89.05 & 54.04 & 70.18 & \multicolumn{1}{r|}{53.59} & 64.95 & 9.58 & 52.04 & \multicolumn{1}{r|}{10.38} & 82.69 & 37.27 & 62.15 & 37.57 \\
\multicolumn{1}{l|}{FreDect} & 67.48 & 55.64 & 62.78 & \multicolumn{1}{r|}{49.63} & {\ul 78.49} & \textbf{44.77} & {\ul 63.60} & \multicolumn{1}{r|}{{\ul 42.92}} & 63.59 & 21.33 & 54.81 & 24.74 \\
\multicolumn{1}{l|}{GramNet} & 57.52 & 14.30 & 50.10 & \multicolumn{1}{r|}{14.57} & 63.01 & 33.06 & 59.09 & \multicolumn{1}{r|}{33.10} & 50.00 & 0.41 & 49.99 & 0.77 \\
\multicolumn{1}{l|}{Fusing} & 91.21 & 42.63 & 65.84 & \multicolumn{1}{r|}{43.07} & 73.88 & 6.03 & 51.64 & \multicolumn{1}{r|}{7.15} & 90.55 & 58.54 & 71.55 & 57.12 \\
\multicolumn{1}{l|}{LNP} & 64.89 & 15.88 & 53.00 & \multicolumn{1}{r|}{16.63} & 58.52 & 10.65 & 52.01 & \multicolumn{1}{r|}{11.35} & 60.25 & 9.52 & 51.05 & 10.98 \\
\multicolumn{1}{l|}{SPrompts} & 90.19 & 39.73 & 63.55 & \multicolumn{1}{r|}{40.04} & 76.79 & 11.67 & 53.44 & \multicolumn{1}{r|}{12.58} & 88.45 & 24.02 & 57.55 & 24.75 \\
\multicolumn{1}{l|}{UnivFD} & {\ul 92.96} & {\ul 60.49} & {\ul 74.69} & \multicolumn{1}{r|}{{\ul 59.73}} & 77.59 & 30.39 & 60.22 & \multicolumn{1}{r|}{31.42} & {\ul 91.91} & {\ul 79.95} & {\ul 83.68} & {\ul 75.87} \\
\multicolumn{1}{l|}{LGrad} & 53.12 & 3.90 & 48.57 & \multicolumn{1}{r|}{4.46} & 51.93 & 2.12 & 50.30 & \multicolumn{1}{r|}{2.63} & 52.19 & 0 & 49.99 & 0.37 \\
\multicolumn{1}{l|}{NPR} & 47.60 & 12.98 & 50.95 & \multicolumn{1}{r|}{13.19} & 55.28 & 21.79 & 53.69 & \multicolumn{1}{r|}{21.92} & 48.07 & 47.65 & 49.97 & 47.73 \\
\multicolumn{1}{l|}{Freqnet} & 53.12 & 8.58 & 50.38 & \multicolumn{1}{r|}{9.00} & 53.83 & 3.55 & 50.20 & \multicolumn{1}{r|}{4.02} & 49.62 & 49.49 & 51.04 & 49.37 \\ \hline
\multicolumn{1}{l|}{\methodname} & \textbf{94.09} & \textbf{79.21} & \textbf{80.37} & \multicolumn{1}{r|}{\textbf{77.65}} & \textbf{80.98} & \textbf{47.77} & \textbf{68.02} & \multicolumn{1}{r|}{\textbf{47.68}} & \textbf{93.22} & \textbf{85.79} & \textbf{84.49} & \textbf{83.67} \\ \hline
 & \multicolumn{1}{l}{} & \multicolumn{1}{l}{} & \multicolumn{1}{l}{} & \multicolumn{1}{l}{} & \multicolumn{1}{l}{} & \multicolumn{1}{l}{} & \multicolumn{1}{l}{} & \multicolumn{1}{l}{} & \multicolumn{1}{l}{} & \multicolumn{1}{l}{} & \multicolumn{1}{l}{} & \multicolumn{1}{l}{} \\ \hline
\multicolumn{1}{c|}{\multirow{2}{*}{Methods}} & \multicolumn{4}{c|}{DiffusionForensics} & \multicolumn{4}{c|}{Ojha} & \multicolumn{4}{c}{DIF} \\ \cline{2-13} 
\multicolumn{1}{c|}{} & $\mathrm{AP}$ & $\mathrm{F1}$ & $\mathrm{ACC}$ & \multicolumn{1}{c|}{$\mathrm{AUC_{f1}}$} & $\mathrm{AP}$ & $\mathrm{F1}$ & $\mathrm{ACC}$ & \multicolumn{1}{c|}{$\mathrm{AUC_{f1}}$} & $\mathrm{AP}$ & $\mathrm{F1}$ & $\mathrm{ACC}$ & $\mathrm{AUC_{f1}}$ \\ \hline
\multicolumn{1}{l|}{CNNDet} & 64.70 & 11.14 & 57.31 & \multicolumn{1}{r|}{11.60} & 63.41 & 6.80 & 51.31 & \multicolumn{1}{r|}{7.50} & 74.31 & 34.65 & 62.50 & 34.61 \\
\multicolumn{1}{l|}{FreDect} & 45.53 & {\ul 34.48} & 49.16 & \multicolumn{1}{r|}{{\ul 31.71}} & 68.14 & 54.99 & 62.04 & \multicolumn{1}{r|}{50.49} & 76.76 & 55.08 & 67.57 & 50.98 \\
\multicolumn{1}{l|}{GramNet} & 77.71 & 28.89 & 60.31 & \multicolumn{1}{r|}{28.99} & 52.89 & 2.32 & 50.00 & \multicolumn{1}{r|}{2.65} & 49.87 & 9.73 & 43.45 & 10.00 \\
\multicolumn{1}{l|}{Fusing} & 62.89 & 1.57 & 56.23 & \multicolumn{1}{r|}{2.77} & 73.61 & 7.71 & 51.88 & \multicolumn{1}{r|}{9.53} & 81.64 & 33.92 & 63.36 & 34.01 \\
\multicolumn{1}{l|}{LNP} & {\ul 79.50} & 22.58 & {\ul 61.04} & \multicolumn{1}{r|}{22.86} & 46.45 & 7.84 & 49.80 & \multicolumn{1}{r|}{8.39} & 56.42 & 10.46 & 50.07 & 11.23 \\
\multicolumn{1}{l|}{SPrompts} & \textbf{81.01} & 12.97 & 57.38 & \multicolumn{1}{r|}{13.41} & 71.93 & 12.79 & 53.76 & \multicolumn{1}{r|}{13.53} & 79.02 & 29.71 & 59.97 & 30.07 \\
\multicolumn{1}{l|}{UnivFD} & 62.42 & 18.39 & 60.68 & \multicolumn{1}{r|}{19.65} & \textbf{93.87} & {\ul 67.45} & {\ul 76.19} & \multicolumn{1}{r|}{{\ul 65.54}} & {\ul 88.31} & {\ul 56.39} & {\ul 73.72} & {\ul 55.92} \\
\multicolumn{1}{l|}{LGrad} & 67.90 & 4.88 & 56.81 & \multicolumn{1}{r|}{5.33} & 38.07 & 0.20 & 49.86 & \multicolumn{1}{r|}{0.62} & 50.36 & 0.47 & 49.92 & 0.97 \\
\multicolumn{1}{l|}{NPR} & 56.78 & 10.12 & 53.26 & \multicolumn{1}{r|}{10.31} & 44.57 & 4.95 & 50.42 & \multicolumn{1}{r|}{5.28} & 45.75 & 8.68 & 47.68 & 8.90 \\
\multicolumn{1}{l|}{Freqnet} & 51.63 & 4.93 & 56.27 & \multicolumn{1}{r|}{5.26} & 50.82 & 2.71 & 50.01 & \multicolumn{1}{r|}{3.22} & 51.71 & 3.39 & 50.22 & 3.91 \\ \hline
\multicolumn{1}{l|}{\methodname} & 63.48 & \textbf{42.91} & \textbf{67.24} & \multicolumn{1}{r|}{\textbf{42.41}} & {\ul 91.98} & \textbf{81.79} & \textbf{82.90} & \multicolumn{1}{r|}{\textbf{79.46}} & \textbf{88.76} & \textbf{70.32} & \textbf{79.23} & \textbf{69.17} \\
\hline
\\
\end{tabular}
}

\resizebox{0.6\textheight}{!}{
\begin{tabular}{l|rrrr|rrrr|r}
\hline
\multicolumn{1}{c|}{} & \multicolumn{4}{c|}{Celeb-DF-v1/v2} & \multicolumn{4}{c|}{UADFV} & \multicolumn{1}{c}{AntiFake} \\ \cline{2-10} 
\multicolumn{1}{c|}{} & \multicolumn{1}{c}{$\mathrm{AP}$} & \multicolumn{1}{c}{$\mathrm{F1}$} & \multicolumn{1}{c}{$\mathrm{ACC}$} & \multicolumn{1}{c|}{$\mathrm{AUC_{f1}}$} & \multicolumn{1}{c}{$\mathrm{AP}$} & \multicolumn{1}{c}{$\mathrm{F1}$} & \multicolumn{1}{c}{$\mathrm{ACC}$} & \multicolumn{1}{c|}{$\mathrm{AUC_{f1}}$} & \multicolumn{1}{c}{$\mathrm{ACC}$} \\ \hline
CNNDet & 64.40/87.60 & 0.14/0.19 & 33.94/13.57 & 0.72/0.92 & 61.65 & 0.13 & 50.42 & 0.65 & 16.63 \\
FreDect & 72.54/88.10 & {\ul 53.93/55.18} & {\ul 51.88/43.61} & {\ul 46.57/51.06} & 62.42 & {\ul 52.11} & {\ul 60.03} & {\ul 44.37} & 27.46 \\
GramNet & 67.44/88.43 & 0.04/0.08 & 33.94/13.53 & 0.44/0.55 & 39.31 & 0 & 49.77 & 0.33 & 30.40 \\
Fusing & 61.87/86.66 & 0.21/0.56 & 33.86/13.70 & 1.32/2.23 & 53.18 & 0.65 & 50.49 & 1.69 & 15.72 \\
LNP & 64.16/86.76 & 1.66/3.30 & 34.11/14.75 & 2.17/3.96 & 44.87 & 10.18 & 51.73 & 10.34 & 20.06 \\
SPrompts & 66.38/87.30 & 1.75/1.94 & 34.33/14.29 & 2.53/2.82 & 46.20 & 11.72 & 48.01 & 12.34 & 22.84 \\
UnivFD & \textbf{83.73}/{\ul 92.49} & 16.64/10.28 & 39.66/18.10 & 20.54/14.96 & \textbf{91.50} & 43.08 & 63.87 & 44.23 & {\ul 38.19} \\
LGrad & 65.98/87.33 & 0.14/0.56 & 33.95/13.71 & 0.60/1.16 & 44.47 & 0.13 & 49.90 & 0.87 & 15.50 \\
NPR & 62.13/86.15 & 7.28/7.48 & 34.45/16.23 & 7.59/7.88 & 43.40 & 10.78 & 50.42 & 11.11 & 26.64 \\
Freqnet & 64.60/86.11 & 18.85/17.32 & 36.67/20.41 & 19.13/17.77 & 44.55 & 11.44 & 45.05 & 11.89 & 16.49 \\ \hline
\methodname & {\ul 83.69}/\textbf{93.16} & \textbf{71.28/76.95} & \textbf{65.93/65.69} & \textbf{65.69/71.35} & {\ul 85.92} & \textbf{79.17} & \textbf{77.86} & \textbf{71.41} & \textbf{48.24} \\ 
\hline 
\end{tabular}
}

\end{table}

\begin{figure}[]
\centering
\centerline{\includegraphics[width=0.9\linewidth]{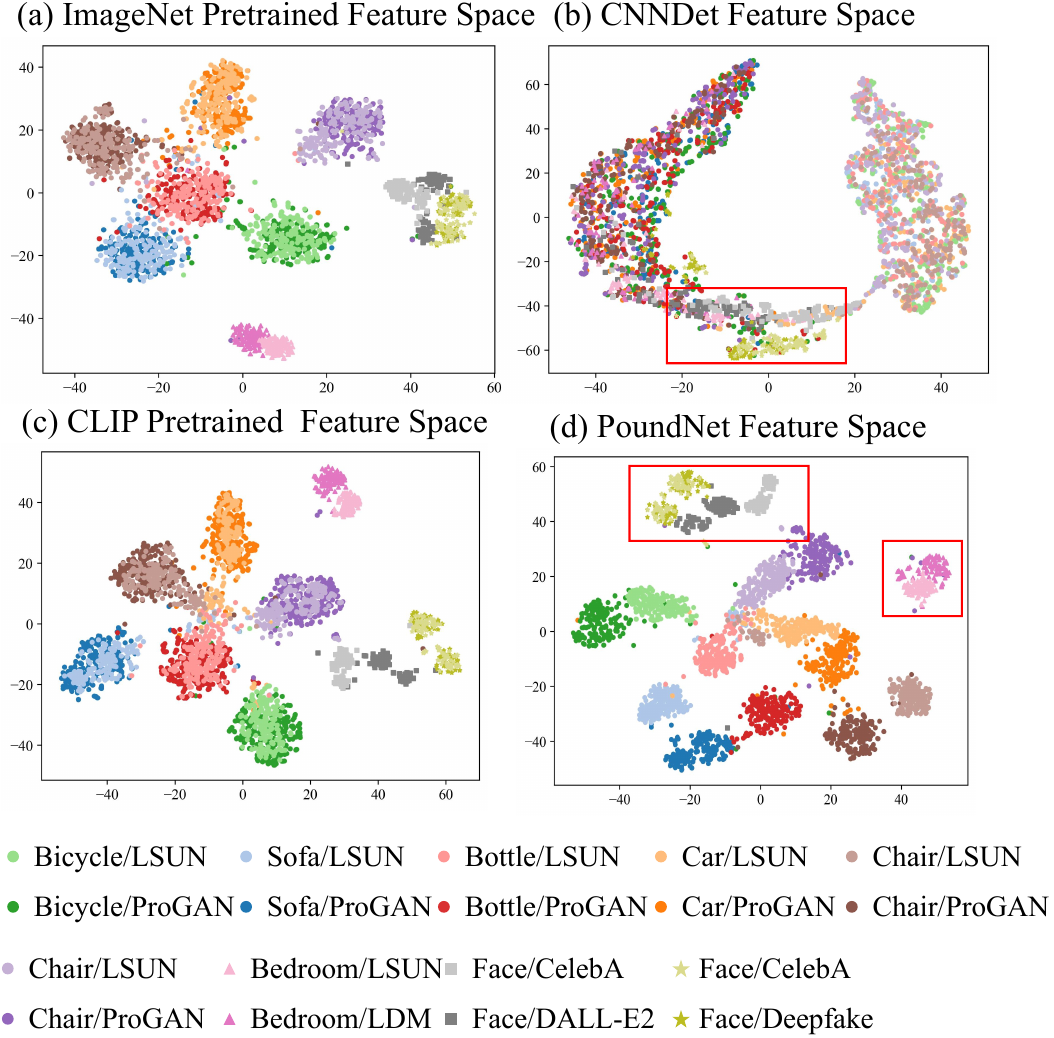}}
\caption{Feature spaces for in-domain deepfakes (produced by ProGAN) and three unseen deepfakes (generated by LDM, DALL-E2, and Deepfake). Reals:  [CLASS]/LSUN, [CLASS]/CelebA. }
\label{fig:featurespace}
\vspace{-1em}
\end{figure}

In addition, we examine the t-SNE visualization of feature spaces from two perspectives: 1) comparing pre-trained feature spaces (Fig.~\ref{fig:featurespace} (a, c)) of CNNDet and PoundNet, and 2) analyzing their evolution (Fig.~\ref{fig:featurespace} (b, d)) after fine-tuning on the same deepfake detection dataset. We use in-domain test data from ProGAN~\cite{karras2017progressive} and out-of-domain unseen deepfakes (generated by LDM~\cite{rombach2022high}, DALL-E 2~\cite{ramesh2022hierarchical} and Deepfake~\cite{rossler2019faceforensics++}) to assess generalization. As shown in Fig.~\ref{fig:featurespace}, CNNDet learns distinct feature representations for in-domain data but struggles with unseen deepfakes, indicated by overlapping clusters of real and fake samples. This suggests CNNDet's features are overly specialized to the training data. In contrast, PoundNet preserves the structure of the pre-trained feature space while adapting to deepfake detection, resulting in better generalization to unseen deepfakes. This is evidenced by more distinct clusters of real and fake samples in both in-domain and out-of-domain data, indicating PoundNet's ability to learn robust and transferable features.

\begin{table}
\centering
\caption{Impact of the two main hyperparameters $\lambda_1$ and $\lambda_2$ of the proposed PoundNet for deepfake detection on three used datasets. \zhiwu{Appendix \ref{sec:ablation} studies more values of $\lambda_1$ and $\lambda_2$.}
}
\label{tabl:blance_factor_eval}
\scalebox{0.8}{
\begin{tabular}{rr|rrrr|rrrr|rrrr|rrrr}
\hline
\multirow{2}{*}{$\lambda_1$} & \multirow{2}{*}{$\lambda_2$} & \multicolumn{4}{c|}{ForenS} & \multicolumn{4}{c|}{DIF} & \multicolumn{4}{c|}{Ojha} & \multicolumn{4}{c}{Averge} \\ \cline{3-18} 
 &  & \multicolumn{1}{c}{$\mathrm{AP}$} & \multicolumn{1}{c}{$\mathrm{F1}$} & \multicolumn{1}{c}{$\mathrm{ACC}$} & \multicolumn{1}{c|}{$\mathrm{AUC_{f1}}$} & \multicolumn{1}{c}{$\mathrm{AP}$} & \multicolumn{1}{c}{$\mathrm{F1}$} & \multicolumn{1}{c}{$\mathrm{ACC}$} & \multicolumn{1}{c|}{$\mathrm{AUC_{f1}}$} & \multicolumn{1}{c}{$\mathrm{AP}$} & \multicolumn{1}{c}{$\mathrm{F1}$} & \multicolumn{1}{c}{$\mathrm{ACC}$} & \multicolumn{1}{c|}{$\mathrm{AUC_{f1}}$} & \multicolumn{1}{c}{$\mathrm{AP}$} & \multicolumn{1}{c}{$\mathrm{F1}$} & \multicolumn{1}{c}{$\mathrm{ACC}$} & \multicolumn{1}{c}{$\mathrm{AUC_{f1}}$} \\ \hline
0 & 0 & 90.13 & 69.89 & 78.89 & 62.31 & 85.35 & 52.18 & 70.96 & 50.48 & 90.47 & 53.16 & 63.61 & 51.63 & 88.65 & 58.41 & 71.15 & 54.81 \\
0 & 1 & 95.30 & 74.42 & 81.25 & 72.69 & 87.61 & 58.45 & 74.51 & 57.54 & 93.18 & 61.16 & 72.84 & 59.70 & 92.03 & 64.68 & 76.20 & 63.31 \\
1 & 0 & 93.95 & 65.72 & 75.92 & 64.14 & 85.90 & 54.35 & 71.65 & 53.07 & 89.31 & 56.34 & 69.80 & 54.17 & 89.72 & 58.80 & 72.46 & 57.13 \\
1 & 1 & 94.09 & 79.21 & 80.37 & 77.65 & 88.76 & 70.32 & 79.23 & 69.17 & 91.98 & 81.79 & 82.90 & 79.46 & 91.61 & \textbf{77.11} & \textbf{80.83} & \textbf{75.43} \\
\hline
\end{tabular}
}
\vspace{-1em}
\end{table}

\textbf{Ablation analysis.}
This ablation study aims to validate the influence of the main components of the proposed PoundNet. 
We train PoundNet on ForenSynths~\cite{wang2020cnn} and test it on three datasets: ForenSynths, DIF, and Ojha, evaluating performance in terms of $\mathrm{AP}$, $\mathrm{F1}$, $\mathrm{ACC}$, and $\mathrm{AUC_{f1}}$.
The impact of prompt depth, prompt length, number of classes for training, model architectures and other affect factors are provided in Appendix~\ref{sec:ablation}.
The most important components of \methodname are the semantic-preserving term $\mathcal{L}_{spm}$ and class-aware binary term $\mathcal{L}_{cab}$, so we analyze the impact of the two hyperparameters, $\lambda_1$ and $\lambda_2$ given in Eq.~\eqref{eq:total_loss}. 
Table~\ref{tabl:blance_factor_eval} indicates that \methodname is robust to various hyperparameter values, with the best average performance observed when $\lambda_1=1$ and $\lambda_2=1$. This result aligns with our assumption of maintaining balance.

\section{Conclusion and Outlook}

This paper identifies a "Penny-Wise and Pound-Foolish" phenomenon in deepfake detection and proposes PoundNet, which uses a learnable prompt design and a balanced objective to address this issue. Evaluation on the largest test sets to date (comprising 10 deepfake detection benchmark datasets and 5 object classification datasets) demonstrates that PoundNet achieves a remarkable 19\% improvement in deepfake detection compared to state-of-the-art methods. Additionally, PoundNet maintains strong performance in object classification, achieving a 63\% accuracy, while other deepfake detection methods typically fail in this area.

For future work, we believe the proposed learning scheme can be applied to various other domains that involve transferring pre-trained deep learning models from upstream to downstream tasks. This approach would shed a light on the direction for continuous learning across a wide range of task categories. Furthermore, the current learning scheme has been tested in two continual learning sessions (pre-training for zero-shot object classification and fine-tuning on a single deepfake dataset for deepfake detection). Future work will explore extended continual learning sessions.

\bibliography{iclr2025_conference}
\bibliographystyle{iclr2025_conference}

\appendix


\appendix

In the appendix, we provide additional ablation studies on PoundNet's robustness and performance. We examine the effects of varying prompt depth and length, as well as the impact of training with different class numbers from the ForenSynths dataset. We also compare PoundNet's performance with various pre-trained models and assess its robustness to JPEG compression across multiple datasets. Additionally, we provide a detailed analysis of the benchmarks used for evaluation, along with the experimental results supporting the main paper.

\section{More Ablation Studies}
\label{sec:ablation}

\zhiwu{Table~\ref{tabl:blance_factor_eval1} supplements the main paper's study on the influence of the two hyperparameters $\lambda_1$ and $\lambda_2$, showing the proposed PoundNet is relatively insensitive to their varying settings.} Table~\ref{tab:ab_depth_length} studies the effects of different prompt depths ($J$) and prompt lengths ($M$) on the model's performance. 
The study is measured across the ForenSynths, DIF, and Ojha test sets, and presented as Average Precision (AP) and Accuracy (ACC) percentages. 
While deeper prompts and longer prompts tend to yield slightly better results, the model's performance does not drastically fluctuate with changes in different prompts, indicating robustness to this parameter.

\begin{table}[ht]
\centering
\caption{Impact of the two main hyperparameters $\lambda_1$ and $\lambda_2$ deepfake detection on three used datasets.
}
\label{tabl:blance_factor_eval1}
\scalebox{0.64}{
\begin{tabular}{rr|rrrr|rrrr|rrrr|rrrr}
\hline
\multirow{2}{*}{$\lambda_1$} & \multirow{2}{*}{$\lambda_2$} & \multicolumn{4}{c|}{ForenS} & \multicolumn{4}{c|}{DIF} & \multicolumn{4}{c|}{Ojha} & \multicolumn{4}{c}{Averge} \\ \cline{3-18} 
 &  & \multicolumn{1}{c}{$\mathrm{AP}$} & \multicolumn{1}{c}{$\mathrm{F1}$} & \multicolumn{1}{c}{$\mathrm{ACC}$} & \multicolumn{1}{c|}{$\mathrm{AUC_{f1}}$} & \multicolumn{1}{c}{$\mathrm{AP}$} & \multicolumn{1}{c}{$\mathrm{F1}$} & \multicolumn{1}{c}{$\mathrm{ACC}$} & \multicolumn{1}{c|}{$\mathrm{AUC_{f1}}$} & \multicolumn{1}{c}{$\mathrm{AP}$} & \multicolumn{1}{c}{$\mathrm{F1}$} & \multicolumn{1}{c}{$\mathrm{ACC}$} & \multicolumn{1}{c|}{$\mathrm{AUC_{f1}}$} & \multicolumn{1}{c}{$\mathrm{AP}$} & \multicolumn{1}{c}{$\mathrm{F1}$} & \multicolumn{1}{c}{$\mathrm{ACC}$} & \multicolumn{1}{c}{$\mathrm{AUC_{f1}}$} \\ \hline
0 & 0 & 90.13 & 69.89 & 78.89 & 62.31 & 85.35 & 52.18 & 70.96 & 50.48 & 90.47 & 53.16 & 63.61 & 51.63 & 88.65 & 58.41 & 71.15 & 54.81 \\
0 & 1 & 95.30 & 74.42 & 81.25 & 72.69 & 87.61 & 58.45 & 74.51 & 57.54 & 93.18 & 61.16 & 72.84 & 59.70 & 92.03 & 64.68 & 76.20 & 63.31 \\
1 & 0 & 93.95 & 65.72 & 75.92 & 64.14 & 85.90 & 54.35 & 71.65 & 53.07 & 89.31 & 56.34 & 69.80 & 54.17 & 89.72 & 58.80 & 72.46 & 57.13 \\
1 & 1 & 94.09 & 79.21 & 80.37 & 77.65 & 88.76 & 70.32 & 79.23 & 69.17 & 91.98 & 81.79 & 82.90 & 79.46 & 91.61 & \textbf{77.11} & \textbf{80.83} & \textbf{75.43} \\
0.1 & 0.1 & 94.93 & 78.87 & 85.11 & 77.14 & 90.03 & 63.12 & 76.95 & 62.36 & 92.12 & 69.93 & 77.26 & 67.90 & \textbf{92.36} & 70.64 & 79.77 & 69.13 \\
0.1 & 1 & 95.18 & 75.23 & 82.76 & 73.34 & 88.53 & 60.28 & 75.56 & 59.54 & 90.68 & 62.28 & 72.94 & 60.74 & 91.46 & 65.93 & 77.09 & 64.54 \\
1 & 0.1 & 94.20 & 77.84 & 81.90 & 75.56 & 86.53 & 62.03 & 75.72 & 60.55 & 93.63 & 70.02 & 77.79 & 67.49 & 91.45 & 69.96 & 78.47 & 67.87 \\
0.5 & 0.5 & 94.76 & 74.34 & 81.08 & 72.87 & 88.58 & 60.48 & 75.21 & 59.74 & 91.68 & 63.62 & 73.89 & 61.73 & 91.67 & 66.15 & 76.73 & 64.78 \\
0.5 & 1 & 94.35 & 79.29 & 82.93 & 77.56 & 87.41 & 65.56 & 77.28 & 64.74 & 92.85 & 75.29 & 79.96 & 72.89 & 91.54 & 73.38 & 80.06 & 71.73 \\
1 & 0.5 & 94.42 & 68.92 & 77.03 & 67.85 & 84.21 & 59.87 & 73.75 & 58.74 & 91.10 & 62.05 & 72.76 & 60.19 & 89.91 & 63.61 & 74.51 & 62.26 \\
2 & 2 & 93.56 & 76.04 & 82.19 & 74.14 & 85.72 & 60.91 & 74.34 & 60.04 & 86.23 & 61.43 & 71.75 & 58.95 & 88.50 & 66.13 & 76.09 & 64.38 \\
2 & 1 & 92.89 & 69.80 & 77.38 & 67.83 & 83.56 & 57.70 & 72.17 & 56.17 & 89.51 & 57.53 & 70.61 & 55.58 & 88.65 & 61.68 & 73.39 & 59.86 \\
1 & 2 & 91.33 & 73.07 & 81.55 & 72.01 & 82.83 & 61.32 & 76.32 & 60.42 & 90.34 & 72.35 & 78.70 & 70.13 & 88.17 & 68.91 & 78.86 & 67.52 \\ \hline
\end{tabular}
}
\vspace{-1.5em}
\end{table}

\begin{table}[h]
\centering
\caption{Influence of prompt depth $J$ and prompt length $M$ for deepfake detection on three used datasets.}
\label{tab:ab_depth_length}
\begin{tabular}{lrrrr}
\hline
\multicolumn{1}{c|}{\multirow{2}{*}{$J$}} & \multicolumn{4}{c}{Test   Set AP/ACC (\%)} \\ \cline{2-5} 
\multicolumn{1}{c|}{} & \multicolumn{1}{c}{ForenSynths} & \multicolumn{1}{c}{DIF} & \multicolumn{1}{c}{Ojha} & \multicolumn{1}{c}{Average} \\ \hline
\multicolumn{1}{r|}{1} & 94.11/84.54 & 85.89/76.9 & 86.49/73.46 & 88.83/78.30 \\
\multicolumn{1}{r|}{2} & 91.99/78.72 & 88.49/75.54 & 92.14/78.6 & 90.87/77.62 \\
\multicolumn{1}{r|}{4} & 94.39/76.82 & 89.77/73.92 & 90.89/72.13 & 91.68/74.29 \\
\multicolumn{1}{r|}{6} & 93.12/79.65 & 89.15/77.37 & 90.24/78.37 & 90.84/78.46 \\
\multicolumn{1}{r|}{8} & 94.09/80.37 & 88.76/79.23 & 91.98/82.9 & 91.61/80.83 \\ \hline
 & \multicolumn{1}{l}{} & \multicolumn{1}{l}{} & \multicolumn{1}{l}{} & \multicolumn{1}{l}{} \\ \hline
\multicolumn{1}{c|}{\multirow{2}{*}{$M$}} & \multicolumn{4}{c}{Test Set   AP/ACC (\%)} \\ \cline{2-5} 
\multicolumn{1}{c|}{} & \multicolumn{1}{c}{ForenSynths} & \multicolumn{1}{c}{DIF} & \multicolumn{1}{c}{Ojha} & \multicolumn{1}{c}{Average} \\ \hline
\multicolumn{1}{r|}{1} & 90.9/73.31 & 73.9/62.57 & 76.43/69.25 & 80.41/68.38 \\
\multicolumn{1}{r|}{2} & 93.61/81.21 & 88.95/78.3 & 87.52/80.64 & 90.03/80.05 \\
\multicolumn{1}{r|}{4} & 94.13/84.12 & 85.43/73.81 & 89.81/75.27 & 89.79/77.73 \\
\multicolumn{1}{r|}{8} & 92.01/82.85 & 86.48/77.78 & 91.11/83.34 & 89.87/81.32 \\
\multicolumn{1}{r|}{16} & 94.09/80.37 & 88.76/79.23 & 91.98/82.9 & 91.61/80.83 \\ \hline
\end{tabular}
\end{table}

Table~\ref{tab:ab_classnumber} shows the performance of PoundNet when trained with different numbers of classes from the ForenSynths training set, which contains a total of 20 classes. 
The results are presented as Average Precision (AP) and Accuracy (ACC) percentages across the ForenSynths, DIF, and Ojha test sets. 
Although PoundNet's performance generally improves with an increase in the number of classes, it remains relatively stable, indicating that the method is not sensitive to the number of classes used for training.

\begin{table}[]
\centering
\caption{Influence of the number of classes used for training PoundNet. }
\label{tab:ab_classnumber} 
\begin{tabular}{r|rrrr}
\hline
\multirow{2}{*}{Class Number} & \multicolumn{4}{c}{Test Set AP/ACC (\%)} \\ \cline{2-5} 
 & \multicolumn{1}{c}{ForenSynths} & \multicolumn{1}{c}{DIF} & \multicolumn{1}{c}{Ojha} & \multicolumn{1}{c}{Averge} \\ \hline
2 & 90.90/73.31 & 86.23/72.52 & 90.70/68.86 & 89.28/71.56 \\
4 & 92.42/79.87 & 88.08/77.92 & 93.09/77.88 & 91.20/78.56 \\
8 & 93.62/78.61 & 86.69/75.35 & 90.91/72.80 & 90.41/75.59 \\
16 & 93.26/80.83 & 85.95/76.45 & 87.57/72.86 & 88.93/76.71 \\
20 & 94.09/80.37 & 88.76/79.23 & 91.98/82.90 & 91.61/80.83 \\ \hline
\end{tabular}
\end{table}

Table~\ref{tab:ab_pretrain} compares the performance of various pre-trained models used to train PoundNet on the ForenSynths, DIF, and Ojha test sets. We keep all other hyperparameters the same, and only change the pre-trained weights of the CLIP architecture. We can see that larger models (ViT-L/14) generally offer better performance, with OpenAI ViT-L/14 being the best performer. This suggests that OpenAI ViT-L/14 is more suitable for deepfake detection.

\begin{table}[]
\centering
\caption{Performance of different pre-trained models.}
\label{tab:ab_pretrain}
\begin{tabular}{c|rccc}
\hline
\multirow{2}{*}{Pretrained Model} & \multicolumn{4}{c}{Test Set AP/ACC (\%)}                    \\ 
\cline{2-5} 
 & ForenSynths & DIF & \multicolumn{1}{c}{Ojha} & Average \\ 
 \hline
 \hline
Openai ViT-B/16  & 86.04/74.33 & 85.44/78.12 & \multicolumn{1}{c}{87.75/73.82}    & 86.41/75.42   \\
Openai ViT-L/14 & 94.09/80.37 & 88.76/79.23 & \multicolumn{1}{c}{91.98/82.90}    & 91.61/80.83  \\
OpenCLIP ViT-L/14 &  93.46/79.67 & 87.63/77.32 & \multicolumn{1}{c}{91.73/82.09}    & 90.94/79.69  \\
 \hline
\end{tabular}
\end{table}

\section{Implementation Details}
We utilize the architecture of CLIP ViT-L/14 as our detector, which contains 389 million parameters and has a storage size of around 1.4 GB in a PyTorch-compatible format. 
The prompts we add to CLIP are negligible in size, consisting of only 241K parameters with a depth of 8 and a length of 16. 
Using learnable prompts does not increase the inference time for the CLIP model. As a result, PoundNet can process approximately 300 samples per second on an A100 GPU under full optimization. 
Training PoundNet takes about 1 hour on an A100 GPU, with a single epoch.

The inference process is the same as other prompt-tuning methods, like CoOp~\cite{zhou2022conditional}, where we append class names to the learnable context to obtain the text prompts. 
PoundNet has a pair of contexts for real and fake images. 
Then, for each class (e.g., "Face"), we can generate a pair of text embeddings: one for the real class and one for the fake class. 
It would be preferable to use the grounding class names for the test samples to make predictions, as this aligns with the training objective of our method. 
However, since this may be too idealistic in real-world scenarios and requires more computation for inference, we use the mean of a wide spread of class embeddings to represent the general concept for testing. 
We compare the cosine similarity of image features, extracted by the CLIP vision encoder, with these text features to make predictions.
The class corresponding to the text feature with the highest cosine similarity to the image feature is selected as the predicted class.

For classical deepfake benchmarks such as Celeb-DF-v1/v2 and UADFV, we utilize a face detector to crop human faces from the images. This strong prior assumption allows us to use the class name "Face" for detection. We find that using specific class names can enhance the performance of deepfake detection. 


\section{Benchmark Details}

\label{sec:benchmark_details}

\textbf{Data-side Bias.}
Recent methodologies often assess their main performances without employing cropping, resizing, or compression techniques during the testing phase.
However, as highlighted in recent studies~\cite{grommelt2024fake}, real images in most datasets are predominantly stored in JPEG format, featuring a variety of sizes and aspect ratios, while their AI-generated counterparts are generally saved in the lossless PNG format with a uniform square aspect ratio.
Since JPEG compression is a prevalent post-processing operation, using raw images for primary evaluation does not accurately reflect real-world scenarios.
Such evaluation may inadvertently leverage bias to give some methods an unfair advantage.
To ensure that deepfake detectors are truly identifying deepfakes and not JPEG compression artifacts, recent studies~\cite{10.1145/3652027, yu2024diff} have applied JPEG compression to all test images, which significantly reduce the effectiveness of existing state-of-the-art methods.

\textbf{Metric-side Bias.}
In the evaluation of deepfake detection,  standard, widely used metrics include the ROC Area Under Curve (ROC-AUC) and Average Precision (AP).
These metrics can effectively reflect the sorting ability of a binary classifier and provide insight into its performance.

However, we find that these metrics can't fully reflect the performance of detectors when applied to unseen generators, also known as out-of-domain data. 
As illustrated in Figures 4 to 15, we present the precision-recall curves with thresholds across all benchmarks and methods. 
After we have a close look, we can find that many precision-recall curves have similar areas under the curve, indicating comparable AP or ROC-AUC.
However, some models achieve high precision and recall with a narrow threshold range.
Moreover, some F1-threshold curves have a dramatic drop near the $0$ threshold, indicating that some methods only achieve high F1 scores within a narrow, limited threshold range. 
This narrow score range can make detectors significantly sensitive to the chosen decision threshold, where small changes in the threshold result in substantial fluctuations in classification performance.

Thus, even ROC-AUC or Average Precision (AP) scores of some methods may be satisfactory, but this can be misleading. 
As it is not feasible to use a validation set to calibrate detectors for unseen deepfakes, a robust AI-generated image detector should ideally perform well across a wide range of threshold selections. 
This robustness ensures that the model's performance remains consistent and reliable, without being overly sensitive to the specific threshold chosen for classification.

To address this issue, we design a new metric to reflect the threshold sensitivity of detectors: $\mathrm{AUC_{f1}}$. 
This metric calculates the area under the curve of the F1 score across different thresholds. 
By evaluating the F1 score over a range of thresholds, $\mathrm{AUC_{f1}}$ provides a more comprehensive measure of a detector's robustness and performance consistency.

\begin{equation}
\label{eq:f1}
F1(\tau) = \frac{2 \cdot \text{Precision}(\tau) \cdot \text{Recall}(\tau)}{\text{Precision}(\tau) + \text{Recall}(\tau)}, 
\end{equation}

\begin{equation}
\label{eq:aucf1}
\mathrm{AUC_{f1}} = \int_0^{1} \mathrm{F1}(\tau) \, d\tau,
\end{equation}

\textbf{Datasets descriptions.} 
We validate our method on $10$ large, publicly available ai-generated image detection benchmarks with state-of-the-art methods. 
These benchmarks encompass a diverse range of generative models, including Generative Adversarial Networks (GANs), Flow models, and Diffusion techniques. 
Table~\ref{tab:10datasets_stas} provides statistics on various datasets used for benchmarking deepfake detection algorithms.

\begin{table}[]
\centering
\caption{Statistics of Datasets Used for Benchmarking. It includes the number of real and fake samples across different deepfake types for each dataset. }
\label{tab:10datasets_stas}
\small
\begin{tabular}{lrrrlrr}
\cline{1-3} \cline{5-7}
\multicolumn{3}{c}{ForenSynths} &  & \multicolumn{3}{c}{DIF} \\ \cline{1-3} \cline{5-7} 
\multicolumn{1}{c}{Deepfake Types} & \multicolumn{1}{c}{Num. Real} & \multicolumn{1}{c}{Num. Fake} & \multicolumn{1}{c}{} & \multicolumn{1}{c}{Deepfake Types} & \multicolumn{1}{c}{Num. Real} & \multicolumn{1}{c}{Num. Fake} \\ \cline{1-3} \cline{5-7} 
BigGAN & 2,000 & 2,000 &  & BigGAN & 1,488 & 1,488 \\
CRN & 6,382 & 6,382 &  & CycleGAN & 809 & 809 \\
CycleGAN & 1,321 & 1,321 &  & DALL-E 2 & 487 & 487 \\
DeepFake & 2,707 & 2,698 &  & DALL-E Mini & 2,498 & 2,498 \\
GauGAN & 5,000 & 5,000 &  & GauGAN & 4,488 & 4,488 \\
IMLE & 6,382 & 6,382 &  & GLIDE & 2,498 & 2,498 \\
ProGAN & 4,000 & 4,000 &  & MJ & 2,498 & 2,498 \\
SAN & 219 & 219 &  & ProGAN & 3,488 & 3,488 \\
SeeingDark & 180 & 180 &  & SD-v1.4 & 2,488 & 2,488 \\
StarGAN & 1,999 & 1,999 &  & SD-v2.1 & 2,488 & 2,488 \\
StyleGAN & 5,991 & 5,991 &  & StarGAN & 1,487 & 1,487 \\
StyleGAN2 & 7,988 & 7,988 &  & StyleGAN & 5,479 & 5,479 \\
Which Face Is Real & 1,000 & 1,000 &  & StyleGAN2 & 7,476 & 7,476 \\ \cline{1-3} \cline{5-7} 
 &  &  &  &  &  &  \\ \cline{1-3} \cline{5-7} 
\multicolumn{3}{c}{Ojha} &  & \multicolumn{3}{c}{GenImage} \\ \cline{1-3} \cline{5-7} 
\multicolumn{1}{c}{Deepfake Types} & \multicolumn{1}{c}{Num. Real} & \multicolumn{1}{c}{Num. Fake} & \multicolumn{1}{c}{} & \multicolumn{1}{c}{Deepfake Types} & \multicolumn{1}{c}{Num. Real} & \multicolumn{1}{c}{Num. Fake} \\ \cline{1-3} \cline{5-7} 
DALL-E & 1,000 & 1,000 &  & WuKong & 6,000 & 6,000 \\
GLIDE\_100\_10 & 1,000 & 1,000 &  & BigGAN & 2,000 & 2,000 \\
GLIDE\_100\_27 & 1,000 & 1,000 &  & MJ & 6,000 & 6,000 \\
GLIDE\_50\_27 & 1,000 & 1,000 &  & SD-v1.5 & 8,000 & 8,000 \\
Guided & 1,000 & 1,000 &  & ADM & 6,000 & 6,000 \\
LDM\_100 & 1,000 & 1,000 &  & GLIDE & 6,000 & 6,000 \\
LDM\_200 & 1,000 & 1,000 &  & SD-v1.4 & 6,000 & 6,000 \\
LDM\_200\_CFG & 1,000 & 1,000 &  & VQDM & 6,000 & 6,000 \\ \cline{1-3} \cline{5-7} 
 &  &  &  &  &  &  \\ \cline{1-3} \cline{5-7}
\multicolumn{3}{c}{GANGen-Detection} &  & \multicolumn{3}{c}{DiffusionForensics} \\ \cline{1-3} \cline{5-7} 
\multicolumn{1}{c}{Deepfake Types} & \multicolumn{1}{c}{Num. Real} & \multicolumn{1}{c}{Num. Fake} &  & \multicolumn{1}{c}{Deepfake Types} & \multicolumn{1}{c}{Num. Real} & \multicolumn{1}{c}{Num. Fake} \\ \cline{1-3} \cline{5-7} 
AttGAN & 2,000 & 2,000 &  & ImageNet SD-v1 & 5,000 & 10,000 \\
BEGAN & 2,000 & 2,000 &  & ImageNet ADM & 5,000 & 5,000 \\
CramerGAN & 2,000 & 2,000 &  & CelebA-HQ SD-v2 & 1,000 & 1,000 \\
GANimation & 2,000 & 2,000 &  & CelebA-HQ IF & 1,000 & 1,000 \\
InfoMaxGAN & 2,000 & 2,000 &  & CelebA-HQ DALL-E 2 & 1,000 & 500 \\
MMDGAN & 2,000 & 2,000 &  & CelebA-HQ MJ & 1,000 & 100 \\
RelGAN & 2,000 & 2,000 &  &  Bedroom SD-v1 Part1 & 1,000 & 1,000 \\
S3GAN & 2,000 & 2,000 &  &  Bedroom SD-v1 Part2 & 1,000 & 1,000 \\
SNGAN & 2,000 & 2,000 &  &  Bedroom ADM & 1,000 & 1,000 \\
STGAN & 2,000 & 2,000 &  &  Bedroom IDDPM & 1,000 & 1,000 \\ \cline{1-3}
 &  &  &  &  Bedroom DDPM & 1,000 & 768 \\ \cline{1-3}
\multicolumn{3}{c}{Antifake} &  &  Bedroom PNDM & 1,000 & 1,000 \\ \cline{1-3}
\multicolumn{1}{c}{Deepfake Types} & \multicolumn{1}{c}{Num. Real} & \multicolumn{1}{c}{Num. Fake} &  &  Bedroom SD-v2 & 1,000 & 1,000 \\ \cline{1-3}
COCO & 3,000 & - &  &  Bedroom LDM & 1,000 & 1,000 \\
Flickr & 3,000 & - &  &  Bedroom VQDiffusion & 1,000 & 1,000 \\
AdvAtk & - & 3,000 &  &  Bedroom IF & 1,000 & 1,000 \\
DALLE2 & - & 3,000 & \multicolumn{1}{l}{} &  Bedroom DALL-E 2 & 1,000 & 500 \\
Deeperforensics & - & 3,000 & \multicolumn{1}{l}{} &  Bedroom MJ & 1,000 & 100 \\ \cline{5-7} 
IF & - & 3,000 & \multicolumn{1}{l}{} &  & \multicolumn{1}{l}{} & \multicolumn{1}{l}{} \\ \cline{5-7} 
lte & - & 3,000 & \multicolumn{1}{l}{} & \multicolumn{3}{c}{Celeb-DF-v1} \\ \cline{5-7} 
SD2Inpaint & - & 3,000 & \multicolumn{1}{l}{} & \multicolumn{1}{c}{Deepfake Types} & \multicolumn{1}{l}{Real Videos} & \multicolumn{1}{l}{Fake Videos} \\ \cline{5-7} 
SDXL & - & 3,000 & \multicolumn{1}{l}{} & DeepFake & \multicolumn{1}{r}{408} & \multicolumn{1}{r}{795} \\ \cline{5-7} 
Backdoor & - & 3,000 & \multicolumn{1}{l}{} &  & \multicolumn{1}{l}{} & \multicolumn{1}{l}{} \\ \cline{5-7} 
Control & - & 3,000 & \multicolumn{1}{l}{} & \multicolumn{3}{c}{Celeb-DF-v1} \\ \cline{5-7} 
DataPoison & - & 3,000 & \multicolumn{1}{l}{} & \multicolumn{1}{c}{Deepfake Types} & \multicolumn{1}{l}{Real Videos} & \multicolumn{1}{l}{Fake Videos} \\ \cline{5-7} 
Lama & - & 3,000 & \multicolumn{1}{l}{} & DeepFake & \multicolumn{1}{r}{590} & \multicolumn{1}{r}{5,639} \\ \cline{5-7} 
SD2 & - & 3,000 & \multicolumn{1}{l}{} &  & \multicolumn{1}{l}{} & \multicolumn{1}{l}{} \\ \cline{5-7} 
SD2SuperRes & - & 3,000 & \multicolumn{1}{l}{} & \multicolumn{3}{c}{UADFV} \\ \cline{5-7} 
SGXL & - & 3,000 & \multicolumn{1}{l}{} & \multicolumn{1}{c}{Deepfake Types} & \multicolumn{1}{l}{Real Videos} & \multicolumn{1}{l}{Fake Videos} \\ \cline{1-3} \cline{5-7} 
 & \multicolumn{1}{l}{} & \multicolumn{1}{l}{} & \multicolumn{1}{l}{} & DeepFake & \multicolumn{1}{r}{49} & \multicolumn{1}{r}{49} \\ \cline{5-7} 
\end{tabular}
\end{table}

\section{Influence of JPEG Compression}
\label{sec:jpeg_compression}

We explore the robustness of our method to JPEG compression, a common form of image degradation that can affect the performance of deepfake detection models. To assess the impact of JPEG compression on various methods, we conduct experiments and present the results in Table~\ref{tab:7datasets_withoutjpg}, Table~\ref{tab:7datasets_80jpg}, and Table~\ref{tab:7datasets_70jpg}.

The results demonstrate that as the image quality decreases due to JPEG compression, all models experience a decline in performance. This observation highlights the fact that interfering factors, such as compression artifacts, significantly increase the difficulty of the AI-generated image detection task. However, despite the performance degradation, our proposed model still outperforms the latest state-of-the-art methods, showcasing its robustness to JPEG compression.

Furthermore, we observe that methods relying heavily on low-level features, such as pixel-level artifacts or high-frequency noise patterns, suffer a more dramatic decrease in performance when faced with JPEG compression. This is because JPEG compression tends to smooth out high-frequency details and introduce blocking artifacts, which can obscure or distort the subtle low-level cues that these methods rely on for detection.

In contrast, our approach, which leverages high-level semantic features learned through the pre-training of the CLIP model, demonstrates greater resilience to JPEG compression. By focusing on more abstract and robust representations of the image content, our method is less sensitive to the local distortions introduced by compression.

\begin{table}[]
\centering
\caption{Comprehensive comparisons of our method and other AI-generated images detectors on $7$ datasets without unified JPEG Compression on all real and fake images. \zhiwu{\emph{NOTE: In this scenario, all fake images are in PNG format, while all real images are in JPEG format, resulting in a highly biased evaluation. The following results are provided for reference only, and we do not endorse the outcomes in this table. For less biased results, please refer to Table~\ref{tab:7datasets_80jpg} and Table~\ref{tab:7datasets_70jpg}.}} $\mathbf{ACC_r}$, $\mathbf{ACC_f}$ represent detection accurracies for real images and fake images, respectively. For each test subset, the best results are highlighted in boldface and the second best results are underlined.
}
\label{tab:7datasets_withoutjpg}
\small
\resizebox{0.65\textheight}{!}{

\begin{tabular}{rrrrrrrrrrrrrrrrr}
\hline
\multicolumn{9}{c}{ForenSynths} & \multicolumn{8}{c}{GANGen-Detection} \\ \hline
\multicolumn{1}{r|}{Method} & $\mathbf{AP}$ & $\mathbf{F1}$ & $\mathbf{ACC_r}$ & $\mathbf{ACC_f}$ & $\mathbf{ACC}$ & $\mathbf{AUC_{roc}}$ & $\mathbf{AUC_{f1}}$ & \multicolumn{1}{r|}{$\mathbf{AUC_{f2}}$} & $\mathbf{AP}$ & $\mathbf{F1}$ & $\mathbf{ACC_r}$ & $\mathbf{ACC_f}$ & $\mathbf{ACC}$ & $\mathbf{AUC_{roc}}$ & $\mathbf{AUC_{f1}}$ & $\mathbf{AUC_{f2}}$ \\ \hline
\multicolumn{1}{r|}{CNNDet} & 89.68 & 69.35 & 91.6 & 64.45 & 78.03 & 90.53 & 68.8 & \multicolumn{1}{r|}{65.63} & 92.97 & 76.56 & 97.87 & 68.19 & 83.03 & 93.23 & 75.59 & 70.42 \\
\multicolumn{1}{r|}{FreDect} & 75.75 & 63.11 & 71.7 & 65.64 & 68.67 & 75.5 & 61.53 & \multicolumn{1}{r|}{63.03} & 60.64 & 30.59 & 78.13 & 29.58 & 53.86 & 62.14 & 35.88 & 36.78 \\
\multicolumn{1}{r|}{Freqnet} & 87.69 & 79.17 & 74.92 & 82.27 & 78.59 & 88.6 & 78.86 & \multicolumn{1}{r|}{80.08} & 84.73 & 76.23 & 40.34 & 94.76 & 67.55 & 85.84 & 75.92 & 85.23 \\
\multicolumn{1}{r|}{Fusing} & 91.75 & 72.72 & 93.29 & 67.17 & 80.23 & 92.36 & 71.92 & \multicolumn{1}{r|}{68.4} & 91.83 & 75.83 & 96.52 & 66.97 & 81.74 & 91.71 & 73.8 & 68.57 \\
\multicolumn{1}{r|}{GramNet} & 74.68 & 73.83 & 51.13 & 86.15 & 68.65 & 77.16 & 73.75 & \multicolumn{1}{r|}{79.67} & 88.38 & 73.34 & 89.11 & 71.44 & 80.27 & 89.73 & 73.04 & 71.63 \\
\multicolumn{1}{r|}{LGrad} & 77.24 & 49.56 & 82.79 & 43 & 62.9 & 77.31 & 49.27 & \multicolumn{1}{r|}{44.73} & 85.54 & 20.67 & 99.15 & 12.15 & 55.65 & 86.52 & 21.61 & 15.58 \\
\multicolumn{1}{r|}{LNP} & 80.45 & 66.62 & 76.21 & 67.25 & 71.74 & 81.75 & 66.08 & \multicolumn{1}{r|}{65.85} & 65.22 & 26.76 & 97.61 & 21.18 & 59.39 & 64.88 & 27.17 & 23.51 \\
\multicolumn{1}{r|}{NPR} & 82.43 & 84.15 & 61.92 & 96.69 & 79.3 & 84.67 & 83.85 & \multicolumn{1}{r|}{90.44} & 65.54 & 72.2 & 27.27 & 95.21 & 61.24 & 68.49 & 72.11 & 83.54 \\
\multicolumn{1}{r|}{PoundNet} & 96.12 & 84.2 & 76.71 & 89.14 & 82.92 & 95.91 & 82.95 & \multicolumn{1}{r|}{85.42} & 96.72 & 92.59 & 88.78 & 95.85 & 92.31 & 96.55 & 90.95 & 93.21 \\
\multicolumn{1}{r|}{SPrompts} & 97.39 & 86.06 & 78.67 & 90.59 & 84.63 & 97.14 & 85.6 & \multicolumn{1}{r|}{87.56} & 98.6 & 89.16 & 99.45 & 87.74 & 93.6 & 98.46 & 88.32 & 87.35 \\
\multicolumn{1}{r|}{UnivFD} & 93.9 & 68.12 & 98.63 & 59.49 & 79.06 & 93.46 & 67.08 & \multicolumn{1}{r|}{62.01} & 94.5 & 86.65 & 91.68 & 85.96 & 88.82 & 93.97 & 83.08 & 83.03 \\ \hline
 &  &  &  &  &  &  &  &  &  &  &  &  &  &  &  &  \\ \hline
\multicolumn{9}{c}{DIF} & \multicolumn{8}{c}{GenImage} \\ \hline
\multicolumn{1}{r|}{Method} & $\mathbf{AP}$ & $\mathbf{F1}$ & $\mathbf{ACC_r}$ & $\mathbf{ACC_f}$ & $\mathbf{ACC}$ & $\mathbf{AUC_{roc}}$ & $\mathbf{AUC_{f1}}$ & \multicolumn{1}{r|}{$\mathbf{AUC_{f2}}$} & $\mathbf{AP}$ & $\mathbf{F1}$ & $\mathbf{ACC_r}$ & $\mathbf{ACC_f}$ & $\mathbf{ACC}$ & $\mathbf{AUC_{roc}}$ & $\mathbf{AUC_{f1}}$ & $\mathbf{AUC_{f2}}$ \\ \hline
\multicolumn{1}{r|}{CNNDet} & 78.19 & 52.29 & 95.77 & 44.36 & 70.06 & 79.08 & 51.89 & \multicolumn{1}{r|}{46.73} & 63.34 & 28.12 & 92.4 & 18.55 & 55.47 & 66.44 & 28.39 & 21.94 \\
\multicolumn{1}{r|}{FreDect} & 76.3 & 63.14 & 80.48 & 63.22 & 71.85 & 75.94 & 61.67 & \multicolumn{1}{r|}{61.66} & 57.84 & 44.72 & 69.32 & 43.16 & 56.24 & 57.48 & 43.88 & 43.29 \\
\multicolumn{1}{r|}{Freqnet} & 91.57 & 77.84 & 93.44 & 73.12 & 83.28 & 91.85 & 77.5 & \multicolumn{1}{r|}{74.42} & 83.28 & 67.63 & 89.06 & 59.44 & 74.25 & 84.16 & 67.21 & 62.02 \\
\multicolumn{1}{r|}{Fusing} & 83.76 & 54.59 & 97.81 & 46.73 & 72.27 & 83.93 & 54.18 & \multicolumn{1}{r|}{49.12} & 72.98 & 23.12 & 97.55 & 15.57 & 56.56 & 75.54 & 23.97 & 18.73 \\
\multicolumn{1}{r|}{GramNet} & 77.39 & 78.45 & 62.75 & 87.12 & 74.94 & 81.02 & 78.28 & \multicolumn{1}{r|}{82.88} & 63.49 & 69.82 & 35.39 & 88.41 & 61.9 & 68.63 & 69.74 & 79.77 \\
\multicolumn{1}{r|}{LGrad} & 84.25 & 55.8 & 95.28 & 43.34 & 69.31 & 84.86 & 55.51 & \multicolumn{1}{r|}{47.41} & 71.15 & 49.79 & 85.95 & 38.97 & 62.46 & 72.01 & 49.64 & 42.72 \\
\multicolumn{1}{r|}{LNP} & 80.44 & 68.37 & 88.07 & 61.42 & 74.75 & 81.82 & 67.54 & \multicolumn{1}{r|}{63.38} & 69.4 & 48.57 & 87.9 & 37.4 & 62.65 & 69.4 & 48.05 & 41.17 \\
\multicolumn{1}{r|}{NPR} & 92.03 & 88.96 & 79.99 & 95.39 & 87.69 & 93.98 & 88.61 & \multicolumn{1}{r|}{92.24} & 87.21 & 80.83 & 63.49 & 92.71 & 78.1 & 89.11 & 80.58 & 87.24 \\
\multicolumn{1}{r|}{PoundNet} & 92.97 & 79.56 & 91.82 & 78.13 & 84.98 & 94 & 78.46 & \multicolumn{1}{r|}{77.52} & 83.76 & 63.02 & 93.72 & 53.64 & 73.68 & 84.52 & 61.33 & 56.24 \\
\multicolumn{1}{r|}{SPrompts} & 94.32 & 83.69 & 96.96 & 77.97 & 87.46 & 94.76 & 82.49 & \multicolumn{1}{r|}{79.11} & 90.63 & 66.04 & 98.06 & 52.88 & 75.47 & 89.86 & 64.83 & 56.86 \\
\multicolumn{1}{r|}{UnivFD} & 90.96 & 66.29 & 98.97 & 59.84 & 79.41 & 91.08 & 65.59 & \multicolumn{1}{r|}{61.7} & 79.94 & 39.27 & 98.62 & 31.34 & 64.98 & 81.02 & 40.1 & 35.04 \\ \hline
 &  &  &  &  &  &  &  &  &  &  &  &  &  &  &  &  \\ \hline
\multicolumn{9}{c|}{DiffusionForensics} & \multicolumn{8}{c}{Ojha} \\ \hline
\multicolumn{1}{r|}{Method} & $\mathbf{AP}$ & $\mathbf{F1}$ & $\mathbf{ACC_r}$ & $\mathbf{ACC_f}$ & $\mathbf{ACC}$ & $\mathbf{AUC_{roc}}$ & $\mathbf{AUC_{f1}}$ & \multicolumn{1}{r|}{$\mathbf{AUC_{f2}}$} & $\mathbf{AP}$ & $\mathbf{F1}$ & $\mathbf{ACC_r}$ & $\mathbf{ACC_f}$ & $\mathbf{ACC}$ & $\mathbf{AUC_{roc}}$ & $\mathbf{AUC_{f1}}$ & $\mathbf{AUC_{f2}}$ \\ \hline
\multicolumn{1}{r|}{CNNDet} & 79.88 & 28.93 & 99.62 & 20.57 & 62.92 & 81.88 & 29.15 & \multicolumn{1}{r|}{23.47} & 76.12 & 29.02 & 98.72 & 17.66 & 58.19 & 75.68 & 29.22 & 21.43 \\
\multicolumn{1}{r|}{FreDect} & 58.76 & 45.6 & 67.78 & 50.81 & 58.29 & 59.11 & 42.64 & \multicolumn{1}{r|}{44.43} & 81.5 & 69.75 & 83.24 & 66.35 & 74.79 & 82.93 & 66.89 & 65.87 \\
\multicolumn{1}{r|}{Freqnet} & 96.12 & 75.03 & 99.51 & 65.55 & 83.07 & 95.79 & 74.32 & \multicolumn{1}{r|}{68.2} & 95.35 & 87.27 & 98.32 & 80.36 & 89.34 & 94.3 & 86.79 & 82.49 \\
\multicolumn{1}{r|}{Fusing} & 79.84 & 19.97 & 99.99 & 12.64 & 61.09 & 82.45 & 20.94 & \multicolumn{1}{r|}{15.99} & 80 & 27.81 & 99.45 & 16.61 & 58.03 & 78.43 & 28.16 & 20.46 \\
\multicolumn{1}{r|}{GramNet} & 93.09 & 71.01 & 93.15 & 76.01 & 84.01 & 93.64 & 70.97 & \multicolumn{1}{r|}{73.1} & 94.68 & 90.08 & 89.46 & 89.94 & 89.7 & 95.41 & 89.83 & 89.69 \\
\multicolumn{1}{r|}{LGrad} & 96.8 & 64.95 & 99.91 & 55.27 & 79.03 & 96.6 & 64.44 & \multicolumn{1}{r|}{58.07} & 84.87 & 64.36 & 96.85 & 51.88 & 74.36 & 82.09 & 63.95 & 55.91 \\
\multicolumn{1}{r|}{LNP} & 92.78 & 55.79 & 99.61 & 46.31 & 73.71 & 93.07 & 55.08 & \multicolumn{1}{r|}{48.85} & 82.15 & 63.63 & 92.5 & 50.79 & 71.64 & 80.95 & 62.41 & 54.69 \\
\multicolumn{1}{r|}{NPR} & 98.64 & 90.92 & 95.4 & 92.23 & 93.23 & 99.15 & 90.64 & \multicolumn{1}{r|}{90.87} & 97.4 & 94.14 & 94.17 & 93.79 & 93.98 & 97.58 & 93.69 & 93.48 \\
\multicolumn{1}{r|}{PoundNet} & 85.96 & 59.5 & 96.2 & 51.94 & 77.23 & 93.75 & 58.7 & \multicolumn{1}{r|}{54.42} & 97.87 & 91.63 & 97.16 & 87.79 & 92.47 & 97.81 & 89.26 & 87.1 \\
\multicolumn{1}{r|}{SPrompts} & 94.62 & 64.23 & 99.65 & 55.34 & 78.46 & 95.27 & 61.6 & \multicolumn{1}{r|}{56.15} & 97.97 & 84.31 & 99.79 & 75.52 & 87.66 & 97.81 & 83.26 & 77.85 \\
\multicolumn{1}{r|}{UnivFD} & 66.87 & 25.19 & 97.72 & 18.51 & 63.48 & 70.54 & 26.14 & \multicolumn{1}{r|}{22.04} & 95.85 & 77.41 & 99.04 & 65.42 & 82.23 & 95.48 & 74.62 & 67.77 \\ \hline
 &  &  &  &  &  &  &  &  &  &  &  &  &  &  &  &  \\ \cline{1-10}
\multicolumn{2}{c|}{AntiFake} & \multicolumn{8}{c}{Average} &  &  &  &  &  &  &  \\ \cline{1-10}
\multicolumn{1}{r|}{Method} & \multicolumn{1}{r|}{$\mathbf{ACC_r}$} & $\mathbf{AP}$ & $\mathbf{F1}$ & $\mathbf{ACC_r}$ & $\mathbf{ACC_f}$ & $\mathbf{ACC}$ & $\mathbf{AUC_{roc}}$ & $\mathbf{AUC_{f1}}$ & $\mathbf{AUC_{f2}}$ &  &  &  &  &  &  &  \\ \cline{1-10}
\multicolumn{1}{r|}{CNNDet} & \multicolumn{1}{r|}{20.25} & 80.03 & 47.38 & 96.00 & 38.96 & 61.14 & 81.14 & 47.17 & 41.60 &  &  &  &  &  &  &  \\
\multicolumn{1}{r|}{FreDect} & \multicolumn{1}{r|}{42.04} & 68.47 & 52.82 & 75.11 & 53.13 & 60.82 & 68.85 & 52.08 & 52.51 &  &  &  &  &  &  &  \\
\multicolumn{1}{r|}{Freqnet} & \multicolumn{1}{r|}{35.86} & 89.79 & 77.20 & 82.60 & 75.92 & 73.13 & 90.09 & 76.77 & 75.41 &  &  &  &  &  &  &  \\
\multicolumn{1}{r|}{Fusing} & \multicolumn{1}{r|}{21.97} & 83.36 & 45.67 & \textbf{97.44} & 37.62 & 61.70 & 84.07 & 45.50 & 40.21 &  &  &  &  &  &  &  \\
\multicolumn{1}{r|}{GramNet} & \multicolumn{1}{r|}{46.76} & 81.95 & 76.09 & 70.17 & 83.18 & 72.32 & 84.27 & 75.94 & {\ul 79.46} &  &  &  &  &  &  &  \\
\multicolumn{1}{r|}{LGrad} & \multicolumn{1}{r|}{29.64} & 83.31 & 50.86 & 93.32 & 40.77 & 61.91 & 83.23 & 50.74 & 44.07 &  & {\ul } &  &  &  &  &  \\
\multicolumn{1}{r|}{LNP} & \multicolumn{1}{r|}{32.16} & 78.41 & 54.96 & 90.32 & 47.39 & 63.72 & 78.65 & 54.39 & 49.58 &  &  &  &  &  &  &  \\
\multicolumn{1}{r|}{NPR} & \multicolumn{1}{r|}{54.78} & 87.21 & \textbf{85.20} & 70.37 & \textbf{94.34} & 78.33 & 88.83 & \textbf{84.91} & \textbf{89.64} &  &  &  &  &  &  &  \\
\multicolumn{1}{r|}{PoundNet} & \multicolumn{1}{r|}{52.66} & {\ul 92.23} & 78.42 & 90.73 & {\ul 76.08} & \textbf{79.46} & {\ul 93.76} & 76.94 & 75.65 & \textbf{} &  & \textbf{} & \textbf{} & \textbf{} & \textbf{} & \textbf{} \\
\multicolumn{1}{r|}{SPrompts} & \multicolumn{1}{r|}{45.88} & \textbf{95.59} & {\ul 78.92} & 95.43 & 73.34 & {\ul 79.02} & \textbf{95.55} & {\ul 77.68} & 74.15 &  & \textbf{} &  &  &  &  &  \\
\multicolumn{1}{r|}{UnivFD} & \multicolumn{1}{r|}{42.99} & 87.00 & 60.49 & \textbf{97.44} & 53.43 & 71.57 & 87.59 & 59.44 & 55.27 & {\ul } &  & {\ul } & {\ul } & \textbf{} & {\ul } & {\ul } \\ \cline{1-10}
\end{tabular}
}
\end{table}

\begin{table}[]
\centering
\caption{Comprehensive comparisons of our method and other AI-generated images detectors on $7$ datasets with \zhiwu{unified JPEG Compression $80\%$ on all the real and fake images}. $\mathbf{ACC_r}$, $\mathbf{ACC_f}$ represent detection accurracies for real images and fake images, respectively. For each test subset, the best results are highlighted in boldface and the second best results are underlined.
}
\label{tab:7datasets_80jpg}
\small
\resizebox{0.65\textheight}{!}{

\begin{tabular}{rrrrrrrrrrrrrrrrr}
\hline
\multicolumn{9}{c|}{ForenSynths} & \multicolumn{8}{c}{GANGen-Detection} \\ \hline
\multicolumn{1}{r|}{Method} & {$\mathbf{AP}$} & $\mathbf{F1}$ & $\mathbf{ACC_r}$ & $\mathbf{ACC_f}$ & $\mathbf{ACC}$ & $\mathbf{AUC_{roc}}$ & $\mathbf{AUC_{f1}}$ & \multicolumn{1}{r|}{$\mathbf{AUC_{f2}}$} & $\mathbf{AP}$ & $\mathbf{F1}$ & $\mathbf{ACC_r}$ & $\mathbf{ACC_f}$ & $\mathbf{ACC}$ & $\mathbf{AUC_{roc}}$ & $\mathbf{AUC_{f1}}$ & $\mathbf{AUC_{f2}}$ \\ \hline
\multicolumn{1}{r|}{CNNDet} & {88.15} & 49.4 & 98.05 & 37.8 & 67.93 & 88.6 & 49.05 & \multicolumn{1}{r|}{41.62} & 79.51 & 34.94 & 98.78 & 23.09 & 60.93 & 79.44 & 35.23 & 27.7 \\
\multicolumn{1}{r|}{FreDect} & {67.15} & 55.54 & 72.37 & 52.11 & 62.24 & 66.25 & 47.65 & \multicolumn{1}{r|}{47.11} & 60.81 & 21.35 & 87.08 & 22.87 & 54.97 & 61.41 & 24.16 & 24.81 \\
\multicolumn{1}{r|}{Freqnet} & {50.58} & 5.49 & 96.49 & 3.85 & 50.18 & 50.22 & 5.95 & \multicolumn{1}{r|}{4.85} & 49.31 & 47.9 & 36.62 & 65.64 & 51.13 & 49.84 & 47.65 & 56.48 \\
\multicolumn{1}{r|}{Fusing} & {89.69} & 38.83 & 99.49 & 29.86 & 64.68 & 89.72 & 39.41 & \multicolumn{1}{r|}{33.52} & 88.23 & 57.82 & 97.1 & 44.39 & 70.74 & 88.38 & 56.01 & 48.41 \\
\multicolumn{1}{r|}{GramNet} & {56.28} & 12.49 & 83.83 & 16.6 & 50.22 & 57.79 & 12.78 & \multicolumn{1}{r|}{14.8} & 51.08 & 0.21 & 99.83 & 0.11 & 49.97 & 51.66 & 0.53 & 0.54 \\
\multicolumn{1}{r|}{LGrad} & {52.54} & 1.77 & 95.42 & 1.5 & 48.47 & 53.14 & 2.19 & \multicolumn{1}{r|}{2.09} & 50.22 & 0.05 & 99.99 & 0.03 & 50.01 & 50.81 & 0.38 & 0.45 \\
\multicolumn{1}{r|}{LNP} & {60.63} & 8.22 & 96.07 & 7.47 & 51.78 & 62.28 & 8.69 & \multicolumn{1}{r|}{8.07} & 56.5 & 8.16 & 96.06 & 4.61 & 50.33 & 57.33 & 9.45 & 6.94 \\
\multicolumn{1}{r|}{NPR} & {46.73} & 8.17 & 92.74 & 7.56 & 50.15 & 43.45 & 8.54 & \multicolumn{1}{r|}{8.07} & 48.47 & 46.85 & 29.9 & 70.08 & 49.99 & 47.97 & 46.94 & 58.56 \\
\multicolumn{1}{r|}{PoundNet} & {93.09} & 72.54 & 91.23 & 68.42 & 79.83 & 92.83 & 70.85 & \multicolumn{1}{r|}{68.4} & 90.71 & 72.61 & 95.12 & 63.68 & 79.4 & 90.65 & 69.59 & 64.53 \\
\multicolumn{1}{r|}{SPrompts} & {82.67} & 17.29 & 95.51 & 14.07 & 54.8 & 80.76 & 17.77 & \multicolumn{1}{r|}{15.36} & 78.04 & 2.32 & 100 & 1.21 & 50.61 & 76.55 & 3.06 & 2.19 \\
\multicolumn{1}{r|}{UnivFD} & {91.54} & 58.63 & 98.69 & 48.19 & 73.44 & 90.91 & 57.71 & \multicolumn{1}{r|}{51.5} & 90.29 & 75.87 & 94.53 & 67.32 & 80.92 & 89.39 & 71.74 & 67.6 \\ \hline
 &  &  &  &  &  &  &  &  &  &  &  &  &  &  &  &  \\ \hline
\multicolumn{9}{c|}{DIF} & \multicolumn{8}{c}{GenImage} \\ \hline
\multicolumn{1}{r|}{Method} & {$\mathbf{AP}$} & $\mathbf{F1}$ & $\mathbf{ACC_r}$ & $\mathbf{ACC_f}$ & $\mathbf{ACC}$ & $\mathbf{AUC_{roc}}$ & $\mathbf{AUC_{f1}}$ & \multicolumn{1}{r|}{$\mathbf{AUC_{f2}}$} & $\mathbf{AP}$ & $\mathbf{F1}$ & $\mathbf{ACC_r}$ & $\mathbf{ACC_f}$ & $\mathbf{ACC}$ & $\mathbf{AUC_{roc}}$ & $\mathbf{AUC_{f1}}$ & $\mathbf{AUC_{f2}}$ \\ \hline
\multicolumn{1}{r|}{CNNDet} & {75.23} & 31.97 & 98.81 & 24.73 & 61.77 & 75.92 & 32.11 & \multicolumn{1}{r|}{27.37} & 68 & 7.35 & 99.4 & 4.17 & 51.78 & 70.55 & 8.2 & 5.88 \\
\multicolumn{1}{r|}{FreDect} & {75.5} & 55.59 & 85.8 & 47.27 & 66.54 & 77.15 & 50.3 & \multicolumn{1}{r|}{47.07} & 78.58 & 52.04 & 90.98 & 40.85 & 65.91 & 79.06 & 48.43 & 43.18 \\
\multicolumn{1}{r|}{Freqnet} & {49.25} & 2.23 & 99.2 & 1.16 & 50.18 & 47.21 & 2.68 & \multicolumn{1}{r|}{1.94} & 51.21 & 1.97 & 99.24 & 1.01 & 50.13 & 49.04 & 2.39 & 1.74 \\
\multicolumn{1}{r|}{Fusing} & {81.71} & 34.45 & 99.78 & 27.34 & 63.56 & 81.51 & 34.5 & \multicolumn{1}{r|}{29.89} & 72.73 & 6.59 & 99.84 & 3.82 & 51.83 & 75.02 & 7.73 & 5.67 \\
\multicolumn{1}{r|}{GramNet} & {45.35} & 2.49 & 89.64 & 1.63 & 45.64 & 39.16 & 2.78 & \multicolumn{1}{r|}{2.28} & 68.13 & 14.25 & 98.06 & 9.34 & 53.7 & 74.56 & 14.45 & 11.11 \\
\multicolumn{1}{r|}{LGrad} & {48.51} & 0.19 & 99.9 & 0.1 & 50 & 47.04 & 0.58 & \multicolumn{1}{r|}{0.58} & 53.14 & 0.33 & 99.82 & 0.17 & 49.99 & 54.09 & 0.73 & 0.67 \\
\multicolumn{1}{r|}{LNP} & {51.11} & 2.98 & 98.16 & 1.56 & 49.86 & 48.8 & 3.57 & \multicolumn{1}{r|}{2.54} & 57.46 & 3.49 & 99.33 & 1.82 & 50.57 & 57.47 & 4.12 & 2.88 \\
\multicolumn{1}{r|}{NPR} & {46.18} & 5.57 & 94.44 & 3.29 & 48.86 & 42.77 & 5.91 & \multicolumn{1}{r|}{4.36} & 62.64 & 14.02 & 97.89 & 8.17 & 53.03 & 64.35 & 14.29 & 10.19 \\
\multicolumn{1}{r|}{PoundNet} & {87.29} & 57.34 & 96.69 & 49.78 & 73.23 & 87.87 & 56.49 & \multicolumn{1}{r|}{51.68} & 79.26 & 29.89 & 99.07 & 21.07 & 60.07 & 80.25 & 30.38 & 24.68 \\
\multicolumn{1}{r|}{SPrompts} & {71.21} & 10.13 & 99.78 & 5.78 & 52.78 & 69.23 & 10.64 & \multicolumn{1}{r|}{7.55} & 67.45 & 4.07 & 99.92 & 2.13 & 51.02 & 66.58 & 4.75 & 3.3 \\
\multicolumn{1}{r|}{UnivFD} & {87.28} & 54.03 & 99.06 & 45.59 & 72.33 & 87.3 & 53.52 & \multicolumn{1}{r|}{48.39} & 76.44 & 29.29 & 99.01 & 20.31 & 59.66 & 77.44 & 30.29 & 24.63 \\ \hline
 &  &  &  &  &  &  &  &  &  &  &  &  &  &  &  &  \\ \hline
\multicolumn{9}{c|}{DiffusionForensics} & \multicolumn{8}{c}{Ojha} \\ \hline
\multicolumn{1}{r|}{Method} & {$\mathbf{AP}$} & $\mathbf{F1}$ & $\mathbf{ACC_r}$ & $\mathbf{ACC_f}$ & $\mathbf{ACC}$ & $\mathbf{AUC_{roc}}$ & $\mathbf{AUC_{f1}}$ & \multicolumn{1}{r|}{$\mathbf{AUC_{f2}}$} & $\mathbf{AP}$ & $\mathbf{F1}$ & $\mathbf{ACC_r}$ & $\mathbf{ACC_f}$ & $\mathbf{ACC}$ & $\mathbf{AUC_{roc}}$ & $\mathbf{AUC_{f1}}$ & $\mathbf{AUC_{f2}}$ \\ \hline
\multicolumn{1}{r|}{CNNDet} & {60.62} & 6.56 & 99.57 & 4.17 & 56.48 & 66.87 & 7.02 & \multicolumn{1}{r|}{5.38} & 64.42 & 6.2 & 99.06 & 3.25 & 51.16 & 67.08 & 7.13 & 4.9 \\
\multicolumn{1}{r|}{FreDect} & {47.4} & 36.35 & 62.78 & 38.73 & 49.61 & 49.36 & 33.58 & \multicolumn{1}{r|}{35.28} & 67.15 & 53.4 & 79.71 & 44.21 & 61.96 & 73.99 & 47.7 & 46.39 \\
\multicolumn{1}{r|}{Freqnet} & {44.45} & 1.1 & 99.39 & 0.61 & 55.63 & 46.32 & 1.61 & \multicolumn{1}{r|}{1.27} & 48.98 & 1.97 & 98.92 & 1.01 & 49.97 & 45.97 & 2.45 & 1.77 \\
\multicolumn{1}{r|}{Fusing} & {60.18} & 1.4 & 99.98 & 0.71 & 56.18 & 69.2 & 2.51 & \multicolumn{1}{r|}{1.83} & 73.08 & 8.52 & 99.35 & 4.51 & 51.93 & 73.66 & 10.45 & 7.22 \\
\multicolumn{1}{r|}{GramNet} & {69.87} & 12.69 & 93.64 & 11.14 & 54.27 & 72.85 & 12.79 & \multicolumn{1}{r|}{11.64} & 51.43 & 0.17 & 99.64 & 0.09 & 49.86 & 51.82 & 0.53 & 0.54 \\
\multicolumn{1}{r|}{LGrad} & {53.33} & 0.25 & 99.97 & 0.13 & 55.87 & 59.88 & 0.67 & \multicolumn{1}{r|}{0.62} & 39.59 & 0.25 & 99.91 & 0.12 & 50.02 & 30.92 & 0.57 & 0.57 \\
\multicolumn{1}{r|}{LNP} & {61.81} & 6.59 & 99.71 & 3.61 & 57.16 & 67.44 & 7.29 & \multicolumn{1}{r|}{5.11} & 40.96 & 3.19 & 97 & 1.68 & 49.34 & 34.74 & 3.63 & 2.59 \\
\multicolumn{1}{r|}{NPR} & {54.04} & 5.27 & 92.65 & 4.36 & 52.46 & 55.16 & 5.59 & \multicolumn{1}{r|}{4.82} & 45.42 & 3.29 & 99.09 & 1.69 & 50.39 & 38.29 & 3.65 & 2.53 \\
\multicolumn{1}{r|}{PoundNet} & {71.88} & 30.06 & 97.84 & 22.56 & 65.18 & 82.15 & 30.41 & \multicolumn{1}{r|}{25.98} & 92.16 & 66.34 & 98.04 & 52.24 & 75.14 & 91.8 & 63.94 & 55.67 \\
\multicolumn{1}{r|}{SPrompts} & {69.58} & 2.68 & 99.67 & 1.47 & 56.12 & 73.62 & 3.66 & \multicolumn{1}{r|}{2.86} & 60.53 & 4.64 & 99.98 & 2.46 & 51.22 & 58.34 & 5.11 & 3.56 \\
\multicolumn{1}{r|}{UnivFD} & {63.96} & 19.37 & 98.17 & 13.54 & 61.31 & 67.53 & 20.5 & \multicolumn{1}{r|}{16.89} & 92.71 & 65.86 & 98.36 & 51.6 & 74.98 & 92.22 & 63.9 & 55.8 \\ \hline
 &  &  &  &  &  &  &  &  &  &  &  &  &  &  &  &  \\ \cline{1-10}
\multicolumn{2}{c|}{AntiFake} & \multicolumn{8}{c}{Average} &  &  &  &  &  &  &  \\ \cline{1-10}
\multicolumn{1}{r|}{Method} & \multicolumn{1}{r|}{$\mathbf{ACC}$} & $\mathbf{AP}$ & $\mathbf{F1}$ & $\mathbf{ACC_r}$ & $\mathbf{ACC_f}$ & $\mathbf{ACC}$ & $\mathbf{AUC_{roc}}$ & $\mathbf{AUC_{f1}}$ & $\mathbf{AUC_{f2}}$ &  &  &  &  &  &  &  \\ \cline{1-10}
\multicolumn{1}{r|}{CNNDet} & \multicolumn{1}{r|}{15.21} & 72.66 & 22.74 & 98.95 & 16.20 & 52.18 & 74.74 & 23.12 & 18.81 &  &  &  &  &  &  &  \\
\multicolumn{1}{r|}{FreDect} & \multicolumn{1}{r|}{26.47} & 66.10 & 45.71 & 79.79 & 41.01 & 55.39 & 67.87 & 41.97 & 40.64 &  &  &  &  &  &  &  \\
\multicolumn{1}{r|}{Freqnet} & \multicolumn{1}{r|}{14.87} & 48.96 & 10.11 & 88.31 & 12.21 & 46.01 & 48.10 & 10.46 & 11.34 &  &  &  &  &  &  &  \\
\multicolumn{1}{r|}{Fusing} & \multicolumn{1}{r|}{15.56} & 77.60 & 24.60 & \textbf{99.26} & 18.44 & 53.50 & 79.58 & 25.10 & 21.09 &  &  &  &  &  &  &  \\
\multicolumn{1}{r|}{GramNet} & \multicolumn{1}{r|}{22.3} & 57.02 & 7.05 & 94.11 & 6.49 & 46.57 & 57.97 & 7.31 & 6.82 &  &  &  &  &  &  &  \\
\multicolumn{1}{r|}{LGrad} & \multicolumn{1}{r|}{12.78} & 49.56 & 0.47 & {\ul 99.17} & 0.34 & 45.31 & 49.31 & 0.85 & 0.83 &  & {\ul } &  &  &  &  &  \\
\multicolumn{1}{r|}{LNP} & \multicolumn{1}{r|}{14.62} & 54.75 & 5.44 & 97.72 & 3.46 & 46.24 & 54.68 & 6.13 & 4.69 &  &  &  &  &  &  &  \\
\multicolumn{1}{r|}{NPR} & \multicolumn{1}{r|}{22.3} & 50.58 & 13.86 & 84.45 & 15.86 & 46.74 & 48.67 & 14.15 & 14.76 &  &  &  &  &  &  &  \\
\multicolumn{1}{r|}{PoundNet} & \multicolumn{1}{r|}{30.72} & \textbf{85.73} & \textbf{54.80} & 96.33 & \textbf{46.29} & \textbf{66.22} & \textbf{87.59} & \textbf{53.61} & \textbf{48.49} & \textbf{} &  & \textbf{} & \textbf{} & \textbf{} & \textbf{} & \textbf{} \\
\multicolumn{1}{r|}{SPrompts} & \multicolumn{1}{r|}{17.07} & 71.58 & 6.86 & 99.14 & 4.52 & 47.66 & 70.85 & 7.50 & 5.80 &  & \textbf{} &  &  &  &  &  \\
\multicolumn{1}{r|}{UnivFD} & \multicolumn{1}{r|}{37.55} & 83.70 & {\ul 50.51} & 97.97 & {\ul 41.09} & {\ul 65.74} & {\ul 84.13} & {\ul 49.61} & {\ul \textbf{44.14}} & {\ul } &  & {\ul } & {\ul } & \textbf{} & {\ul } & {\ul } \\ \cline{1-10}
\end{tabular}
}
\end{table}

\begin{table}[]
\centering
\caption{Comprehensive comparisons of our method and other AI-generated images detectors on $7$ datasets with \zhiwu{unified JPEG Compression $70\%$ on all the real and fake images}. $\mathbf{ACC_r}$, $\mathbf{ACC_f}$ represent detection accurracies for real images and fake images, respectively. For each test subset, the best results are highlighted in boldface and the second best results are underlined.
}
\label{tab:7datasets_70jpg}
\small
\resizebox{0.65\textheight}{!}{

\begin{tabular}{rrrrrrrrrrrrrrrrr}
\hline
\multicolumn{9}{c|}{ForenSynths} & \multicolumn{8}{c}{GANGen-Detection} \\ \hline
\multicolumn{1}{r|}{Method} & {$\mathbf{AP}$} & $\mathbf{F1}$ & $\mathbf{ACC_r}$ & $\mathbf{ACC_f}$ & $\mathbf{ACC}$ & $\mathbf{AUC_{roc}}$ & $\mathbf{AUC_{f1}}$ & \multicolumn{1}{r|}{$\mathbf{AUC_{f2}}$} & $\mathbf{AP}$ & $\mathbf{F1}$ & $\mathbf{ACC_r}$ & $\mathbf{ACC_f}$ & $\mathbf{ACC}$ & $\mathbf{AUC_{roc}}$ & $\mathbf{AUC_{f1}}$ & $\mathbf{AUC_{f2}}$ \\ \hline
\multicolumn{1}{r|}{CNNDet} & {86.71} & 45.56 & 98.25 & 34.13 & 66.19 & 87.29 & 45.5 & \multicolumn{1}{r|}{38.18} & 75.78 & 33.18 & 97.95 & 21.61 & 59.78 & 75.51 & 33.48 & 26.19 \\
\multicolumn{1}{r|}{FreDect} & {67} & 54.44 & 72.52 & 50.98 & 61.75 & 66.2 & 46.42 & \multicolumn{1}{r|}{46.32} & 59.99 & 20.51 & 90.18 & 20.9 & 55.54 & 60.33 & 23.12 & 23.33 \\
\multicolumn{1}{r|}{Freqnet} & {48.48} & 2.08 & 98.73 & 1.17 & 49.96 & 47.08 & 2.47 & \multicolumn{1}{r|}{1.88} & 48.3 & 45.09 & 41.12 & 60.22 & 50.67 & 48.44 & 44.89 & 52.67 \\
\multicolumn{1}{r|}{Fusing} & {88.46} & 37.05 & 99.3 & 28.67 & 63.99 & 88.55 & 37.77 & \multicolumn{1}{r|}{32.32} & 86.76 & 57.19 & 96.52 & 43.73 & 70.12 & 86.97 & 54.81 & 47.48 \\
\multicolumn{1}{r|}{GramNet} & {54.07} & 9.82 & 89.61 & 11.15 & 50.39 & 55.72 & 10.06 & \multicolumn{1}{r|}{10.69} & 50.72 & 0.39 & 99.69 & 0.2 & 49.94 & 51.19 & 0.73 & 0.67 \\
\multicolumn{1}{r|}{LGrad} & {52.95} & 1.33 & 95.83 & 1.02 & 48.43 & 53.25 & 1.8 & \multicolumn{1}{r|}{1.66} & 49.56 & 0.02 & 99.99 & 0.01 & 50 & 49.66 & 0.37 & 0.44 \\
\multicolumn{1}{r|}{LNP} & {54.28} & 5.05 & 96.56 & 4.59 & 50.58 & 55.16 & 5.63 & \multicolumn{1}{r|}{5.41} & 53.5 & 8.24 & 95.16 & 4.69 & 49.92 & 54.45 & 9.8 & 7.3 \\
\multicolumn{1}{r|}{NPR} & {45.81} & 5.29 & 95.76 & 3.69 & 49.73 & 42.19 & 5.44 & \multicolumn{1}{r|}{4.48} & 48.29 & 46.8 & 30.01 & 70.03 & 50.02 & 47.51 & 46.9 & 58.51 \\
\multicolumn{1}{r|}{PoundNet} & {91.54} & 70.19 & 90.31 & 65.98 & 78.15 & 91.12 & 68.3 & \multicolumn{1}{r|}{65.77} & 87.56 & 72.12 & 91.98 & 63.48 & 77.73 & 87.74 & 68.64 & 64.02 \\
\multicolumn{1}{r|}{SPrompts} & {78.17} & 10.59 & 95.91 & 9.64 & 52.78 & 76.57 & 11.38 & \multicolumn{1}{r|}{10.35} & 71.88 & 0.6 & 100 & 0.3 & 50.15 & 70.73 & 1.03 & 0.86 \\
\multicolumn{1}{r|}{UnivFD} & {89.86} & 58.02 & 98 & 47.38 & 72.69 & 89.08 & 57.1 & \multicolumn{1}{r|}{50.84} & 88.99 & 73.89 & 93.02 & 65.23 & 79.13 & 87.97 & 69.69 & 65.61 \\ \hline
 &  &  &  &  &  &  &  &  &  &  &  &  &  &  &  &  \\ \hline
\multicolumn{9}{c|}{DIF} & \multicolumn{8}{c}{GenImage} \\ \hline
\multicolumn{1}{r|}{Method} & {$\mathbf{AP}$} & $\mathbf{F1}$ & $\mathbf{ACC_r}$ & $\mathbf{ACC_f}$ & $\mathbf{ACC}$ & $\mathbf{AUC_{roc}}$ & $\mathbf{AUC_{f1}}$ & \multicolumn{1}{r|}{$\mathbf{AUC_{f2}}$} & $\mathbf{AP}$ & $\mathbf{F1}$ & $\mathbf{ACC_r}$ & $\mathbf{ACC_f}$ & $\mathbf{ACC}$ & $\mathbf{AUC_{roc}}$ & $\mathbf{AUC_{f1}}$ & $\mathbf{AUC_{f2}}$ \\ \hline
\multicolumn{1}{r|}{CNNDet} & {76.93} & 31.15 & 98.98 & 24.16 & 61.57 & 77.63 & 31.35 & \multicolumn{1}{r|}{26.77} & 68.5 & 6.78 & 99.55 & 3.88 & 51.72 & 70.84 & 7.69 & 5.56 \\
\multicolumn{1}{r|}{FreDect} & {74.58} & 54.59 & 86.2 & 44.75 & 65.48 & 76.54 & 48.86 & \multicolumn{1}{r|}{45.52} & 77.3 & 54.08 & 89.67 & 42.7 & 66.19 & 78.19 & 50.01 & 45.28 \\
\multicolumn{1}{r|}{Freqnet} & {48.15} & 1.12 & 99.59 & 0.58 & 50.09 & 44.7 & 1.52 & \multicolumn{1}{r|}{1.18} & 50.64 & 1.11 & 99.54 & 0.57 & 50.05 & 46.3 & 1.5 & 1.16 \\
\multicolumn{1}{r|}{Fusing} & {81.94} & 34.17 & 99.67 & 27.04 & 63.35 & 81.8 & 34.33 & \multicolumn{1}{r|}{29.72} & 71.41 & 6.8 & 99.82 & 3.91 & 51.86 & 73.27 & 8.18 & 6 \\
\multicolumn{1}{r|}{GramNet} & {44.51} & 0.51 & 98.08 & 0.27 & 49.17 & 40.21 & 0.84 & \multicolumn{1}{r|}{0.75} & 70.42 & 5.28 & 99.26 & 3.06 & 51.16 & 77.77 & 5.56 & 4.06 \\
\multicolumn{1}{r|}{LGrad} & {50.51} & 0.3 & 99.93 & 0.15 & 50.04 & 49.76 & 0.68 & \multicolumn{1}{r|}{0.64} & 56.46 & 0.21 & 99.88 & 0.1 & 49.99 & 58.33 & 0.63 & 0.6 \\
\multicolumn{1}{r|}{LNP} & {48.32} & 1.03 & 99.15 & 0.52 & 49.83 & 46.57 & 1.49 & \multicolumn{1}{r|}{1.16} & 56.27 & 1.48 & 99.69 & 0.75 & 50.22 & 56.89 & 1.99 & 1.47 \\
\multicolumn{1}{r|}{NPR} & {47.96} & 4.27 & 97.36 & 2.37 & 49.87 & 46.2 & 4.61 & \multicolumn{1}{r|}{3.31} & 66.6 & 9.71 & 98.56 & 5.42 & 51.99 & 69.96 & 10.07 & 7.03 \\
\multicolumn{1}{r|}{PoundNet} & {86.54} & 56.42 & 96.59 & 49.11 & 72.85 & 87 & 55.43 & \multicolumn{1}{r|}{50.85} & 76.37 & 28.03 & 99.07 & 19.99 & 59.53 & 77.37 & 28.46 & 23.32 \\
\multicolumn{1}{r|}{SPrompts} & {67.77} & 5.19 & 99.84 & 2.76 & 51.3 & 65.57 & 5.78 & \multicolumn{1}{r|}{4.01} & 61.46 & 1.89 & 99.95 & 0.96 & 50.46 & 59.27 & 2.4 & 1.74 \\
\multicolumn{1}{r|}{UnivFD} & {86.6} & 53.63 & 98.53 & 44.86 & 71.7 & 86.6 & 53.02 & \multicolumn{1}{r|}{47.77} & 74.77 & 29.37 & 98.6 & 20.23 & 59.41 & 75.7 & 30.48 & 24.8 \\ \hline
 &  &  &  &  &  &  &  &  &  &  &  &  &  &  &  &  \\ \hline
\multicolumn{9}{c|}{DiffusionForensics} & \multicolumn{8}{c}{Ojha} \\ \hline
\multicolumn{1}{r|}{Method} & {$\mathbf{AP}$} & $\mathbf{F1}$ & $\mathbf{ACC_r}$ & $\mathbf{ACC_f}$ & $\mathbf{ACC}$ & $\mathbf{AUC_{roc}}$ & $\mathbf{AUC_{f1}}$ & \multicolumn{1}{r|}{$\mathbf{AUC_{f2}}$} & $\mathbf{AP}$ & $\mathbf{F1}$ & $\mathbf{ACC_r}$ & $\mathbf{ACC_f}$ & $\mathbf{ACC}$ & $\mathbf{AUC_{roc}}$ & $\mathbf{AUC_{f1}}$ & $\mathbf{AUC_{f2}}$ \\ \hline
\multicolumn{1}{r|}{CNNDet} & {56.23} & 3.36 & 99.62 & 1.8 & 56.21 & 61.76 & 4.17 & \multicolumn{1}{r|}{2.99} & 64.46 & 6.7 & 99.1 & 3.53 & 51.31 & 67.12 & 8.02 & 5.53 \\
\multicolumn{1}{r|}{FreDect} & {57.35} & 41.16 & 74.64 & 41.24 & 57.19 & 60.51 & 37.82 & \multicolumn{1}{r|}{38.79} & 67.35 & 50.7 & 83.56 & 39.94 & 61.75 & 74.52 & 44.93 & 43.63 \\
\multicolumn{1}{r|}{Freqnet} & {41.09} & 0.24 & 99.96 & 0.12 & 55.87 & 40.98 & 0.58 & \multicolumn{1}{r|}{0.56} & 46.77 & 1.35 & 98.92 & 0.69 & 49.81 & 41.82 & 1.72 & 1.3 \\
\multicolumn{1}{r|}{Fusing} & {62.91} & 1.93 & 99.99 & 0.99 & 56.32 & 71.75 & 3.31 & \multicolumn{1}{r|}{2.37} & 72.18 & 10.25 & 99.36 & 5.47 & 52.42 & 72.86 & 12.46 & 8.68 \\
\multicolumn{1}{r|}{GramNet} & {65.95} & 3.49 & 97.97 & 2.19 & 54.89 & 67.93 & 3.81 & \multicolumn{1}{r|}{2.98} & 49.26 & 0.12 & 99.81 & 0.06 & 49.94 & 49.93 & 0.46 & 0.5 \\
\multicolumn{1}{r|}{LGrad} & {55.18} & 0.03 & 99.91 & 0.01 & 55.8 & 63.39 & 0.36 & \multicolumn{1}{r|}{0.42} & 42.26 & 0.27 & 99.92 & 0.14 & 50.03 & 34.44 & 0.66 & 0.62 \\
\multicolumn{1}{r|}{LNP} & {54.1} & 1.11 & 99.8 & 0.57 & 55.98 & 59.38 & 1.57 & \multicolumn{1}{r|}{1.2} & 38.82 & 1.41 & 98 & 0.73 & 49.36 & 30.59 & 1.85 & 1.4 \\
\multicolumn{1}{r|}{NPR} & {50.58} & 3.82 & 97.34 & 2.53 & 54.94 & 52.81 & 4.12 & \multicolumn{1}{r|}{3.28} & 44.93 & 2.43 & 99.34 & 1.24 & 50.29 & 38.07 & 2.81 & 1.99 \\
\multicolumn{1}{r|}{PoundNet} & {58.97} & 28.82 & 94.61 & 23.31 & 63.8 & 64.21 & 28.72 & \multicolumn{1}{r|}{25.55} & 91.57 & 66.67 & 98.24 & 52.24 & 75.24 & 91.23 & 63.89 & 55.64 \\
\multicolumn{1}{r|}{SPrompts} & {68.02} & 1.58 & 99.88 & 0.84 & 56.1 & 72.5 & 2.27 & \multicolumn{1}{r|}{1.74} & 55.57 & 2.53 & 99.98 & 1.3 & 50.64 & 52 & 3.08 & 2.18 \\
\multicolumn{1}{r|}{UnivFD} & {64.4} & 19.71 & 97.23 & 13.81 & 60.87 & 68.17 & 21.12 & \multicolumn{1}{r|}{17.44} & 91.52 & 67.01 & 97.26 & 53.41 & 75.34 & 91.15 & 64.55 & 56.89 \\ \hline
 &  &  &  &  &  &  &  &  &  &  &  &  &  &  &  &  \\ \cline{1-10}
\multicolumn{2}{c|}{AntiFake} & \multicolumn{8}{c}{Average} &  &  &  &  &  &  &  \\ \cline{1-10}
\multicolumn{1}{c|}{Method} & \multicolumn{1}{c|}{$\mathbf{ACC}$} & $\mathbf{AP}$ & $\mathbf{F1}$ & $\mathbf{ACC_r}$ & $\mathbf{ACC_f}$ & $\mathbf{ACC}$ & $\mathbf{AUC_{roc}}$ & $\mathbf{AUC_{f1}}$ & $\mathbf{AUC_{f2}}$ &  &  &  &  &  &  &  \\ \cline{1-10}
\multicolumn{1}{l|}{CNNDet} & \multicolumn{1}{l|}{14.79} & 71.44 & 21.12 & 98.91 & 14.85 & 51.65 & 73.36 & 21.70 & 17.54 &  &  &  &  &  &  &  \\
\multicolumn{1}{l|}{FreDect} & \multicolumn{1}{l|}{28.33} & 67.26 & 45.91 & 82.80 & 40.09 & 56.60 & 69.38 & 41.86 & 40.48 &  &  &  &  &  &  &  \\
\multicolumn{1}{l|}{Freqnet} & \multicolumn{1}{l|}{14.15} & 47.24 & 8.50 & 89.64 & 10.56 & 45.80 & 44.89 & 8.78 & 9.79 &  &  &  &  &  &  &  \\
\multicolumn{1}{l|}{Fusing} & \multicolumn{1}{l|}{15.21} & 77.28 & 24.57 & 99.11 & 18.30 & 53.32 & 79.20 & 25.14 & 21.10 &  &  &  &  &  &  &  \\
\multicolumn{1}{l|}{GramNet} & \multicolumn{1}{l|}{15.48} & 55.82 & 3.27 & 97.40 & 2.82 & 45.85 & 57.13 & 3.58 & 3.28 &  &  &  &  &  &  &  \\
\multicolumn{1}{l|}{LGrad} & \multicolumn{1}{l|}{12.52} & 51.15 & 0.36 & {\ul 99.24} & 0.24 & 45.26 & 51.47 & 0.75 & 0.73 &  & {\ul } &  &  &  &  &  \\
\multicolumn{1}{l|}{LNP} & \multicolumn{1}{l|}{13.29} & 50.88 & 3.05 & 98.06 & 1.98 & 45.60 & 50.51 & 3.72 & 2.99 &  &  &  &  &  &  &  \\
\multicolumn{1}{l|}{NPR} & \multicolumn{1}{l|}{20.64} & 50.70 & 12.05 & 86.40 & 14.21 & 46.78 & 49.46 & 12.33 & 13.10 &  &  &  &  &  &  &  \\
\multicolumn{1}{l|}{PoundNet} & \multicolumn{1}{l|}{32.34} & {\ul 82.09} & \textbf{53.71} & 95.13 & \textbf{45.69} & \textbf{65.66} & \textbf{83.11} & \textbf{52.24} & \textbf{47.53} & \textbf{} &  & \textbf{} & \textbf{} & \textbf{} & \textbf{} & \textbf{} \\
\multicolumn{1}{l|}{SPrompts} & \multicolumn{1}{l|}{15.19} & 67.15 & 3.73 & \textbf{99.26} & 2.63 & 46.66 & 66.11 & 4.32 & 3.48 &  & \textbf{} &  &  &  &  &  \\
\multicolumn{1}{l|}{UnivFD} & \multicolumn{1}{l|}{38.54} & \textbf{82.69} & {\ul 50.27} & 97.11 & {\ul 40.82} & {\ul 65.38} & \textbf{83.11} & {\ul 49.33} & {\ul 43.89} & {\ul } &  & {\ul } & {\ul } & \textbf{} & {\ul } & {\ul } \\ \cline{1-10}
\end{tabular}
}
\end{table}

\section{Main Results of 10 Benchmarks}
\label{sec:mainresults_20bench}

In the main paper, we focus on presenting the total experimental results obtained from $10$ testing datasets. 
In order to provide a more comprehensive view of our experimental results, we present the complete set of results here. 
Tables (from Table 12 to Table 30) give full results of all subsets. Figures (from Fig.4 to Fig.15) provide curves of all main results.


\newpage

\begin{table}[]
\small
\label{tab:dif}
\caption{Comprehensive comparisons of our method and other ai-generated images detectors on DIF dataset. \zhiwu{$\mathbf{ACC_r}$, $\mathbf{ACC_f}$ represent detection accurracies for real images and fake images, respectively.} For each test subset, the best results are highlighted in boldface and the second best results are underlined. }
\begin{tabular}{c|l|rrrrrrrr}
\hline
& \multicolumn{1}{c|}{\textbf{Method}} & \multicolumn{1}{c}{$\mathbf{AP}$} & \multicolumn{1}{c}{$\mathbf{F1}$} & \multicolumn{1}{c}{$\mathbf{ACC_r}$} & \multicolumn{1}{c}{$\mathbf{ACC_f}$} & \multicolumn{1}{c}{$\mathbf{ACC}$} & \multicolumn{1}{c}{$\mathbf{AUC_{roc}}$} & \multicolumn{1}{c}{$\mathbf{AUC_{f1}}$} & \multicolumn{1}{c}{$\mathbf{AUC_{f2}}$} \\ \hline

 \multirow{11}{*}{\rotatebox{90}{BigGAN}} & CNNDet\cite{wang2020cnn} & 88.30 & 34.18 & 98.92 & 20.83 & 59.88 & 89.49 & 34.78 & 25.73 \\
 & FreDect\cite{frank2020leveraging} & 77.47 & 68.88 & 63.51 & {\ul 71.71} & 67.61 & 73.73 & 64.26 & {\ul 65.61} \\
 & GramNet\cite{liu2020global} & 48.66 & 0.93 & 99.46 & 0.47 & 49.97 & 49.22 & 1.30 & 1.03 \\
 & Fusing\cite{ju2022fusing} & 89.81 & 32.23 & 99.60 & 19.29 & 59.44 & 90.36 & 32.43 & 23.87 \\
 & LNP\cite{liu2022detecting} & 52.26 & 4.00 & 97.98 & 2.08 & 50.03 & 53.21 & 4.66 & 3.23 \\
 & SPrompts\cite{wang2022s} & 92.82 & 50.60 & {\ul 99.80} & 33.94 & 66.87 & 91.88 & 49.65 & 38.79 \\
 & UnivFD\cite{ojha2023towards} & {\ul 95.62} & {\ul 74.10} & 98.72 & 59.61 & {\ul 79.17} & {\ul 95.18} & {\ul 73.06} & 64.60 \\
 & LGrad\cite{tan2023learning} & 51.26 & 0.40 & \textbf{100.00} & 0.20 & 50.10 & 52.41 & 0.89 & 0.77 \\
 & NPR\cite{tan2023rethinking} & 52.05 & 11.29 & 96.64 & 6.18 & 51.41 & 50.48 & 11.32 & 7.78 \\
 & Freqnet\cite{tan2024frequencyaware} & 50.13 & 3.99 & 97.65 & 2.08 & 49.87 & 48.28 & 4.64 & 3.21 \\
  \cline{2-10} 
 & PoundNet & \textbf{97.59} & \textbf{91.78} & 89.45 & \textbf{93.75} & \textbf{91.60} & \textbf{97.46} & \textbf{89.79} & \textbf{91.09} \\
 \hline

 \multirow{11}{*}{\rotatebox{90}{CycleGAN}} & CNNDet\cite{wang2020cnn} & 96.64 & 77.18 & 98.89 & 63.54 & 81.21 & 96.44 & 74.85 & 66.36 \\
 & FreDect\cite{frank2020leveraging} & 80.97 & 64.58 & 85.17 & 54.76 & 69.96 & 77.82 & 62.67 & 58.59 \\
 & GramNet\cite{liu2020global} & 63.88 & 3.61 & 99.26 & 1.85 & 50.56 & 66.16 & 3.95 & 2.72 \\
 & Fusing\cite{ju2022fusing} & 96.18 & 68.02 & {\ul 99.51} & 51.79 & 75.65 & 96.37 & 65.38 & 55.73 \\
 & LNP\cite{liu2022detecting} & 87.40 & 28.06 & 99.26 & 16.44 & 57.85 & 87.57 & 29.36 & 21.22 \\
 & SPrompts\cite{wang2022s} & 94.80 & 73.09 & 99.13 & 58.10 & 78.62 & 93.88 & 72.61 & 63.30 \\
 & UnivFD\cite{ojha2023towards} & \textbf{98.81} & \textbf{91.13} & 99.13 & {\ul 84.43} & \textbf{91.78} & \textbf{98.71} & \textbf{88.91} & {\ul 85.15} \\
 & LGrad\cite{tan2023learning} & 47.82 & 0.00 & \textbf{100.00} & 0.00 & 50.00 & 48.29 & 0.63 & 0.60 \\
 & NPR\cite{tan2023rethinking} & 55.41 & 17.43 & 93.82 & 10.14 & 51.98 & 53.93 & 17.20 & 12.20 \\
 & Freqnet\cite{tan2024frequencyaware} & 69.36 & 9.26 & 98.15 & 4.94 & 51.55 & 72.59 & 10.53 & 7.18 \\
  \cline{2-10} 
 & PoundNet & {\ul 98.26} & {\ul 89.67} & 77.87 & \textbf{99.26} & {\ul 88.57} & {\ul 98.29} & {\ul 88.47} & \textbf{93.76} \\
 \hline

\multirow{11}{*}{\rotatebox{90}{GauGAN}} & CNNDet\cite{wang2020cnn} & 96.18 & 52.00 & 99.73 & 35.23 & 67.48 & 96.40 & 51.63 & 40.79 \\
 & FreDect\cite{frank2020leveraging} & 70.23 & 68.27 & 66.09 & 69.41 & 67.75 & 73.53 & 60.94 & 62.09 \\
 & GramNet\cite{liu2020global} & 53.72 & 0.40 & 99.64 & 0.20 & 49.92 & 55.38 & 0.75 & 0.68 \\
 & Fusing\cite{ju2022fusing} & 97.03 & 49.66 & {\ul 99.75} & 33.11 & 66.43 & 97.14 & 49.16 & 38.70 \\
 & LNP\cite{liu2022detecting} & 57.57 & 0.62 & 99.18 & 0.31 & 49.74 & 60.91 & 1.20 & 0.97 \\
 & SPrompts\cite{wang2022s} & 93.32 & 59.39 & 99.67 & 42.38 & 71.02 & 92.26 & 58.41 & 47.53 \\
 & UnivFD\cite{ojha2023towards} & \textbf{99.75} & \textbf{96.52} & 99.53 & {\ul 93.72} & \textbf{96.62} & \textbf{99.72} & \textbf{94.87} & {\ul 93.08} \\
 & LGrad\cite{tan2023learning} & 48.82 & 0.04 & \textbf{99.96} & 0.02 & 49.99 & 49.75 & 0.43 & 0.48 \\
 & NPR\cite{tan2023rethinking} & 44.32 & 1.39 & 98.20 & 0.71 & 49.45 & 41.24 & 1.69 & 1.29 \\
 & Freqnet\cite{tan2024frequencyaware} & 55.75 & 1.41 & 99.53 & 0.71 & 50.12 & 56.44 & 1.93 & 1.43 \\
  \cline{2-10} 
 & PoundNet & {\ul 99.60} & {\ul 95.67} & 92.38 & \textbf{98.69} & {\ul 95.53} & {\ul 99.57} & {\ul 94.12} & \textbf{96.25} \\
 \hline

 \multirow{11}{*}{\rotatebox{90}{ProGAN}} & CNNDet\cite{wang2020cnn} & \textbf{100.00} & \textbf{99.93} & 99.91 & \textbf{99.94} & \textbf{99.93} & \textbf{100.00} & \textbf{99.22} & \textbf{99.31} \\
 & FreDect\cite{frank2020leveraging} & 93.63 & 78.26 & 95.33 & 67.29 & 81.31 & 93.87 & 70.48 & 65.91 \\
 & GramNet\cite{liu2020global} & 61.99 & 0.40 & {\ul 99.97} & 0.20 & 50.09 & 64.06 & 0.72 & 0.66 \\
 & Fusing\cite{ju2022fusing} & \textbf{100.00} & {\ul 99.65} & 99.94 & {\ul 99.37} & {\ul 99.66} & \textbf{100.00} & {\ul 98.72} & {\ul 98.50} \\
 & LNP\cite{liu2022detecting} & 82.43 & 27.93 & 98.60 & 16.46 & 57.53 & 84.26 & 28.35 & 20.49 \\
 & SPrompts\cite{wang2022s} & 98.46 & 57.11 & 99.94 & 39.99 & 69.97 & 98.35 & 56.25 & 45.21 \\
 & UnivFD\cite{ojha2023towards} & 99.94 & 98.71 & 99.57 & 97.88 & 98.72 & 99.94 & 96.98 & 96.51 \\
 & LGrad\cite{tan2023learning} & 53.56 & 0.17 & \textbf{100.00} & 0.09 & 50.04 & 54.62 & 0.58 & 0.57 \\
 & NPR\cite{tan2023rethinking} & 49.97 & 0.80 & 99.57 & 0.40 & 49.99 & 50.46 & 1.12 & 0.91 \\
 & Freqnet\cite{tan2024frequencyaware} & 56.83 & 2.26 & 99.51 & 1.15 & 50.33 & 56.41 & 2.71 & 1.93 \\
  \cline{2-10}
 & PoundNet & 99.95 & 98.73 & 98.08 & {\ul 99.37} & 98.72 & 99.95 & 97.54 & 98.16 \\
 \hline

\end{tabular}
\end{table}

\begin{table}[]
\small
\label{tab:10benchmarks_6}
\caption{Comprehensive comparisons of our method and other ai-generated images detectors on DIF dataset. \zhiwu{$\mathbf{ACC_r}$, $\mathbf{ACC_f}$ represent detection accurracies for real images and fake images, respectively.} For each test subset, the best results are highlighted in boldface and the second best results are underlined. }
\begin{tabular}{c|l|rrrrrrrr}
\hline
& \multicolumn{1}{c|}{\textbf{Method}} & \multicolumn{1}{c}{$\mathbf{AP}$} & \multicolumn{1}{c}{$\mathbf{F1}$} & \multicolumn{1}{c}{$\mathbf{ACC_r}$} & \multicolumn{1}{c}{$\mathbf{ACC_f}$} & \multicolumn{1}{c}{$\mathbf{ACC}$} & \multicolumn{1}{c}{$\mathbf{AUC_{roc}}$} & \multicolumn{1}{c}{$\mathbf{AUC_{f1}}$} & \multicolumn{1}{c}{$\mathbf{AUC_{f2}}$} \\ \hline

\multirow{11}{*}{\rotatebox{90}{StarGAN}}  & CNNDet\cite{wang2020cnn} & 91.89 & 70.07 & 97.24 & 55.41 & 76.33 & 91.00 & 67.75 & 58.86 \\
 & FreDect\cite{frank2020leveraging} & 90.22 & 80.26 & 77.00 & 82.45 & 79.72 & 88.24 & 70.72 & 72.16 \\
 & GramNet\cite{liu2020global} & 68.00 & 0.40 & 99.87 & 0.20 & 50.03 & 69.43 & 0.84 & 0.73 \\
 & Fusing\cite{ju2022fusing} & {\ul 99.07} & {\ul 89.94} & 99.60 & 82.04 & {\ul 90.82} & {\ul 98.93} & {\ul 87.64} & 82.79 \\
 & LNP\cite{liu2022detecting} & 51.79 & 1.32 & 98.92 & 0.67 & 49.80 & 53.28 & 2.47 & 1.81 \\
 & SPrompts\cite{wang2022s} & \textbf{99.75} & 59.72 & \textbf{100.00} & 42.57 & 71.28 & \textbf{99.75} & 59.54 & 48.96 \\
 & UnivFD\cite{ojha2023towards} & 98.60 & \textbf{94.05} & 95.23 & {\ul 93.01} & \textbf{94.12} & 98.44 & \textbf{89.82} & {\ul 89.18} \\
 & LGrad\cite{tan2023learning} & 53.86 & 0.00 & {\ul 99.93} & 0.00 & 49.97 & 54.72 & 0.34 & 0.42 \\
 & NPR\cite{tan2023rethinking} & 42.12 & 0.00 & {\ul 99.93} & 0.00 & 49.97 & 37.41 & 0.35 & 0.43 \\
 & Freqnet\cite{tan2024frequencyaware} & 58.41 & 13.39 & 95.97 & 7.46 & 51.71 & 60.24 & 13.51 & 9.47 \\
  \cline{2-10} 
 & PoundNet & 97.53 & 79.25 & 47.75 & \textbf{99.93} & 73.84 & 97.70 & 79.26 & \textbf{90.09} \\
\hline
 
\multirow{11}{*}{\rotatebox{90}{StyleGAN}} & CNNDet\cite{wang2020cnn} & {\ul 97.73} & 53.67 & 99.93 & 36.70 & 68.32 & {\ul 97.76} & 53.39 & 42.36 \\
 & FreDect\cite{frank2020leveraging} & 78.43 & {\ul 65.78} & 85.64 & {\ul 56.05} & 70.84 & 71.28 & {\ul 59.81} & {\ul 54.78} \\
 & GramNet\cite{liu2020global} & 62.26 & 0.84 & 99.95 & 0.42 & 50.18 & 59.77 & 1.15 & 0.93 \\
 & Fusing\cite{ju2022fusing} & \textbf{98.11} & 54.53 & 99.89 & 37.53 & 68.71 & \textbf{98.12} & 53.66 & 43.00 \\
 & LNP\cite{liu2022detecting} & 73.91 & 13.96 & 99.29 & 7.56 & 53.42 & 74.07 & 14.69 & 10.05 \\
 & SPrompts\cite{wang2022s} & 94.54 & 10.74 & {\ul 99.98} & 5.68 & 52.83 & 94.12 & 12.12 & 8.23 \\
 & UnivFD\cite{ojha2023towards} & 94.73 & 60.64 & 99.76 & 43.62 & {\ul 71.69} & 94.15 & 59.73 & 49.62 \\
 & LGrad\cite{tan2023learning} & 60.37 & 0.15 & \textbf{100.00} & 0.07 & 50.04 & 63.17 & 0.59 & 0.58 \\
 & NPR\cite{tan2023rethinking} & 51.89 & 0.58 & 99.85 & 0.29 & 50.07 & 47.76 & 1.01 & 0.84 \\
 & Freqnet\cite{tan2024frequencyaware} & 55.17 & 1.51 & 99.32 & 0.77 & 50.05 & 55.83 & 2.09 & 1.53 \\
  \cline{2-10} 
 & PoundNet & 96.56 & \textbf{85.04} & 96.51 & \textbf{76.55} & \textbf{86.53} & 96.72 & \textbf{82.24} & \textbf{77.14} \\
\hline
 
\multirow{11}{*}{\rotatebox{90}{StyleGAN2}} & CNNDet\cite{wang2020cnn} & {\ul 96.81} & 41.81 & 99.91 & 26.46 & 63.18 & {\ul 96.89} & 41.99 & 31.74 \\
 & FreDect\cite{frank2020leveraging} & 78.17 & {\ul 59.71} & 89.99 & {\ul 46.82} & {\ul 68.41} & 75.53 & {\ul 53.68} & {\ul 48.20} \\
 & GramNet\cite{liu2020global} & 59.40 & 0.16 & {\ul 99.95} & 0.08 & 50.01 & 60.39 & 0.52 & 0.53 \\
 & Fusing\cite{ju2022fusing} & \textbf{97.30} & 29.90 & 99.93 & 17.59 & 58.76 & \textbf{97.51} & 31.18 & 23.05 \\
 & LNP\cite{liu2022detecting} & 73.41 & 7.00 & 99.26 & 3.65 & 51.46 & 75.65 & 7.97 & 5.41 \\
 & SPrompts\cite{wang2022s} & 87.71 & 4.47 & \textbf{99.96} & 2.29 & 51.12 & 88.30 & 5.48 & 3.73 \\
 & UnivFD\cite{ojha2023towards} & 93.41 & 39.60 & 99.61 & 24.79 & 62.20 & 93.49 & 42.05 & 33.09 \\
 & LGrad\cite{tan2023learning} & 62.54 & 0.61 & 99.93 & 0.31 & 50.12 & 65.12 & 1.12 & 0.91 \\
 & NPR\cite{tan2023rethinking} & 48.29 & 0.00 & 99.77 & 0.00 & 49.89 & 47.64 & 0.34 & 0.42 \\
 & Freqnet\cite{tan2024frequencyaware} & 62.61 & 6.84 & 99.21 & 3.57 & 51.39 & 62.36 & 7.30 & 4.93 \\
  \cline{2-10} 
 & PoundNet & 93.61 & \textbf{67.67} & 98.06 & \textbf{52.13} & \textbf{75.09} & 93.69 & \textbf{65.90} & \textbf{56.60} \\
\hline

\multirow{11}{*}{\rotatebox{90}{Dalle-2}} & CNNDet\cite{wang2020cnn} & 45.22 & 1.58 & 97.13 & 0.82 & 48.97 & 43.87 & 2.13 & 1.58 \\
 & FreDect\cite{frank2020leveraging} & 54.04 & 8.70 & 96.10 & 4.72 & 50.41 & 57.91 & 9.93 & 7.24 \\
 & GramNet\cite{liu2020global} & 37.94 & {\ul 19.74} & 55.65 & {\ul 15.81} & 35.73 & 31.80 & {\ul 19.95} & {\ul 17.51} \\
 & Fusing\cite{ju2022fusing} & 61.94 & 0.41 & \textbf{100.00} & 0.21 & 50.10 & 60.01 & 1.50 & 1.16 \\
 & LNP\cite{liu2022detecting} & 49.99 & 15.52 & 90.14 & 9.24 & 49.69 & 51.11 & 16.10 & 11.88 \\
 & SPrompts\cite{wang2022s} & 60.05 & 5.12 & 98.36 & 2.67 & 50.51 & 63.27 & 6.05 & 4.18 \\
 & UnivFD\cite{ojha2023towards} & {\ul 71.11} & 11.22 & {\ul 99.79} & 5.95 & {\ul 52.87} & {\ul 70.68} & 12.77 & 8.94 \\
 & LGrad\cite{tan2023learning} & 59.96 & 1.21 & 99.18 & 0.62 & 49.90 & 63.39 & 2.29 & 1.67 \\
 & NPR\cite{tan2023rethinking} & 40.76 & 11.71 & 77.82 & 7.60 & 42.71 & 36.13 & 11.99 & 9.25 \\
 & Freqnet\cite{tan2024frequencyaware} & 41.65 & 0.41 & 99.18 & 0.21 & 49.69 & 37.96 & 0.94 & 0.80 \\
  \cline{2-10} 
 & PoundNet & \textbf{73.29} & \textbf{30.23} & 95.07 & \textbf{18.69} & \textbf{56.88} & \textbf{77.12} & \textbf{30.73} & \textbf{23.48} \\
\hline
 
\multirow{11}{*}{\rotatebox{90}{Dalle Mini}} & CNNDet\cite{wang2020cnn} & 50.35 & 4.26 & 96.88 & 2.24 & 49.56 & 53.12 & 4.83 & 3.38 \\
 & FreDect\cite{frank2020leveraging} & 90.86 & {\ul 76.73} & 94.28 & {\ul 65.81} & {\ul 80.04} & 92.95 & {\ul 69.86} & {\ul 64.01} \\
 & GramNet\cite{liu2020global} & 32.44 & 1.50 & 52.12 & 1.12 & 26.62 & 11.78 & 1.83 & 1.66 \\
 & Fusing\cite{ju2022fusing} & 71.35 & 6.05 & \textbf{99.92} & 3.12 & 51.52 & 70.25 & 7.34 & 4.98 \\
 & LNP\cite{liu2022detecting} & 36.44 & 2.62 & 88.31 & 1.48 & 44.90 & 27.01 & 3.31 & 2.51 \\
 & SPrompts\cite{wang2022s} & 87.19 & 48.59 & 98.72 & 32.51 & 65.61 & 85.76 & 48.09 & 37.86 \\
 & UnivFD\cite{ojha2023towards} & {\ul 96.75} & 70.47 & {\ul 99.64} & 54.60 & 77.12 & {\ul 96.43} & 68.64 & 59.32 \\
 & LGrad\cite{tan2023learning} & 39.11 & 0.47 & 99.04 & 0.24 & 49.64 & 32.88 & 0.83 & 0.73 \\
 & NPR\cite{tan2023rethinking} & 37.24 & 7.41 & 80.26 & 4.60 & 42.43 & 29.03 & 7.57 & 5.74 \\
 & Freqnet\cite{tan2024frequencyaware} & 51.28 & 1.96 & 98.64 & 1.00 & 49.82 & 52.16 & 2.57 & 1.85 \\
  \cline{2-10} 
 & PoundNet & \textbf{97.72} & \textbf{91.33} & 95.88 & \textbf{87.51} & \textbf{91.69} & \textbf{97.84} & \textbf{89.11} & \textbf{86.66} \\
 \hline

\end{tabular}
\end{table}

\begin{table}[]
\small
\label{tab:10benchmarks_7}
\caption{Comprehensive comparisons of our method and other ai-generated images detectors on DIF dataset. \zhiwu{$\mathbf{ACC_r}$, $\mathbf{ACC_f}$ represent detection accurracies for real images and fake images, respectively.} For each test subset, the best results are highlighted in boldface and the second best results are underlined. }

\begin{tabular}{c|l|rrrrrrrr}
\hline
& \multicolumn{1}{c|}{\textbf{Method}} & \multicolumn{1}{c}{$\mathbf{AP}$} & \multicolumn{1}{c}{$\mathbf{F1}$} & \multicolumn{1}{c}{$\mathbf{ACC_r}$} & \multicolumn{1}{c}{$\mathbf{ACC_f}$} & \multicolumn{1}{c}{$\mathbf{ACC}$} & \multicolumn{1}{c}{$\mathbf{AUC_{roc}}$} & \multicolumn{1}{c}{$\mathbf{AUC_{f1}}$} & \multicolumn{1}{c}{$\mathbf{AUC_{f2}}$} \\ \hline

\multirow{11}{*}{\rotatebox{90}{Glide}} & CNNDet\cite{wang2020cnn} & 61.10 & \textbf{7.62} & 96.88 & 4.08 & \textbf{50.48} & 66.69 & \textbf{9.17} & {\ul 6.38} \\
 & FreDect\cite{frank2020leveraging} & 84.21 & 50.82 & 94.72 & 35.87 & 65.29 & 92.83 & 47.60 & 43.89 \\
 & GramNet\cite{liu2020global} & 33.20 & 2.84 & 52.52 & 2.12 & 27.32 & 15.37 & 3.11 & 2.73 \\
 & Fusing\cite{ju2022fusing} & 81.18 & 7.70 & \textbf{99.96} & 4.00 & 51.98 & 80.59 & 9.61 & 6.57 \\
 & LNP\cite{liu2022detecting} & 46.53 & 13.07 & 87.75 & 7.85 & 47.80 & 46.39 & 13.92 & 10.38 \\
 & SPrompts\cite{wang2022s} & 42.68 & 1.64 & 98.48 & 0.84 & 49.66 & 38.96 & 2.38 & 1.74 \\
 & UnivFD\cite{ojha2023towards} & {\ul 96.84} & {\ul 71.14} & {\ul 99.64} & {\ul 55.40} & {\ul 77.52} & {\ul 96.57} & {\ul 68.08} & {\ul 58.90} \\
 & LGrad\cite{tan2023learning} & 39.01 & 0.32 & 99.04 & 0.16 & 49.60 & 32.78 & 0.74 & 0.68 \\
 & NPR\cite{tan2023rethinking} & 34.14 & 4.65 & 80.58 & 2.84 & 41.71 & 18.88 & 4.91 & 3.74 \\
 & Freqnet\cite{tan2024frequencyaware} & 43.76 & 0.39 & 98.76 & 0.20 & 49.48 & 42.02 & 0.73 & 0.67 \\
  \cline{2-10} 
 & PoundNet & \textbf{97.86} & \textbf{91.21} & 95.68 & \textbf{87.47} & \textbf{91.57} & \textbf{98.13} & \textbf{88.81} & \textbf{86.42} \\
 \hline

\multirow{11}{*}{\rotatebox{90}{MJ}} & CNNDet\cite{wang2020cnn} & 46.08 & 1.47 & 97.00 & 0.76 & 48.88 & 46.87 & 2.31 & 1.71 \\
 & FreDect\cite{frank2020leveraging} & {\ul 87.88} & {\ul 70.77} & 94.44 & {\ul 57.81} & {\ul 76.12} & {\ul 90.08} & {\ul 64.26} & {\ul 57.56} \\
 & GramNet\cite{liu2020global} & 31.97 & 1.13 & 52.20 & 0.84 & 26.52 & 9.41 & 1.45 & 1.35 \\
 & Fusing\cite{ju2022fusing} & 67.20 & 2.22 & \textbf{99.92} & 1.12 & 50.52 & 66.06 & 3.46 & 2.43 \\
 & LNP\cite{liu2022detecting} & 37.73 & 6.58 & 88.23 & 3.80 & 46.02 & 27.75 & 7.23 & 5.31 \\
 & SPrompts\cite{wang2022s} & 63.74 & 8.55 & 98.68 & 4.52 & 51.60 & 64.05 & 10.25 & 7.08 \\
 & UnivFD\cite{ojha2023towards} & 83.64 & 19.32 & {\ul 99.68} & 10.73 & 55.20 & 83.61 & 22.21 & 16.21 \\
 & LGrad\cite{tan2023learning} & 43.24 & 1.10 & 99.00 & 0.56 & 49.78 & 39.49 & 1.63 & 1.24 \\
 & NPR\cite{tan2023rethinking} & 35.08 & 4.46 & 80.66 & 2.72 & 41.69 & 22.29 & 4.72 & 3.59 \\
 & Freqnet\cite{tan2024frequencyaware} & 44.07 & 0.86 & 98.60 & 0.44 & 49.52 & 42.03 & 1.32 & 1.05 \\
  \cline{2-10} 
 & PoundNet & \textbf{92.69} & \textbf{74.01} & 96.00 & \textbf{61.09} & \textbf{78.54} & \textbf{93.52} & \textbf{71.16} & \textbf{63.58} \\
 \hline
 
\multirow{11}{*}{\rotatebox{90}{SD-v1.4}} & CNNDet\cite{wang2020cnn} & 49.40 & 3.67 & 96.91 & 1.93 & 49.42 & 51.65 & 4.16 & 2.94 \\
 & FreDect\cite{frank2020leveraging} & 57.36 & 12.41 & 94.86 & 6.95 & 50.90 & 61.87 & 15.18 & 11.22 \\
 & GramNet\cite{liu2020global} & 48.84 & \textbf{50.04} & 52.21 & \textbf{49.32} & 50.76 & 51.69 & \textbf{49.97} & \textbf{49.57} \\
 & Fusing\cite{ju2022fusing} & 53.10 & 0.32 & \textbf{99.92} & 0.16 & 50.04 & 52.92 & 1.09 & 0.90 \\
 & LNP\cite{liu2022detecting} & 41.69 & 7.53 & 87.94 & 4.38 & 46.16 & 39.41 & 8.16 & 6.02 \\
 & SPrompts\cite{wang2022s} & {\ul 59.67} & 4.65 & 98.71 & 2.41 & 50.56 & {\ul 62.30} & 6.23 & 4.29 \\
 & UnivFD\cite{ojha2023towards} & \textbf{62.93} & 4.24 & {\ul 99.80} & 2.17 & {\ul 50.98} & \textbf{64.27} & 6.24 & 4.36 \\
 & LGrad\cite{tan2023learning} & 47.36 & 0.80 & 99.36 & 0.40 & 49.88 & 48.67 & 1.20 & 0.97 \\
 & NPR\cite{tan2023rethinking} & 52.03 & {\ul 27.44} & 80.47 & {\ul 19.01} & 49.74 & 52.08 & {\ul 27.73} & {\ul 22.12} \\
 & Freqnet\cite{tan2024frequencyaware} & 41.54 & 0.79 & 98.67 & 0.40 & 49.54 & 37.49 & 1.11 & 0.91 \\
  \cline{2-10} 
 & PoundNet & 57.47 & 11.91 & 95.94 & 6.59 & \textbf{51.27} & 59.87 & 13.09 & 9.33 \\
 \hline
 
\multirow{11}{*}{\rotatebox{90}{SD-v2.1}} & CNNDet\cite{wang2020cnn} & 46.31 & 3.05 & 96.10 & 1.61 & 48.85 & 46.97 & 3.68 & 2.62 \\
 & FreDect\cite{frank2020leveraging} & {\ul 54.39} & 10.82 & 93.89 & 6.07 & 49.98 & \textbf{58.30} & 13.40 & 9.90 \\
 & GramNet\cite{liu2020global} & 46.02 & \textbf{44.48} & 52.01 & \textbf{42.32} & 47.17 & 46.35 & \textbf{44.41} & \textbf{43.14} \\
 & Fusing\cite{ju2022fusing} & 49.08 & 0.40 & \textbf{99.88} & 0.20 & 50.04 & 47.23 & 0.93 & 0.79 \\
 & LNP\cite{liu2022detecting} & 42.37 & 7.83 & 88.55 & 4.54 & 46.54 & 40.22 & 8.57 & 6.31 \\
 & SPrompts\cite{wang2022s} & 52.48 & 2.58 & 98.67 & 1.33 & 50.00 & 54.41 & 3.78 & 2.66 \\
 & UnivFD\cite{ojha2023towards} & \textbf{55.96} & 1.90 & {\ul 99.64} & 0.96 & \textbf{50.30} & {\ul 57.33} & 3.63 & 2.59 \\
 & LGrad\cite{tan2023learning} & 47.77 & 0.87 & 99.32 & 0.44 & 49.88 & 48.59 & 1.34 & 1.06 \\
 & NPR\cite{tan2023rethinking} & 51.44 & {\ul 25.69} & 79.98 & {\ul 17.68} & 48.83 & 50.45 & {\ul 25.81} & {\ul 20.50} \\
 & Freqnet\cite{tan2024frequencyaware} & 41.71 & 1.03 & 98.95 & 0.52 & 49.74 & 36.75 & 1.45 & 1.12 \\
  \cline{2-10} 
 & PoundNet & 51.71 & 7.61 & 96.30 & 4.10 & {\ul 50.20} & 53.24 & 8.96 & 6.33 \\
 \hline

\multirow{11}{*}{\rotatebox{90}{Average}} & CNNDet & 74.31 & 34.65 & 98.11 & 26.89 & 62.50 & 75.17 & 34.61 & 29.52 \\
 & FreDect & 76.76 & 55.08 & 87.00 & 48.13 & 67.57 & 77.53 & 50.98 & 47.78 \\
 & Freqnet & 51.71 & 3.39 & 98.63 & 1.80 & 50.22 & 50.81 & 3.91 & 2.78 \\
 & Fusing & 81.64 & 33.92 & \textbf{99.83} & 26.89 & 63.36 & 81.19 & 34.01 & 29.42 \\
 & GramNet & 49.87 & 9.73 & 78.06 & 8.84 & 43.45 & 45.45 & 10.00 & 9.48 \\
 & LGrad & 50.36 & 0.47 & {\ul 99.60} & 0.24 & 49.92 & 50.30 & 0.97 & 0.82 \\
 & LNP & 56.42 & 10.46 & 94.11 & 6.04 & 50.07 & 55.45 & 11.23 & 8.12 \\
 & NPR & 45.75 & 8.68 & 89.81 & 5.55 & 47.68 & 41.37 & 8.90 & 6.83 \\
 & PoundNet & \textbf{88.76} & \textbf{70.32} & 90.38 & \textbf{68.09} & \textbf{79.23} & \textbf{89.47} & \textbf{69.17} & \textbf{67.61} \\
 & SPrompts & 79.02 & 29.71 & 99.24 & 20.71 & 59.97 & 79.02 & 30.07 & 24.12 \\
 & UnivFD & {\ul 88.31} & {\ul 56.39} & 99.21 & {\ul 48.22} & {\ul 73.72} & {\ul 88.35} & {\ul 55.92} & {\ul 50.89} \\
\hline

\end{tabular}
\end{table}

\begin{figure}[]
\centering
\centerline{\includegraphics[width=1\linewidth]{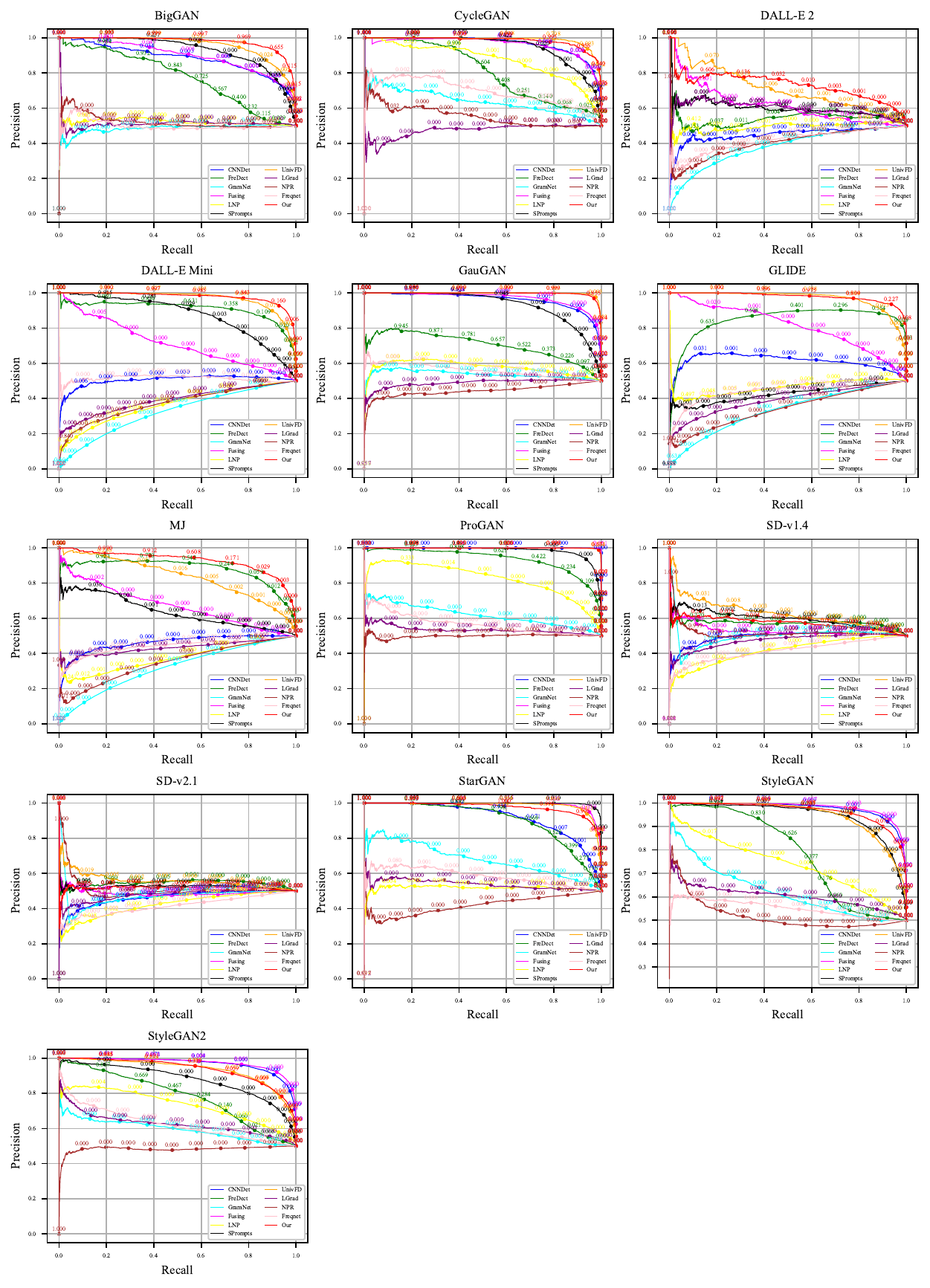}}
\caption{\zhiwu{Precision-Recall Curves of deepfake detection methods on the DIF dataset. The numbers on each curve represent the decision thresholds that define the boundary between positive and negative predictions. The numbers typically fall within a narrow range.}}
\label{fig:}
\end{figure}

\begin{figure}[]
\centering
\centerline{\includegraphics[width=1\linewidth]{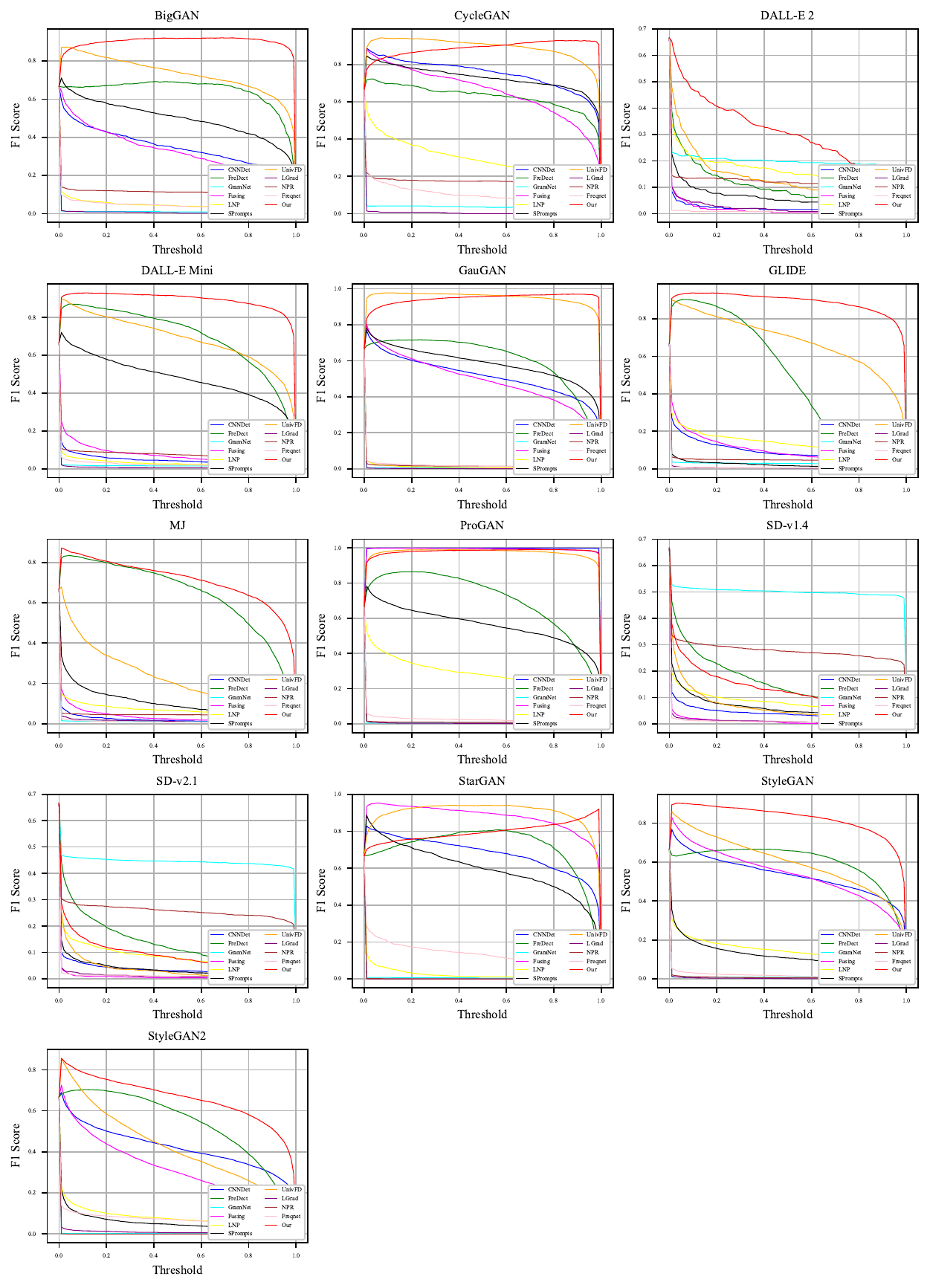}}
\caption{\zhiwu{F1 Curves with different thresholds of logits of deepfake detection methods on various deepfakes on the DIF dataset. The horizontal axis denotes the decision thresholds that determine the boundary between positive and negative predictions.}}
\label{fig:}
\end{figure}

\newpage

\begin{table}[]
\small
\label{tab:diffusion}
\caption{Comprehensive comparisons of our method and other ai-generated images detectors on ImageNet split of DiffusionForensics dataset. \zhiwu{$\mathbf{ACC_r}$, $\mathbf{ACC_f}$ represent detection accurracies for real images and fake images, respectively.} For each test subset, the best results are highlighted in boldface and the second best results are underlined. }
\begin{tabular}{c|l|rrrrrrrr}
\hline
& \multicolumn{1}{c|}{\textbf{Method}} & \multicolumn{1}{c}{$\mathbf{AP}$} & \multicolumn{1}{c}{$\mathbf{F1}$} & \multicolumn{1}{c}{$\mathbf{ACC_r}$} & \multicolumn{1}{c}{$\mathbf{ACC_f}$} & \multicolumn{1}{c}{$\mathbf{ACC}$} & \multicolumn{1}{c}{$\mathbf{AUC_{roc}}$} & \multicolumn{1}{c}{$\mathbf{AUC_{f1}}$} & \multicolumn{1}{c}{$\mathbf{AUC_{f2}}$} \\ \hline

\multirow{11}{*}{\rotatebox{90}{SD-v1}} & CNNDet\cite{wang2020cnn} & 78.45 & 4.45 & 99.40 & 2.28 & 34.65 & 66.81 & 5.66 & 3.85 \\
 & FreDect\cite{frank2020leveraging} & 70.27 & 21.93 & 92.76 & 12.76 & 39.43 & 50.01 & 23.89 & 17.92 \\
 & GramNet\cite{liu2020global} & \textbf{96.53} & \textbf{64.36} & 97.82 & \textbf{47.97} & \textbf{64.59} & \textbf{95.06} & \textbf{64.14} & \textbf{53.33} \\
 & Fusing\cite{ju2022fusing} & 73.70 & 0.62 & \textbf{99.92} & 0.31 & 33.51 & 59.74 & 1.36 & 1.06 \\
 & LNP\cite{liu2022detecting} & 85.71 & 8.36 & 99.30 & 4.38 & 36.02 & 78.06 & 9.54 & 6.45 \\
 & SPrompts\cite{wang2022s} & {\ul 92.99} & 4.59 & {\ul 99.84} & 2.35 & 34.85 & {\ul 88.65} & 6.11 & 4.15 \\
 & UnivFD\cite{ojha2023towards} & 78.62 & 5.32 & 98.88 & 2.75 & 34.79 & 68.52 & 9.58 & 6.77 \\
 & LGrad\cite{tan2023learning} & 79.93 & 0.36 & 99.76 & 0.18 & 33.37 & 71.60 & 0.95 & 0.81 \\
 & NPR\cite{tan2023rethinking} & 87.91 & {\ul 35.08} & 97.84 & {\ul 21.50} & {\ul 46.95} & 80.40 & {\ul 34.99} & {\ul 25.56} \\
 & Freqnet\cite{tan2024frequencyaware} & 69.01 & 1.21 & 99.62 & 0.61 & 33.61 & 51.47 & 1.69 & 1.27 \\
  \cline{2-10} 
 & PoundNet & 83.61 & 16.80 & 98.20 & 9.25 & 38.90 & 74.73 & 18.34 & 13.05 \\
\hline
 
\multirow{11}{*}{\rotatebox{90}{ADM}} & CNNDet\cite{wang2020cnn} & 68.83 & 4.85 & 99.40 & 2.50 & 50.95 & 71.84 & 5.75 & 3.92 \\
 & FreDect\cite{frank2020leveraging} & 80.88 & {\ul 56.17} & 92.76 & {\ul 41.88} & {\ul 67.32} & 83.20 & 50.50 & {\ul 44.86} \\
 & GramNet\cite{liu2020global} & 71.08 & 1.67 & 97.82 & 0.86 & 49.34 & 81.67 & 1.97 & 1.47 \\
 & Fusing\cite{ju2022fusing} & 70.19 & 2.91 & \textbf{99.92} & 1.48 & 50.70 & 68.79 & 4.00 & 2.78 \\
 & LNP\cite{liu2022detecting} & 55.86 & 2.90 & 99.30 & 1.48 & 50.39 & 56.79 & 3.53 & 2.46 \\
 & SPrompts\cite{wang2022s} & 74.81 & 2.68 & {\ul 99.84} & 1.36 & 50.60 & 74.59 & 3.48 & 2.43 \\
 & UnivFD\cite{ojha2023towards} & {\ul 89.93} & 51.21 & 98.88 & 34.80 & 66.84 & {\ul 89.50} & {\ul 50.62} & 40.96 \\
 & LGrad\cite{tan2023learning} & 50.41 & 0.28 & 99.76 & 0.14 & 49.95 & 50.82 & 0.63 & 0.61 \\
 & NPR\cite{tan2023rethinking} & 55.89 & 6.11 & 97.84 & 3.22 & 50.53 & 58.69 & 6.40 & 4.37 \\
 & Freqnet\cite{tan2024frequencyaware} & 62.82 & 4.37 & 99.62 & 2.24 & 50.93 & 60.97 & 4.83 & 3.30 \\
  \cline{2-10} 
 & PoundNet & \textbf{94.91} & \textbf{73.21} & 98.20 & \textbf{58.78} & \textbf{78.49} & \textbf{94.56} & \textbf{71.63} & \textbf{63.10} \\
\hline

\end{tabular}
\end{table}

\begin{table}[]
\small
\caption{Comprehensive comparisons of our method and other ai-generated images detectors on LSUN Bedroom split of DiffusionForensics dataset. \zhiwu{$\mathbf{ACC_r}$, $\mathbf{ACC_f}$ represent detection accurracies for real images and fake images, respectively.} For each test subset, the best results are highlighted in boldface and the second best results are underlined.}
\begin{tabular}{c|l|rrrrrrrr}
\hline
& \multicolumn{1}{c|}{\textbf{Method}} & \multicolumn{1}{c}{$\mathbf{AP}$} & \multicolumn{1}{c}{$\mathbf{F1}$} & \multicolumn{1}{c}{$\mathbf{ACC_r}$} & \multicolumn{1}{c}{$\mathbf{ACC_f}$} & \multicolumn{1}{c}{$\mathbf{ACC}$} & \multicolumn{1}{c}{$\mathbf{AUC_{roc}}$} & \multicolumn{1}{c}{$\mathbf{AUC_{f1}}$} & \multicolumn{1}{c}{$\mathbf{AUC_{f2}}$} \\ \hline
 
\multirow{11}{*}{\rotatebox{90}{SD-v1 part1}} & CNNDet\cite{wang2020cnn} & 51.65 & 0.20 & 99.90 & 0.10 & 50.00 & 51.31 & 0.83 & 0.73 \\
 & FreDect\cite{frank2020leveraging} & 33.33 & {\ul 8.49} & 62.40 & {\ul 6.10} & 34.25 & 11.66 & {\ul 10.31} & {\ul 9.33} \\
 & GramNet\cite{liu2020global} & \textbf{99.87} & \textbf{41.90} & \textbf{100.00} & \textbf{26.50} & \textbf{63.25} & \textbf{99.87} & \textbf{41.61} & \textbf{31.03} \\
 & Fusing\cite{ju2022fusing} & 50.82 & 0.00 & \textbf{100.00} & 0.00 & 50.00 & 51.27 & 0.41 & 0.47 \\
 & LNP\cite{liu2022detecting} & 60.62 & 0.40 & 99.80 & 0.20 & 50.00 & 67.35 & 0.64 & 0.61 \\
 & SPrompts\cite{wang2022s} & {\ul 96.39} & 0.60 & 99.90 & 0.30 & 50.10 & {\ul 98.13} & 1.25 & 1.00 \\
 & UnivFD\cite{ojha2023towards} & 57.29 & 0.00 & 99.80 & 0.00 & 49.90 & 60.48 & 1.11 & 0.93 \\
 & LGrad\cite{tan2023learning} & 81.30 & 0.00 & \textbf{100.00} & 0.00 & 50.00 & 84.51 & 0.43 & 0.48 \\
 & NPR\cite{tan2023rethinking} & 80.41 & 3.15 & \textbf{100.00} & 1.60 & {\ul 50.80} & 81.85 & 3.17 & 2.21 \\
 & Freqnet\cite{tan2024frequencyaware} & 51.75 & 0.20 & 99.90 & 0.10 & 50.00 & 55.85 & 0.56 & 0.56 \\
  \cline{2-10} 
 & PoundNet & 42.80 & 6.82 & 86.70 & 4.00 & 45.35 & 42.94 & 9.62 & 7.79 \\
 \hline

\multirow{11}{*}{\rotatebox{90}{SD-v1 part2}} & CNNDet\cite{wang2020cnn} & 52.83 & 0.99 & 99.90 & 0.50 & 50.20 & 53.51 & 1.48 & 1.14 \\
 & FreDect\cite{frank2020leveraging} & 33.78 & {\ul 8.49} & 62.40 & {\ul 6.10} & 34.25 & 11.32 & {\ul 10.24} & {\ul 9.19} \\
 & GramNet\cite{liu2020global} & \textbf{99.90} & \textbf{38.71} & \textbf{100.00} & \textbf{24.00} & \textbf{62.00} & \textbf{99.90} & \textbf{39.05} & \textbf{28.80} \\
 & Fusing\cite{ju2022fusing} & 53.82 & 0.20 & \textbf{100.00} & 0.10 & 50.05 & 54.52 & 0.63 & 0.60 \\
 & LNP\cite{liu2022detecting} & 61.86 & 0.00 & 99.80 & 0.00 & 49.90 & 69.24 & 0.56 & 0.56 \\
 & SPrompts\cite{wang2022s} & {\ul 96.51} & 0.00 & 99.90 & 0.00 & 49.95 & {\ul 98.32} & 1.04 & 0.87 \\
 & UnivFD\cite{ojha2023towards} & 56.10 & 0.00 & 99.80 & 0.00 & 49.90 & 59.07 & 1.12 & 0.94 \\
 & LGrad\cite{tan2023learning} & 79.47 & 0.00 & \textbf{100.00} & 0.00 & 50.00 & 82.85 & 0.35 & 0.43 \\
 & NPR\cite{tan2023rethinking} & 80.43 & 1.19 & \textbf{100.00} & 0.60 & {\ul 50.30} & 82.25 & 1.83 & 1.36 \\
 & Freqnet\cite{tan2024frequencyaware} & 50.52 & 0.00 & 99.90 & 0.00 & 49.95 & 54.44 & 0.48 & 0.51 \\
  \cline{2-10} 
 & PoundNet & 42.98 & 7.31 & 86.70 & 4.30 & 45.50 & 43.23 & 9.92 & 8.04 \\
 \hline
 
\multirow{11}{*}{\rotatebox{90}{ADM}} & CNNDet\cite{wang2020cnn} & 63.01 & 1.19 & 99.90 & 0.60 & 50.25 & 65.40 & 1.87 & 1.39 \\
 & FreDect\cite{frank2020leveraging} & 56.73 & {\ul 51.96} & 62.40 & {\ul 48.30} & {\ul 55.35} & 58.86 & {\ul 45.23} & {\ul 46.60} \\
 & GramNet\cite{liu2020global} & 74.11 & 0.00 & \textbf{100.00} & 0.00 & 50.00 & 74.66 & 0.34 & 0.42 \\
 & Fusing\cite{ju2022fusing} & 66.40 & 0.60 & \textbf{100.00} & 0.30 & 50.15 & 67.29 & 1.66 & 1.26 \\
 & LNP\cite{liu2022detecting} & 70.43 & 2.56 & 99.80 & 1.30 & 50.55 & 71.34 & 2.90 & 2.05 \\
 & SPrompts\cite{wang2022s} & 60.23 & 0.40 & 99.90 & 0.20 & 50.05 & 63.39 & 0.70 & 0.65 \\
 & UnivFD\cite{ojha2023towards} & {\ul 83.38} & 14.27 & 99.80 & 7.70 & 53.75 & {\ul 83.06} & 17.06 & 12.03 \\
 & LGrad\cite{tan2023learning} & 48.63 & 0.00 & \textbf{100.00} & 0.00 & 50.00 & 48.41 & 0.34 & 0.42 \\
 & NPR\cite{tan2023rethinking} & 42.84 & 0.00 & \textbf{100.00} & 0.00 & 50.00 & 39.17 & 0.34 & 0.42 \\
 & Freqnet\cite{tan2024frequencyaware} & 43.65 & 0.00 & 99.90 & 0.00 & 49.95 & 42.20 & 0.37 & 0.44 \\
  \cline{2-10} 
 & PoundNet & \textbf{86.46} & \textbf{74.39} & 86.70 & \textbf{67.10} & \textbf{76.90} & \textbf{86.17} & \textbf{71.09} & \textbf{67.47} \\
 \hline

\multirow{11}{*}{\rotatebox{90}{Vqdiffusion}} & CNNDet\cite{wang2020cnn} & 58.57 & 0.40 & 99.90 & 0.20 & 50.05 & 61.60 & 1.35 & 1.06 \\
 & FreDect\cite{frank2020leveraging} & 73.75 & \textbf{68.58} & 62.40 & \textbf{71.80} & \textbf{67.10} & 73.69 & \textbf{59.11} & \textbf{61.40} \\
 & GramNet\cite{liu2020global} & 83.81 & 0.00 & \textbf{100.00} & 0.00 & 50.00 & 86.53 & 0.34 & 0.42 \\
 & Fusing\cite{ju2022fusing} & 77.57 & 2.18 & \textbf{100.00} & 1.10 & 50.55 & 76.97 & 3.91 & 2.73 \\
 & LNP\cite{liu2022detecting} & 73.96 & 1.58 & 99.80 & 0.80 & 50.30 & 77.68 & 2.41 & 1.74 \\
 & SPrompts\cite{wang2022s} & {\ul 90.79} & 0.20 & 99.90 & 0.10 & 50.00 & {\ul 94.45} & 0.64 & 0.61 \\
 & UnivFD\cite{ojha2023towards} & \textbf{96.61} & 28.86 & 99.80 & 16.90 & 58.35 & \textbf{96.89} & 33.29 & 26.04 \\
 & LGrad\cite{tan2023learning} & 50.84 & 0.00 & \textbf{100.00} & 0.00 & 50.00 & 52.85 & 0.34 & 0.42 \\
 & NPR\cite{tan2023rethinking} & 36.70 & 0.00 & \textbf{100.00} & 0.00 & 50.00 & 27.94 & 0.34 & 0.42 \\
 & Freqnet\cite{tan2024frequencyaware} & 43.67 & 0.00 & 99.90 & 0.00 & 49.95 & 43.98 & 0.34 & 0.42 \\
  \cline{2-10} 
 & PoundNet & 72.78 & {\ul 55.58} & 86.70 & {\ul 43.60} & {\ul 65.15} & 78.58 & {\ul 51.92} & {\ul 46.71} \\
 \hline
 
\end{tabular}
\end{table}

\begin{table}[]
\small
\caption{Comprehensive comparisons of our method and other ai-generated images detectors on DiffusionForensics dataset. \zhiwu{$\mathbf{ACC_r}$, $\mathbf{ACC_f}$ represent detection accurracies for real images and fake images, respectively.} For each test subset, the best results are highlighted in boldface and the second best results are underlined. }
\begin{tabular}{c|l|rrrrrrrr}
\hline
& \multicolumn{1}{c|}{\textbf{Method}} & \multicolumn{1}{c}{$\mathbf{AP}$} & \multicolumn{1}{c}{$\mathbf{F1}$} & \multicolumn{1}{c}{$\mathbf{ACC_r}$} & \multicolumn{1}{c}{$\mathbf{ACC_f}$} & \multicolumn{1}{c}{$\mathbf{ACC}$} & \multicolumn{1}{c}{$\mathbf{AUC_{roc}}$} & \multicolumn{1}{c}{$\mathbf{AUC_{f1}}$} & \multicolumn{1}{c}{$\mathbf{AUC_{f2}}$} \\ \hline

\multirow{11}{*}{\rotatebox{90}{IF}} & CNNDet\cite{wang2020cnn} & 58.05 & 1.19 & 99.90 & 0.60 & 50.25 & 60.17 & 1.52 & 1.17 \\
 & FreDect\cite{frank2020leveraging} & 55.94 & \textbf{46.09} & 62.40 & \textbf{41.20} & {\ul 51.80} & 53.58 & \textbf{42.40} & \textbf{42.79} \\
 & GramNet\cite{liu2020global} & 70.92 & 0.00 & \textbf{100.00} & 0.00 & 50.00 & 77.88 & 0.34 & 0.42 \\
 & Fusing\cite{ju2022fusing} & {\ul 79.51} & 4.31 & \textbf{100.00} & 2.20 & 51.10 & 77.28 & 6.39 & 4.39 \\
 & LNP\cite{liu2022detecting} & \textbf{92.88} & {\ul 28.42} & 99.80 & {\ul 16.60} & \textbf{58.20} & \textbf{93.41} & {\ul 28.95} & {\ul 20.76} \\
 & SPrompts\cite{wang2022s} & 75.15 & 0.00 & 99.90 & 0.00 & 49.95 & {\ul 85.31} & 0.34 & 0.42 \\
 & UnivFD\cite{ojha2023towards} & 50.42 & 0.00 & 99.80 & 0.00 & 49.90 & 53.87 & 0.61 & 0.60 \\
 & LGrad\cite{tan2023learning} & 60.28 & 0.00 & \textbf{100.00} & 0.00 & 50.00 & 63.65 & 0.34 & 0.42 \\
 & NPR\cite{tan2023rethinking} & 53.47 & 0.00 & \textbf{100.00} & 0.00 & 50.00 & 56.72 & 0.34 & 0.42 \\
 & Freqnet\cite{tan2024frequencyaware} & 37.32 & 0.00 & 99.90 & 0.00 & 49.95 & 29.88 & 0.34 & 0.42 \\
  \cline{2-10} 
 & PoundNet & 37.32 & 2.27 & 86.70 & 1.30 & 44.00 & 29.91 & 4.02 & 3.32 \\
 \hline

\multirow{11}{*}{\rotatebox{90}{Dalle-2}}  & CNNDet\cite{wang2020cnn} & 56.10 & 5.06 & 99.90 & 2.60 & 67.47 & 70.73 & 5.87 & 3.99 \\
 & FreDect\cite{frank2020leveraging} & 26.04 & 15.97 & 62.40 & 15.20 & 46.67 & 21.19 & 15.85 & 16.26 \\
 & GramNet\cite{liu2020global} & \textbf{99.99} & \textbf{88.52} & \textbf{100.00} & \textbf{79.40} & \textbf{93.13} & \textbf{99.99} & \textbf{87.92} & \textbf{82.34} \\
 & Fusing\cite{ju2022fusing} & 46.96 & 0.00 & \textbf{100.00} & 0.00 & 66.67 & 69.64 & 0.38 & 0.44 \\
 & LNP\cite{liu2022detecting} & 97.84 & {\ul 67.72} & 99.80 & {\ul 51.40} & {\ul 83.67} & 98.96 & {\ul 66.43} & {\ul 56.25} \\
 & SPrompts\cite{wang2022s} & {\ul 98.84} & 24.83 & 99.90 & 14.20 & 71.33 & {\ul 99.71} & 29.31 & 22.29 \\
 & UnivFD\cite{ojha2023towards} & 59.90 & 3.52 & 99.80 & 1.80 & 67.13 & 76.77 & 5.99 & 4.19 \\
 & LGrad\cite{tan2023learning} & 97.87 & 34.44 & \textbf{100.00} & 20.80 & 73.60 & 98.92 & 34.40 & 25.29 \\
 & NPR\cite{tan2023rethinking} & 97.18 & 28.77 & \textbf{100.00} & 16.80 & 72.27 & 98.62 & 28.28 & 20.00 \\
 & Freqnet\cite{tan2024frequencyaware} & 85.64 & 18.15 & 99.90 & 10.00 & 69.93 & 92.33 & 19.03 & 13.18 \\
  \cline{2-10} 
 & PoundNet & 33.73 & 10.48 & 86.70 & 7.00 & 60.13 & 56.52 & 13.41 & 12.05 \\
 \hline

\multirow{11}{*}{\rotatebox{90}{MJ}} & CNNDet\cite{wang2020cnn} & 47.94 & 30.25 & 99.90 & 18.00 & 92.45 & 81.34 & 29.29 & 21.13 \\
 & FreDect\cite{frank2020leveraging} & 17.07 & 12.97 & 62.40 & 33.00 & 59.73 & 43.51 & 14.38 & 20.58 \\
 & GramNet\cite{liu2020global} & \textbf{100.00} & \textbf{100.00} & \textbf{100.00} & \textbf{100.00} & \textbf{100.00} & \textbf{100.00} & \textbf{99.37} & \textbf{99.34} \\
 & Fusing\cite{ju2022fusing} & 12.55 & 0.00 & \textbf{100.00} & 0.00 & 90.91 & 60.55 & 0.16 & 0.22 \\
 & LNP\cite{liu2022detecting} & 92.64 & 65.79 & 99.80 & 50.00 & 95.27 & 99.15 & 64.24 & 54.60 \\
 & SPrompts\cite{wang2022s} & 97.82 & 65.33 & 99.90 & 49.00 & 95.27 & {\ul 99.88} & 59.77 & 52.07 \\
 & UnivFD\cite{ojha2023towards} & 5.66 & 0.00 & 99.80 & 0.00 & 90.73 & 24.92 & 0.08 & 0.17 \\
 & LGrad\cite{tan2023learning} & 67.62 & 1.98 & \textbf{100.00} & 1.00 & 91.00 & 94.72 & 3.01 & 2.03 \\
 & NPR\cite{tan2023rethinking} & {\ul 98.19} & {\ul 75.78} & \textbf{100.00} & {\ul 61.00} & {\ul 96.45} & 99.73 & {\ul 75.21} & {\ul 65.85} \\
 & Freqnet\cite{tan2024frequencyaware} & 82.47 & 34.43 & 99.90 & 21.00 & 92.73 & 97.40 & 32.50 & 23.89 \\
  \cline{2-10} 
 & PoundNet & 9.47 & 5.83 & 86.70 & 7.00 & 79.45 & 55.85 & 7.09 & 9.27 \\
 \hline
 
\end{tabular}
\end{table}

\begin{table}[]
\small
\caption{Comprehensive comparisons of our method and other ai-generated images detectors on DiffusionForensics dataset. \zhiwu{$\mathbf{ACC_r}$, $\mathbf{ACC_f}$ represent detection accurracies for real images and fake images, respectively.} For each test subset, the best results are highlighted in boldface and the second best results are underlined.}
\begin{tabular}{c|l|rrrrrrrr}
\hline
& \multicolumn{1}{c|}{\textbf{Method}} & \multicolumn{1}{c}{$\mathbf{AP}$} & \multicolumn{1}{c}{$\mathbf{F1}$} & \multicolumn{1}{c}{$\mathbf{ACC_r}$} & \multicolumn{1}{c}{$\mathbf{ACC_f}$} & \multicolumn{1}{c}{$\mathbf{ACC}$} & \multicolumn{1}{c}{$\mathbf{AUC_{roc}}$} & \multicolumn{1}{c}{$\mathbf{AUC_{f1}}$} & \multicolumn{1}{c}{$\mathbf{AUC_{f2}}$} \\ \hline

\multirow{11}{*}{\rotatebox{90}{iddpm}} & CNNDet\cite{wang2020cnn} & 73.98 & 2.37 & 99.90 & 1.20 & 50.55 & 76.02 & 3.21 & 2.26 \\
 & FreDect\cite{frank2020leveraging} & 60.96 & {\ul 58.87} & 62.40 & {\ul 57.40} & 59.90 & 64.17 & 49.39 & {\ul 51.27} \\
 & GramNet\cite{liu2020global} & 80.54 & 0.20 & \textbf{100.00} & 0.10 & 50.05 & 81.17 & 0.53 & 0.54 \\
 & Fusing\cite{ju2022fusing} & 79.09 & 1.59 & \textbf{100.00} & 0.80 & 50.40 & 79.62 & 3.35 & 2.36 \\
 & LNP\cite{liu2022detecting} & 71.81 & 2.37 & 99.80 & 1.20 & 50.50 & 74.30 & 2.84 & 2.01 \\
 & SPrompts\cite{wang2022s} & 71.22 & 1.59 & 99.90 & 0.80 & 50.35 & 69.69 & 2.49 & 1.79 \\
 & UnivFD\cite{ojha2023towards} & {\ul 96.36} & 54.89 & 99.80 & 37.90 & {\ul 68.85} & {\ul 96.08} & {\ul 54.92} & 44.62 \\
 & LGrad\cite{tan2023learning} & 49.67 & 0.00 & \textbf{100.00} & 0.00 & 50.00 & 49.15 & 0.34 & 0.42 \\
 & NPR\cite{tan2023rethinking} & 44.21 & 0.00 & \textbf{100.00} & 0.00 & 50.00 & 40.40 & 0.36 & 0.44 \\
 & Freqnet\cite{tan2024frequencyaware} & 52.21 & 0.60 & 99.90 & 0.30 & 50.10 & 52.35 & 1.17 & 0.94 \\
  \cline{2-10} 
 & PoundNet & \textbf{97.92} & \textbf{91.53} & 86.70 & \textbf{95.60} & \textbf{91.15} & \textbf{97.81} & \textbf{89.29} & \textbf{91.96} \\
 \hline

\multirow{11}{*}{\rotatebox{90}{ddpm}}& CNNDet\cite{wang2020cnn} & 69.63 & 3.32 & 99.90 & 1.69 & 57.24 & 76.69 & 3.73 & 2.59 \\
 & FreDect\cite{frank2020leveraging} & 49.64 & 52.01 & 62.40 & {\ul 52.34} & 58.03 & 59.79 & 42.46 & {\ul 45.65} \\
 & GramNet\cite{liu2020global} & 78.50 & 0.00 & \textbf{100.00} & 0.00 & 56.56 & 82.48 & 0.34 & 0.42 \\
 & Fusing\cite{ju2022fusing} & 75.31 & 2.82 & \textbf{100.00} & 1.43 & 57.18 & 80.46 & 4.47 & 3.10 \\
 & LNP\cite{liu2022detecting} & 70.86 & 2.31 & 99.80 & 1.17 & 56.96 & 78.47 & 3.60 & 2.50 \\
 & SPrompts\cite{wang2022s} & 54.11 & 1.03 & 99.90 & 0.52 & 56.73 & 59.35 & 1.42 & 1.10 \\
 & UnivFD\cite{ojha2023towards} & \textbf{95.51} & {\ul 53.47} & 99.80 & 36.59 & {\ul 72.34} & \textbf{96.25} & {\ul 53.01} & 42.79 \\
 & LGrad\cite{tan2023learning} & 50.17 & 0.00 & \textbf{100.00} & 0.00 & 56.56 & 57.13 & 0.35 & 0.43 \\
 & NPR\cite{tan2023rethinking} & 36.34 & 0.00 & \textbf{100.00} & 0.00 & 56.56 & 39.04 & 0.31 & 0.40 \\
 & Freqnet\cite{tan2024frequencyaware} & 42.69 & 0.26 & 99.90 & 0.13 & 56.56 & 48.36 & 0.67 & 0.63 \\
  \cline{2-10} 
 & PoundNet & {\ul 94.93} & \textbf{87.59} & 86.70 & \textbf{91.41} & \textbf{88.74} & {\ul 95.81} & \textbf{84.39} & \textbf{86.73} \\
 \hline

\multirow{11}{*}{\rotatebox{90}{pndm}} & CNNDet\cite{wang2020cnn} & 74.10 & 1.59 & 99.90 & 0.80 & 50.35 & 76.33 & 3.23 & 2.27 \\
 & FreDect\cite{frank2020leveraging} & 49.83 & 42.47 & 62.40 & 37.10 & 49.75 & 47.88 & 37.72 & 37.92 \\
 & GramNet\cite{liu2020global} & 89.32 & 0.20 & \textbf{100.00} & 0.10 & 50.05 & 88.54 & 0.53 & 0.54 \\
 & Fusing\cite{ju2022fusing} & 81.84 & 2.76 & \textbf{100.00} & 1.40 & 50.70 & 81.88 & 4.93 & 3.41 \\
 & LNP\cite{liu2022detecting} & 63.50 & 1.58 & 99.80 & 0.80 & 50.30 & 65.71 & 2.08 & 1.52 \\
 & SPrompts\cite{wang2022s} & 91.64 & 19.31 & 99.90 & 10.70 & 55.30 & 90.61 & 20.38 & 14.14 \\
 & UnivFD\cite{ojha2023towards} & {\ul 98.87} & {\ul 80.00} & 99.80 & {\ul 66.80} & {\ul 83.30} & {\ul 98.75} & {\ul 77.92} & {\ul 69.98} \\
 & LGrad\cite{tan2023learning} & 50.23 & 0.00 & \textbf{100.00} & 0.00 & 50.00 & 49.75 & 0.34 & 0.42 \\
 & NPR\cite{tan2023rethinking} & 70.22 & 1.00 & \textbf{100.00} & 0.50 & 50.25 & 66.04 & 1.08 & 0.89 \\
 & Freqnet\cite{tan2024frequencyaware} & 64.45 & 0.80 & 99.90 & 0.40 & 50.15 & 67.95 & 1.26 & 1.00 \\
  \cline{2-10} 
 & PoundNet & \textbf{99.23} & \textbf{93.52} & 86.70 & \textbf{99.50} & \textbf{93.10} & \textbf{99.26} & \textbf{91.44} & \textbf{95.24} \\
 \hline

\multirow{11}{*}{\rotatebox{90}{SD-v2}}  & CNNDet\cite{wang2020cnn} & 74.20 & 8.05 & 99.90 & 4.20 & 52.05 & 73.13 & 9.40 & 6.34 \\
 & FreDect\cite{frank2020leveraging} & 33.51 & 9.29 & 62.40 & 6.70 & 34.55 & 10.91 & 10.34 & 9.23 \\
 & GramNet\cite{liu2020global} & \textbf{99.74} & \textbf{76.16} & \textbf{100.00} & \textbf{61.50} & \textbf{80.75} & \textbf{99.68} & \textbf{75.74} & \textbf{66.37} \\
 & Fusing\cite{ju2022fusing} & 57.32 & 0.00 & \textbf{100.00} & 0.00 & 50.00 & 59.69 & 0.44 & 0.49 \\
 & LNP\cite{liu2022detecting} & 94.15 & {\ul 27.39} & 99.80 & {\ul 15.90} & {\ul 57.85} & 94.65 & {\ul 28.04} & {\ul 20.15} \\
 & SPrompts\cite{wang2022s} & {\ul 97.94} & 0.20 & 99.90 & 0.10 & 50.00 & {\ul 99.10} & 3.47 & 2.54 \\
 & UnivFD\cite{ojha2023towards} & 44.78 & 0.00 & 99.80 & 0.00 & 49.90 & 42.57 & 0.78 & 0.70 \\
 & LGrad\cite{tan2023learning} & 87.77 & 3.54 & \textbf{100.00} & 1.80 & 50.90 & 87.68 & 4.38 & 2.99 \\
 & NPR\cite{tan2023rethinking} & 85.14 & 3.54 & \textbf{100.00} & 1.80 & 50.90 & 85.83 & 3.77 & 2.59 \\
 & Freqnet\cite{tan2024frequencyaware} & 59.00 & 0.60 & 99.90 & 0.30 & 50.10 & 63.93 & 0.95 & 0.81 \\
  \cline{2-10} 
 & PoundNet & 43.66 & 5.49 & 86.70 & 3.20 & 44.95 & 45.31 & 9.34 & 7.67 \\
 \hline

\multirow{11}{*}{\rotatebox{90}{LDM}} & CNNDet\cite{wang2020cnn} & 50.44 & 0.60 & 99.90 & 0.30 & 50.10 & 50.59 & 0.95 & 0.81 \\
 & FreDect\cite{frank2020leveraging} & 49.76 & {\ul 42.83} & 62.40 & {\ul 37.50} & 49.95 & 54.30 & {\ul 38.58} & {\ul 40.80} \\
 & GramNet\cite{liu2020global} & {\ul 91.35} & 0.20 & \textbf{100.00} & 0.10 & 50.05 & {\ul 92.18} & 0.54 & 0.55 \\
 & Fusing\cite{ju2022fusing} & 61.45 & 0.20 & \textbf{100.00} & 0.10 & 50.05 & 61.92 & 1.06 & 0.88 \\
 & LNP\cite{liu2022detecting} & \textbf{93.62} & 19.46 & 99.80 & 10.80 & {\ul 55.30} & \textbf{94.50} & 21.09 & 14.81 \\
 & SPrompts\cite{wang2022s} & 84.44 & 0.00 & 99.90 & 0.00 & 49.95 & 90.51 & 0.37 & 0.44 \\
 & UnivFD\cite{ojha2023towards} & 76.07 & 0.20 & 99.80 & 0.10 & 49.95 & 80.67 & 2.96 & 2.23 \\
 & LGrad\cite{tan2023learning} & 85.95 & 0.00 & \textbf{100.00} & 0.00 & 50.00 & 88.68 & 0.44 & 0.49 \\
 & NPR\cite{tan2023rethinking} & 48.70 & 0.00 & \textbf{100.00} & 0.00 & 50.00 & 49.72 & 0.34 & 0.42 \\
 & Freqnet\cite{tan2024frequencyaware} & 42.85 & 0.40 & 99.90 & 0.20 & 50.05 & 39.90 & 0.73 & 0.67 \\
  \cline{2-10} 
 & PoundNet & 84.85 & \textbf{77.25} & 86.70 & \textbf{71.30} & \textbf{79.00} & 89.36 & \textbf{71.29} & \textbf{68.83} \\
 \hline

\end{tabular}
\end{table}

\begin{table}[]
\small
\caption{Comprehensive comparisons of our method and other ai-generated images detectors on CelebHQ split of DiffusionForensics dataset. \zhiwu{$\mathbf{ACC_r}$, $\mathbf{ACC_f}$ represent detection accurracies for real images and fake images, respectively.} For each test subset, the best results are highlighted in boldface and the second best results are underlined.}
\begin{tabular}{c|l|rrrrrrrr}
\hline
& \multicolumn{1}{c|}{\textbf{Method}} & \multicolumn{1}{c}{$\mathbf{AP}$} & \multicolumn{1}{c}{$\mathbf{F1}$} & \multicolumn{1}{c}{$\mathbf{ACC_r}$} & \multicolumn{1}{c}{$\mathbf{ACC_f}$} & \multicolumn{1}{c}{$\mathbf{ACC}$} & \multicolumn{1}{c}{$\mathbf{AUC_{roc}}$} & \multicolumn{1}{c}{$\mathbf{AUC_{f1}}$} & \multicolumn{1}{c}{$\mathbf{AUC_{f2}}$} \\ \hline

\multirow{11}{*}{\rotatebox{90}{SD-v2}} & CNNDet\cite{wang2020cnn} & 74.57 & 26.37 & 98.60 & 15.40 & 57.00 & 72.69 & 26.46 & 18.94 \\
 & FreDect\cite{frank2020leveraging} & 40.65 & 31.88 & 47.60 & 28.90 & 38.25 & 39.60 & 33.60 & 33.58 \\
 & GramNet\cite{liu2020global} & 36.61 & 13.59 & 69.70 & 9.50 & 39.60 & 19.71 & 13.87 & 11.17 \\
 & Fusing\cite{ju2022fusing} & {\ul 87.64} & 7.51 & \textbf{100.00} & 3.90 & 51.95 & {\ul 87.77} & 10.93 & 7.66 \\
 & LNP\cite{liu2022detecting} & \textbf{97.12} & \textbf{59.05} & 99.50 & \textbf{42.10} & \textbf{70.80} & \textbf{97.41} & \textbf{58.17} & \textbf{47.31} \\
 & SPrompts\cite{wang2022s} & 74.89 & 10.12 & 98.70 & 5.40 & 52.05 & 79.35 & 13.71 & 9.89 \\
 & UnivFD\cite{ojha2023towards} & 56.59 & 24.80 & 91.20 & 15.40 & 53.30 & 55.89 & 25.59 & 20.60 \\
 & LGrad\cite{tan2023learning} & 72.46 & 0.00 & {\ul 99.90} & 0.00 & 49.95 & 74.83 & 0.47 & 0.50 \\
 & NPR\cite{tan2023rethinking} & 34.31 & 0.00 & 68.00 & 0.00 & 34.00 & 19.74 & 0.34 & 0.42 \\
 & Freqnet\cite{tan2024frequencyaware} & 31.50 & 0.79 & 99.00 & 0.40 & 49.70 & 5.66 & 1.17 & 0.95 \\ \cline{2-10} 
 & PoundNet & 70.05 & {\ul 47.15} & 87.20 & {\ul 34.80} & {\ul 61.00} & 77.16 & {\ul 45.68} & {\ul 39.50} \\ \hline

\multirow{11}{*}{\rotatebox{90}{IF}} & CNNDet\cite{wang2020cnn} & 58.98 & 2.92 & 98.60 & 1.50 & 50.05 & 63.30 & 3.86 & 2.71 \\
 & FreDect\cite{frank2020leveraging} & 43.31 & {\ul 48.73} & 47.60 & {\ul 49.10} & 48.35 & 46.52 & {\ul 43.19} & {\ul 45.51} \\
 & GramNet\cite{liu2020global} & 32.12 & 0.00 & 69.70 & 0.00 & 34.85 & 9.69 & 0.37 & 0.45 \\
 & Fusing\cite{ju2022fusing} & 77.87 & 2.57 & \textbf{100.00} & 1.30 & 50.65 & 77.45 & 3.97 & 2.79 \\
 & LNP\cite{liu2022detecting} & \textbf{91.41} & 33.88 & 99.50 & 20.50 & {\ul 60.00} & \textbf{91.71} & 33.80 & 24.69 \\
 & SPrompts\cite{wang2022s} & 39.81 & 0.20 & 98.70 & 0.10 & 49.40 & 36.19 & 0.72 & 0.67 \\
 & UnivFD\cite{ojha2023towards} & 49.46 & 13.06 & 91.20 & 7.60 & 49.40 & 50.78 & 15.87 & 12.66 \\
 & LGrad\cite{tan2023learning} & 59.54 & 0.20 & {\ul 99.90} & 0.10 & 50.00 & 57.82 & 0.57 & 0.57 \\
 & NPR\cite{tan2023rethinking} & 31.18 & 0.00 & 68.00 & 0.00 & 34.00 & 5.67 & 0.34 & 0.42 \\
 & Freqnet\cite{tan2024frequencyaware} & 38.56 & 0.99 & 99.00 & 0.50 & 49.75 & 30.38 & 1.29 & 1.03 \\ \cline{2-10} 
 & PoundNet & {\ul 77.94} & \textbf{62.52} & 87.20 & \textbf{51.30} & \textbf{69.25} & {\ul 82.86} & \textbf{59.50} & \textbf{53.79} \\ \hline
 
\multirow{11}{*}{\rotatebox{90}{Dalle-2}} & CNNDet\cite{wang2020cnn} & 69.08 & 30.64 & 98.60 & 18.60 & 71.93 & 79.82 & 30.20 & 22.24 \\
 & FreDect\cite{frank2020leveraging} & 22.99 & 16.98 & 47.60 & 19.00 & 38.07 & 27.53 & 17.39 & 19.30 \\
 & GramNet\cite{liu2020global} & 67.24 & \textbf{63.20} & 69.70 & \textbf{74.20} & 71.20 & 78.81 & \textbf{63.13} & \textbf{69.28} \\
 & Fusing\cite{ju2022fusing} & 60.38 & 0.00 & \textbf{100.00} & 0.00 & 66.67 & 77.51 & 1.59 & 1.24 \\
 & LNP\cite{liu2022detecting} & {\ul 96.20} & {\ul 60.69} & 99.50 & {\ul 44.00} & \textbf{81.00} & \textbf{98.30} & {\ul 59.90} & {\ul 49.36} \\
 & SPrompts\cite{wang2022s} & 77.58 & 25.21 & 98.70 & 14.80 & 70.73 & 89.92 & 27.98 & 21.79 \\
 & UnivFD\cite{ojha2023towards} & 22.39 & 1.35 & 91.20 & 0.80 & 61.07 & 25.30 & 2.91 & 2.61 \\
 & LGrad\cite{tan2023learning} & \textbf{96.48} & 45.13 & {\ul 99.90} & 29.20 & {\ul 76.33} & {\ul 97.72} & 44.57 & 34.09 \\
 & NPR\cite{tan2023rethinking} & 32.56 & 26.09 & 68.00 & 24.60 & 53.53 & 51.77 & 26.67 & 25.94 \\
 & Freqnet\cite{tan2024frequencyaware} & 53.38 & 15.55 & 99.00 & 8.60 & 68.87 & 66.52 & 16.59 & 11.58 \\ \cline{2-10} 
 & PoundNet & 54.63 & 39.18 & 87.20 & 30.60 & 68.33 & 79.23 & 38.97 & 35.84 \\ \hline

\multirow{11}{*}{\rotatebox{90}{MJ}} & CNNDet\cite{wang2020cnn} & \textbf{84.13} & {\ul 76.09} & 98.60 & 70.00 & {\ul 96.00} & {\ul 97.13} & \textbf{74.13} & \textbf{70.63} \\
 & FreDect\cite{frank2020leveraging} & 21.00 & 26.91 & 47.60 & \textbf{97.00} & 52.09 & 79.10 & 26.28 & 45.50 \\
 & GramNet\cite{liu2020global} & 27.21 & 31.38 & 69.70 & {\ul 75.00} & 70.18 & 78.83 & 31.75 & 48.71 \\
 & Fusing\cite{ju2022fusing} & 19.62 & 0.00 & \textbf{100.00} & 0.00 & 90.91 & 78.78 & 0.20 & 0.25 \\
 & LNP\cite{liu2022detecting} & 60.45 & 22.03 & 99.50 & 13.00 & 91.64 & 93.64 & 22.69 & 16.43 \\
 & SPrompts\cite{wang2022s} & {\ul 82.97} & \textbf{77.17} & 98.70 & 71.00 & \textbf{96.18} & \textbf{98.61} & {\ul 68.16} & {\ul 65.79} \\
 & UnivFD\cite{ojha2023towards} & 5.61 & 0.00 & 91.20 & 0.00 & 82.91 & 23.73 & 0.25 & 0.44 \\
 & LGrad\cite{tan2023learning} & 53.60 & 1.96 & {\ul 99.90} & 1.00 & 90.91 & 91.28 & 3.69 & 2.51 \\
 & NPR\cite{tan2023rethinking} & 6.40 & 1.42 & 68.00 & 3.00 & 62.09 & 33.83 & 1.56 & 2.34 \\
 & Freqnet\cite{tan2024frequencyaware} & 17.78 & 10.34 & 99.00 & 6.00 & 90.55 & 64.04 & 10.75 & 7.89 \\ \cline{2-10} 
 & PoundNet & 15.31 & 15.38 & 87.20 & 19.00 & 81.00 & 73.63 & 16.46 & 20.13 \\ \hline
 
\multirow{11}{*}{\rotatebox{90}{Average}} & CNNDet & 64.70 & 11.14 & 99.56 & 7.84 & 57.31 & 69.36 & 11.60 & 9.29 \\
 & FreDect & 45.53 & {\ul 34.48} & 62.48 & {\ul 36.74} & 49.16 & 46.49 & {\ul 31.71} & {\ul 33.20} \\
 & Freqnet & 51.63 & 4.93 & 99.67 & 2.82 & 56.27 & 53.76 & 5.26 & 3.86 \\
 & Fusing & 62.89 & 1.57 & \textbf{99.99} & 0.80 & 56.23 & 70.62 & 2.77 & 2.01 \\
 & GramNet & 77.71 & 28.89 & 93.02 & 27.73 & 60.31 & 80.37 & 28.99 & 27.53 \\
 & LGrad & 67.90 & 4.88 & {\ul 99.95} & 3.01 & 56.81 & 72.35 & 5.33 & 4.07 \\
 & LNP & {\ul 79.50} & 22.58 & 99.68 & 15.31 & {\ul 61.04} & {\ul 83.37} & 22.86 & 18.01 \\
 & NPR & 56.78 & 10.12 & 92.65 & 7.48 & 53.26 & 56.52 & 10.31 & 8.58 \\
 & PoundNet & 63.48 & \textbf{42.91} & 88.09 & \textbf{38.84} & \textbf{67.24} & 72.38 & \textbf{42.41} & \textbf{40.58} \\
 & SPrompts & \textbf{81.01} & 12.97 & 99.63 & 9.50 & 57.38 & \textbf{84.21} & 13.41 & 11.26 \\ 
 & UnivFD & 62.42 & 18.39 & 97.79 & 12.73 & 60.68 & 65.73 & 19.65 & 16.07 \\ \hline

\end{tabular}
\end{table}

\begin{figure}[]
\centering
\centerline{\includegraphics[width=0.95\linewidth]{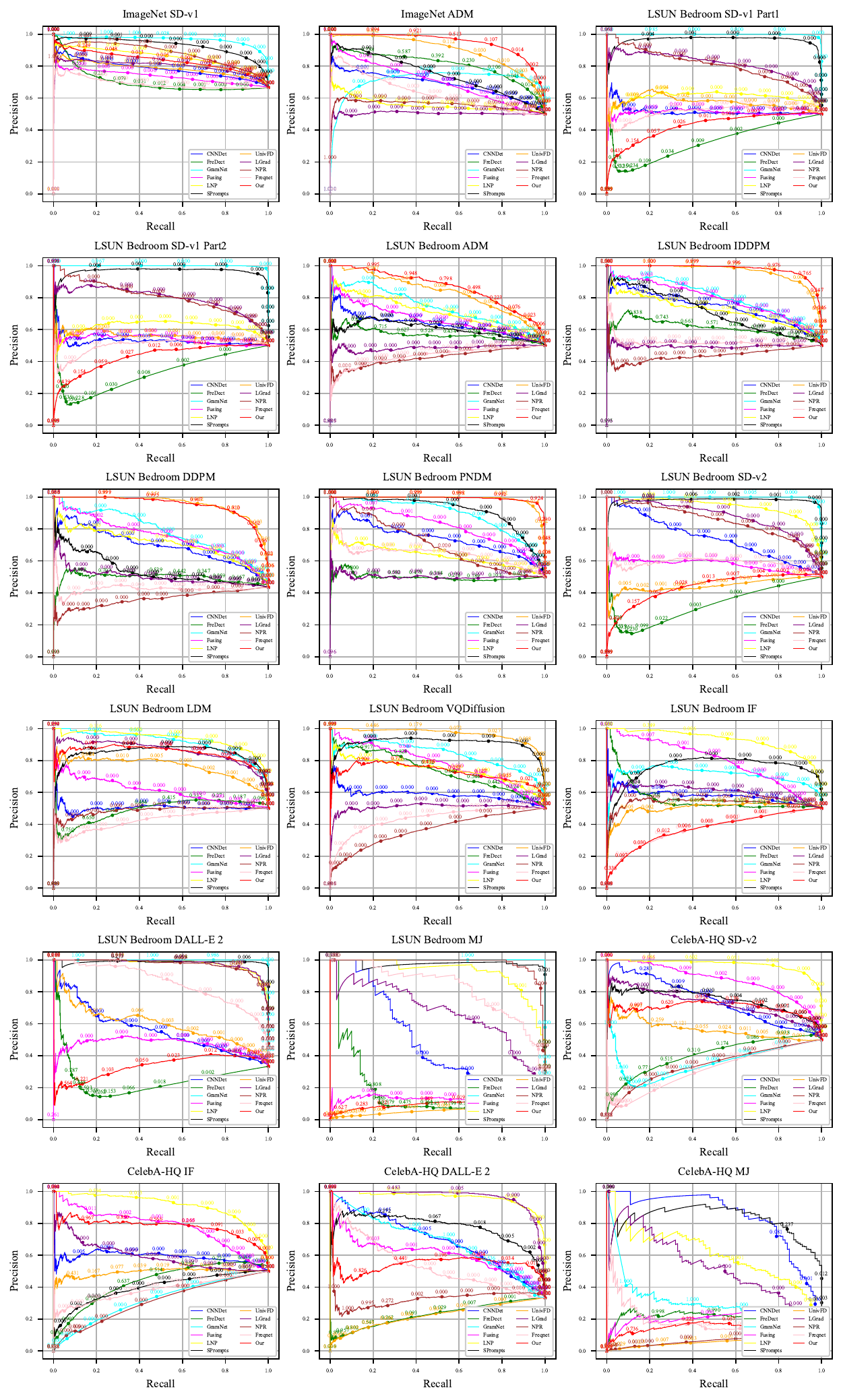}}
\caption{\zhiwu{Precision-Recall Curves of deepfake detection methods on the DiffusionForensics dataset. The numbers on each curve represent the decision thresholds that define the boundary between positive and negative predictions. The numbers typically fall within a narrow range.}}
\label{fig:}
\end{figure}

\begin{figure}[]
\centering
\centerline{\includegraphics[width=0.95\linewidth]{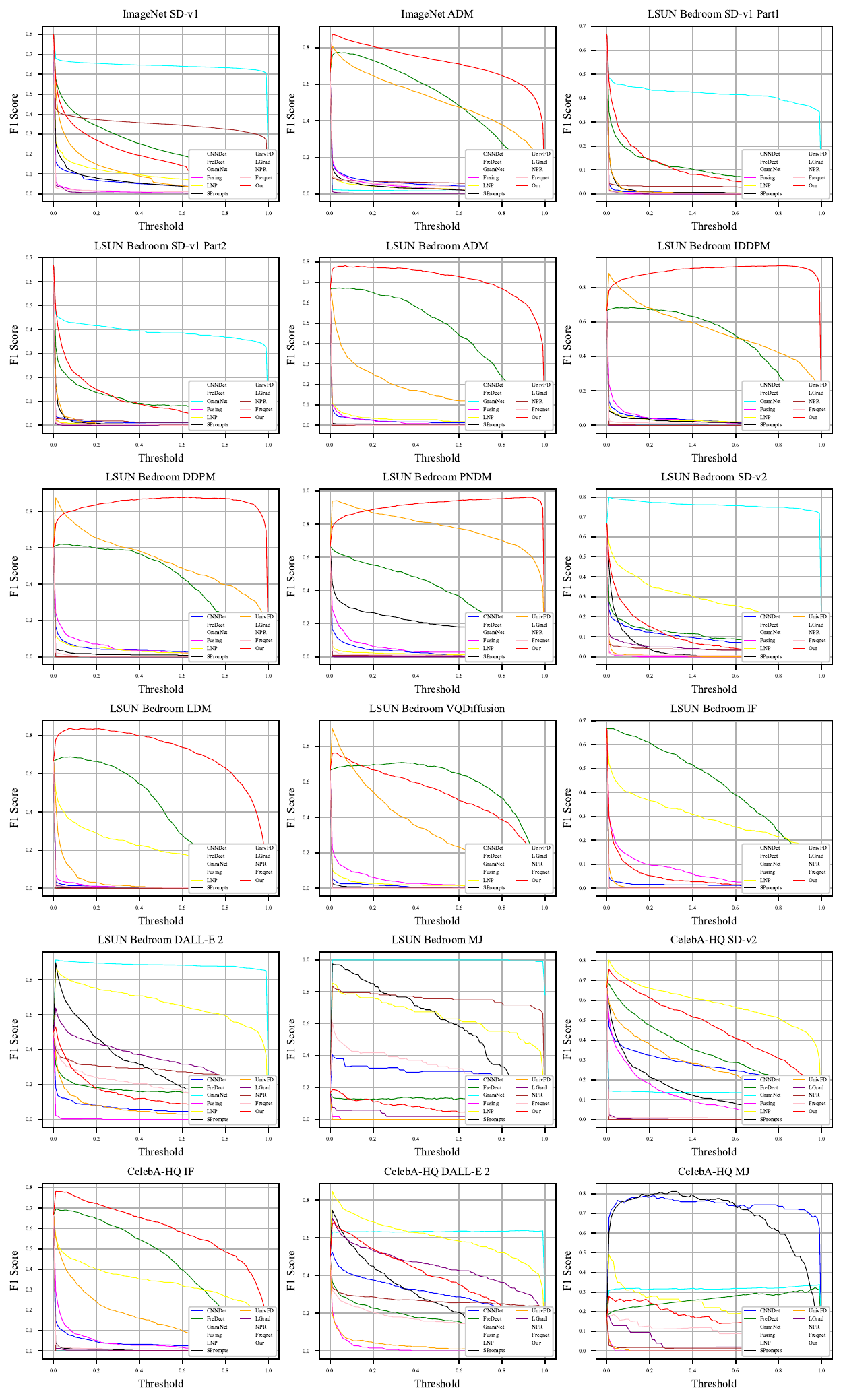}}
\caption{\zhiwu{F1 Curves with different threshold of logits of deepfake detection methods on various deepfakes on the DiffusionForensics dataset. The horizontal axis denotes the decision thresholds that determine the boundary between positive and negative predictions.}}
\label{fig:}
\end{figure}

\newpage

\begin{table}[]
\small
\label{tab:foren}
\caption{Comprehensive comparisons of our method and other ai-generated images detectors on ForenSynths dataset. \zhiwu{$\mathbf{ACC_r}$, $\mathbf{ACC_f}$ represent detection accurracies for real images and fake images, respectively.} For each test subset, the best results are highlighted in boldface and the second best results are underlined.
}
\begin{tabular}{c|l|rrrrrrrr}
\hline
& \multicolumn{1}{c|}{\textbf{Method}} & \multicolumn{1}{c}{$\mathbf{AP}$} & \multicolumn{1}{c}{$\mathbf{F1}$} & \multicolumn{1}{c}{$\mathbf{ACC_r}$} & \multicolumn{1}{c}{$\mathbf{ACC_f}$} & \multicolumn{1}{c}{$\mathbf{ACC}$} & \multicolumn{1}{c}{$\mathbf{AUC_{roc}}$} & \multicolumn{1}{c}{$\mathbf{AUC_{f1}}$} & \multicolumn{1}{c}{$\mathbf{AUC_{f2}}$} \\ \hline
\multirow{11}{*}{\rotatebox{90}{BigGAN}} & CNNDet\cite{wang2020cnn} & 88.04 & 33.81 & 99.00 & 20.55 & 59.77 & 89.25 & 34.53 & 25.54 \\
 & FreDect\cite{frank2020leveraging} & 77.10 & 69.06 & 63.95 & {\ul 71.75} & 67.85 & 73.63 & 64.19 & {\ul 65.56} \\
 & GramNet\cite{liu2020global} & 49.14 & 1.28 & 99.35 & 0.65 & 50.00 & 49.23 & 1.63 & 1.24 \\
 & Fusing\cite{ju2022fusing} & 89.37 & 31.67 & 99.55 & 18.90 & 59.23 & 90.01 & 32.36 & 23.83 \\
 & LNP\cite{liu2022detecting} & 52.31 & 3.85 & 98.05 & 2.00 & 50.02 & 53.54 & 4.50 & 3.13 \\
 & SPrompts\cite{wang2022s} & 92.67 & 49.74 & {\ul 99.85} & 33.15 & 66.50 & 91.59 & 48.90 & 38.06 \\
 & UnivFD\cite{ojha2023towards} & {\ul 95.48} & {\ul 74.44} & 98.80 & 60.00 & {\ul 79.40} & {\ul 95.02} & {\ul 73.04} & 64.60 \\
 & LGrad\cite{tan2023learning} & 50.92 & 0.50 & \textbf{99.95} & 0.25 & 50.10 & 51.93 & 0.94 & 0.80 \\
 & NPR\cite{tan2023rethinking} & 52.42 & 10.87 & 96.50 & 5.95 & 51.23 & 50.83 & 11.16 & 7.67 \\
 & Freqnet\cite{tan2024frequencyaware} & 49.14 & 3.73 & 97.35 & 1.95 & 49.65 & 47.66 & 4.41 & 3.07 \\ \cline{2-10} 
 & PoundNet & \textbf{97.62} & \textbf{91.73} & 89.40 & \textbf{93.70} & \textbf{91.55} & \textbf{97.48} & \textbf{89.77} & \textbf{91.08} \\ \hline

\multirow{11}{*}{\rotatebox{90}{CRN}} & CNNDet\cite{wang2020cnn} & {\ul 98.08} & 64.09 & 99.84 & 47.23 & \textbf{73.53} & 98.08 & {\ul 62.96} & 52.37 \\
 & FreDect\cite{frank2020leveraging} & 59.80 & {\ul 66.38} & 40.14 & {\ul 79.41} & 59.78 & 63.97 & 57.68 & {\ul 63.48} \\
 & GramNet\cite{liu2020global} & 51.85 & 0.00 & 99.98 & 0.00 & 49.99 & 51.84 & 0.34 & 0.42 \\
 & Fusing\cite{ju2022fusing} & 98.00 & 25.70 & \textbf{100.00} & 14.74 & 57.37 & {\ul 98.17} & 27.13 & 19.40 \\
 & LNP\cite{liu2022detecting} & 78.44 & 21.57 & 98.62 & 12.25 & 55.44 & 80.44 & 22.50 & 16.08 \\
 & SPrompts\cite{wang2022s} & 75.30 & 1.00 & 99.97 & 0.50 & 50.24 & 77.54 & 1.71 & 1.29 \\
 & UnivFD\cite{ojha2023towards} & 94.60 & 14.74 & 99.97 & 7.96 & 53.96 & 94.86 & 16.66 & 11.50 \\
 & LGrad\cite{tan2023learning} & 52.56 & 0.00 & \textbf{100.00} & 0.00 & 50.00 & 55.85 & 0.34 & 0.42 \\
 & NPR\cite{tan2023rethinking} & 36.12 & 0.00 & \textbf{100.00} & 0.00 & 50.00 & 26.54 & 0.34 & 0.42 \\
 & Freqnet\cite{tan2024frequencyaware} & 41.78 & 0.03 & 99.98 & 0.02 & 50.00 & 39.89 & 0.38 & 0.45 \\ \cline{2-10} 
 & PoundNet & \textbf{98.52} & \textbf{74.63} & 32.01 & \textbf{100.00} & {\ul 66.01} & \textbf{98.57} & \textbf{75.30} & \textbf{88.00} \\ \hline
 
\multirow{11}{*}{\rotatebox{90}{CycleGAN}} & CNNDet\cite{wang2020cnn} & 95.79 & 76.37 & 98.41 & 62.76 & 80.58 & 95.79 & 74.35 & 66.00 \\
 & FreDect\cite{frank2020leveraging} & 80.08 & 63.43 & 84.94 & 53.44 & 69.19 & 77.44 & 61.79 & 57.52 \\
 & GramNet\cite{liu2020global} & 64.14 & 3.55 & 99.32 & 1.82 & 50.57 & 66.97 & 3.86 & 2.66 \\
 & Fusing\cite{ju2022fusing} & 96.25 & 67.10 & {\ul 99.55} & 50.72 & 75.13 & 96.32 & 64.73 & 54.98 \\
 & LNP\cite{liu2022detecting} & 87.31 & 27.38 & 99.32 & 15.97 & 57.65 & 87.49 & 28.71 & 20.68 \\
 & SPrompts\cite{wang2022s} & 94.44 & 72.63 & 99.24 & 57.46 & 78.35 & 93.54 & 72.00 & 62.58 \\
 & UnivFD\cite{ojha2023towards} & \textbf{98.79} & \textbf{90.57} & 98.94 & {\ul 83.65} & \textbf{91.29} & \textbf{98.72} & {\ul 88.36} & {\ul 84.40} \\
 & LGrad\cite{tan2023learning} & 47.06 & 0.30 & \textbf{99.77} & 0.15 & 49.96 & 47.22 & 0.65 & 0.62 \\
 & NPR\cite{tan2023rethinking} & 55.65 & 18.15 & 93.79 & 10.60 & 52.20 & 53.88 & 17.99 & 12.80 \\
 & Freqnet\cite{tan2024frequencyaware} & 69.26 & 9.37 & 98.33 & 5.00 & 51.67 & 72.57 & 10.03 & 6.82 \\ \cline{2-10} 
 & PoundNet & {\ul 98.26} & {\ul 89.80} & 78.50 & \textbf{99.02} & {\ul 88.76} & {\ul 98.27} & \textbf{88.53} & \textbf{93.65} \\ \hline
 
\multirow{11}{*}{\rotatebox{90}{Deepfake}} & CNNDet\cite{wang2020cnn} & 81.95 & 5.06 & 99.96 & 2.59 & 51.36 & 82.73 & 7.22 & 4.98 \\
 & FreDect\cite{frank2020leveraging} & 60.38 & {\ul 64.98} & 58.66 & {\ul 68.09} & 63.37 & 66.17 & 42.22 & 47.49 \\
 & GramNet\cite{liu2020global} & 59.44 & 2.18 & 99.34 & 1.11 & 50.31 & 61.23 & 2.55 & 1.83 \\
 & Fusing\cite{ju2022fusing} & 66.53 & 20.09 & 96.79 & 11.53 & 54.23 & 68.18 & 22.85 & 17.87 \\
 & LNP\cite{liu2022detecting} & 55.98 & 1.03 & 99.82 & 0.52 & 50.25 & 59.73 & 1.54 & 1.18 \\
 & SPrompts\cite{wang2022s} & {\ul 90.64} & 15.90 & \textbf{100.00} & 8.64 & 54.39 & {\ul 89.03} & 17.41 & 12.16 \\
 & UnivFD\cite{ojha2023towards} & 83.96 & 59.04 & 96.56 & 43.33 & {\ul 69.99} & 81.99 & {\ul 54.45} & {\ul 47.70} \\
 & LGrad\cite{tan2023learning} & 56.40 & 0.22 & 99.93 & 0.11 & 50.10 & 58.05 & 1.12 & 0.92 \\
 & NPR\cite{tan2023rethinking} & 53.91 & 0.07 & \textbf{100.00} & 0.04 & 50.10 & 56.61 & 0.41 & 0.47 \\
 & Freqnet\cite{tan2024frequencyaware} & 48.62 & 0.80 & 98.60 & 0.41 & 49.58 & 51.80 & 1.27 & 1.02 \\ \cline{2-10} 
 & PoundNet & \textbf{96.61} & \textbf{90.51} & 87.62 & \textbf{92.92} & \textbf{90.27} & \textbf{96.31} & \textbf{85.41} & \textbf{86.77} \\ \hline
 
\multirow{11}{*}{\rotatebox{90}{GauGAN}} & CNNDet\cite{wang2020cnn} & 96.25 & 52.12 & 99.72 & 35.34 & 67.53 & 96.47 & 51.79 & 40.95 \\
 & FreDect\cite{frank2020leveraging} & 70.24 & 68.25 & 66.10 & 69.36 & 67.73 & 73.57 & 60.99 & 62.14 \\
 & GramNet\cite{liu2020global} & 53.52 & 0.40 & 99.66 & 0.20 & 49.93 & 55.31 & 0.75 & 0.68 \\
 & Fusing\cite{ju2022fusing} & 97.00 & 49.75 & {\ul 99.74} & 33.20 & 66.47 & 97.12 & 49.28 & 38.81 \\
 & LNP\cite{liu2022detecting} & 57.62 & 0.63 & 99.14 & 0.32 & 49.73 & 60.92 & 1.19 & 0.96 \\
 & SPrompts\cite{wang2022s} & 93.36 & 59.30 & 99.68 & 42.28 & 70.98 & 92.26 & 58.41 & 47.53 \\
 & UnivFD\cite{ojha2023towards} & \textbf{99.75} & \textbf{96.59} & 99.58 & {\ul 93.80} & \textbf{96.69} & \textbf{99.73} & \textbf{94.92} & {\ul 93.12} \\
 & LGrad\cite{tan2023learning} & 49.02 & 0.16 & \textbf{99.94} & 0.08 & 50.01 & 49.80 & 0.50 & 0.52 \\
 & NPR\cite{tan2023rethinking} & 44.58 & 1.48 & 98.28 & 0.76 & 49.52 & 41.73 & 1.78 & 1.34 \\
 & Freqnet\cite{tan2024frequencyaware} & 55.94 & 1.62 & 99.52 & 0.82 & 50.17 & 56.55 & 2.10 & 1.54 \\ \cline{2-10} 
 & PoundNet & {\ul 99.59} & {\ul 95.53} & 92.10 & \textbf{98.66} & {\ul 95.38} & {\ul 99.57} & {\ul 94.05} & \textbf{96.23} \\ \hline
 
\end{tabular}
\end{table}

\begin{table}[]
\small
\caption{Comprehensive comparisons of our method and other ai-generated images detectors on ForenSynths dataset. \zhiwu{$\mathbf{ACC_r}$, $\mathbf{ACC_f}$ represent detection accurracies for real images and fake images, respectively.} For each test subset, the best results are highlighted in boldface and the second best results are underlined.}
\begin{tabular}{c|l|rrrrrrrr}
\hline
& \multicolumn{1}{c|}{\textbf{Method}} & \multicolumn{1}{c}{$\mathbf{AP}$} & \multicolumn{1}{c}{$\mathbf{F1}$} & \multicolumn{1}{c}{$\mathbf{ACC_r}$} & \multicolumn{1}{c}{$\mathbf{ACC_f}$} & \multicolumn{1}{c}{$\mathbf{ACC}$} & \multicolumn{1}{c}{$\mathbf{AUC_{roc}}$} & \multicolumn{1}{c}{$\mathbf{AUC_{f1}}$} & \multicolumn{1}{c}{$\mathbf{AUC_{f2}}$} \\ \hline
\multirow{11}{*}{\rotatebox{90}{IMLE}} & CNNDet\cite{wang2020cnn} & 98.97 & \textbf{77.16} & 99.84 & 62.91 & \textbf{81.38} & 98.92 & \textbf{75.83} & {\ul 67.05} \\
 & FreDect\cite{frank2020leveraging} & 53.73 & 63.03 & 40.14 & {\ul 73.57} & 56.86 & 58.07 & 53.38 & 58.81 \\
 & GramNet\cite{liu2020global} & 66.22 & 0.00 & 99.98 & 0.00 & 49.99 & 66.45 & 0.34 & 0.42 \\
 & Fusing\cite{ju2022fusing} & {\ul 99.01} & 33.76 & \textbf{100.00} & 20.31 & 60.15 & {\ul 99.07} & 34.89 & 25.98 \\
 & LNP\cite{liu2022detecting} & 63.31 & 4.53 & 98.62 & 2.35 & 50.49 & 68.92 & 5.82 & 4.05 \\
 & SPrompts\cite{wang2022s} & 93.76 & 14.12 & 99.97 & 7.60 & 53.78 & 93.52 & 15.83 & 10.95 \\
 & UnivFD\cite{ojha2023towards} & 96.05 & 32.48 & 99.97 & 19.40 & 59.68 & 95.78 & 33.16 & 24.27 \\
 & LGrad\cite{tan2023learning} & 66.90 & 0.00 & \textbf{100.00} & 0.00 & 50.00 & 71.07 & 0.34 & 0.42 \\
 & NPR\cite{tan2023rethinking} & 32.69 & 0.00 & \textbf{100.00} & 0.00 & 50.00 & 13.64 & 0.34 & 0.42 \\
 & Freqnet\cite{tan2024frequencyaware} & 41.06 & 0.09 & 99.98 & 0.05 & 50.02 & 36.11 & 0.42 & 0.47 \\ \cline{2-10} 
 & PoundNet & \textbf{99.25} & {\ul 74.62} & 32.01 & \textbf{99.98} & {\ul 66.00} & \textbf{99.25} & {\ul 75.31} & \textbf{88.01} \\ \hline
 
\multirow{11}{*}{\rotatebox{90}{ProGAN}} & CNNDet\cite{wang2020cnn} & \textbf{100.00} & \textbf{99.92} & 99.92 & \textbf{99.92} & \textbf{99.92} & \textbf{100.00} & \textbf{99.22} & \textbf{99.30} \\
 & FreDect\cite{frank2020leveraging} & 93.55 & 78.18 & 95.17 & 67.27 & 81.23 & 93.77 & 70.28 & 65.76 \\
 & GramNet\cite{liu2020global} & 62.04 & 0.40 & \textbf{99.98} & 0.20 & 50.09 & 64.02 & 0.72 & 0.66 \\
 & Fusing\cite{ju2022fusing} & \textbf{100.00} & {\ul 99.66} & {\ul 99.95} & {\ul 99.38} & {\ul 99.66} & \textbf{100.00} & {\ul 98.73} & {\ul 98.51} \\
 & LNP\cite{liu2022detecting} & 83.12 & 27.94 & 98.70 & 16.45 & 57.57 & 84.81 & 28.52 & 20.62 \\
 & SPrompts\cite{wang2022s} & 98.47 & 57.33 & {\ul 99.95} & 40.20 & 70.08 & 98.34 & 56.56 & 45.53 \\
 & UnivFD\cite{ojha2023towards} & 99.94 & 98.68 & 99.55 & 97.82 & 98.69 & 99.94 & 96.96 & 96.49 \\
 & LGrad\cite{tan2023learning} & 53.49 & 0.15 & {\ul 99.95} & 0.07 & 50.01 & 54.65 & 0.62 & 0.60 \\
 & NPR\cite{tan2023rethinking} & 50.09 & 0.74 & 99.58 & 0.38 & 49.98 & 50.83 & 1.06 & 0.88 \\
 & Freqnet\cite{tan2024frequencyaware} & 56.88 & 2.51 & 99.52 & 1.27 & 50.40 & 56.38 & 2.88 & 2.04 \\ \cline{2-10} 
 & PoundNet & 99.94 & 98.68 & 98.05 & 99.30 & 98.67 & 99.94 & 97.48 & 98.11 \\ \hline
 
\multirow{11}{*}{\rotatebox{90}{SAN}} & CNNDet\cite{wang2020cnn} & 57.32 & 2.69 & {\ul 99.54} & 1.37 & 50.46 & 56.75 & 4.06 & 2.81 \\
 & FreDect\cite{frank2020leveraging} & 48.66 & 8.37 & 95.43 & 4.57 & 50.00 & 46.57 & 12.17 & 9.26 \\
 & GramNet\cite{liu2020global} & 47.13 & \textbf{42.42} & 57.53 & \textbf{38.36} & 47.95 & 45.66 & \textbf{42.77} & \textbf{40.25} \\
 & Fusing\cite{ju2022fusing} & \textbf{71.86} & 3.59 & \textbf{100.00} & 1.83 & 50.91 & {\ul 69.88} & 5.64 & 3.83 \\
 & LNP\cite{liu2022detecting} & 39.37 & 3.38 & 93.61 & 1.83 & 47.72 & 31.23 & 3.88 & 2.78 \\
 & SPrompts\cite{wang2022s} & 60.20 & 7.76 & 98.17 & 4.11 & 51.14 & 61.67 & 10.05 & 7.04 \\
 & UnivFD\cite{ojha2023towards} & 69.15 & 8.70 & {\ul 99.54} & 4.57 & 52.05 & 67.47 & 12.38 & 8.79 \\
 & LGrad\cite{tan2023learning} & 47.85 & 3.51 & 97.72 & 1.83 & 49.77 & 47.57 & 4.16 & 2.93 \\
 & NPR\cite{tan2023rethinking} & 48.84 & {\ul 34.23} & 73.97 & {\ul 26.03} & 50.00 & 46.12 & {\ul 34.12} & {\ul 28.99} \\
 & Freqnet\cite{tan2024frequencyaware} & 56.48 & 14.94 & 98.17 & 8.22 & {\ul 53.20} & 51.12 & 14.98 & 10.29 \\ \cline{2-10} 
 & PoundNet & {\ul 69.43} & 27.17 & 95.43 & 16.44 & \textbf{55.94} & \textbf{72.35} & 27.23 & 20.87 \\ \hline
 
\multirow{11}{*}{\rotatebox{90}{Seeingdark}} & CNNDet\cite{wang2020cnn} & 71.43 & 55.20 & 87.78 & 42.78 & 65.28 & 74.64 & 53.88 & 46.43 \\
 & FreDect\cite{frank2020leveraging} & 34.96 & 19.54 & 46.11 & 16.67 & 31.39 & 20.67 & 19.82 & 18.84 \\
 & GramNet\cite{liu2020global} & 47.59 & 66.79 & 2.78 & \textbf{98.89} & 50.83 & 46.80 & 66.59 & {\ul 82.61} \\
 & Fusing\cite{ju2022fusing} & 79.45 & 35.62 & \textbf{100.00} & 21.67 & 60.83 & 77.10 & 34.61 & 26.31 \\
 & LNP\cite{liu2022detecting} & 70.01 & 69.06 & 48.33 & 80.00 & 64.17 & 72.76 & 68.94 & 75.01 \\
 & SPrompts\cite{wang2022s} & \textbf{95.03} & {\ul 72.77} & 37.78 & {\ul 92.78} & 65.28 & \textbf{92.58} & {\ul 74.49} & \textbf{84.66} \\
 & UnivFD\cite{ojha2023towards} & {\ul 89.82} & 54.84 & \textbf{100.00} & 37.78 & {\ul 68.89} & {\ul 86.49} & 53.62 & 42.76 \\
 & LGrad\cite{tan2023learning} & 35.21 & 23.42 & 36.67 & 21.67 & 29.17 & 22.72 & 24.47 & 23.77 \\
 & NPR\cite{tan2023rethinking} & 54.12 & 70.51 & 31.67 & 91.67 & 61.67 & 60.85 & 70.50 & 81.49 \\
 & Freqnet\cite{tan2024frequencyaware} & 49.70 & 50.70 & 52.78 & 50.00 & 51.39 & 56.38 & 50.90 & 50.70 \\ \cline{2-10} 
 & PoundNet & 87.92 & \textbf{76.57} & 80.00 & 74.44 & \textbf{77.22} & 84.10 & \textbf{75.09} & 74.08 \\ \hline
 
\multirow{11}{*}{\rotatebox{90}{StarGAN}} & CNNDet\cite{wang2020cnn} & 92.40 & 70.73 & 97.60 & 56.03 & 76.81 & 91.49 & 68.47 & 59.55 \\
 & FreDect\cite{frank2020leveraging} & 90.47 & 80.82 & 78.14 & 82.64 & 80.39 & 88.64 & 70.97 & 72.39 \\
 & GramNet\cite{liu2020global} & 68.13 & 0.50 & 99.90 & 0.25 & 50.08 & 69.08 & 0.91 & 0.78 \\
 & Fusing\cite{ju2022fusing} & {\ul 99.09} & {\ul 90.55} & 99.50 & 83.14 & {\ul 91.32} & {\ul 98.97} & {\ul 88.10} & 83.47 \\
 & LNP\cite{liu2022detecting} & 52.58 & 1.77 & 98.90 & 0.90 & 49.90 & 53.98 & 2.90 & 2.09 \\
 & SPrompts\cite{wang2022s} & \textbf{99.79} & 60.50 & \textbf{100.00} & 43.37 & 71.69 & \textbf{99.78} & 60.04 & 49.46 \\
 & UnivFD\cite{ojha2023towards} & 98.68 & \textbf{94.13} & 95.75 & {\ul 92.70} & \textbf{94.22} & 98.55 & \textbf{89.92} & {\ul 89.12} \\
 & LGrad\cite{tan2023learning} & 52.12 & 0.00 & \textbf{100.00} & 0.00 & 50.00 & 53.48 & 0.34 & 0.42 \\
 & NPR\cite{tan2023rethinking} & 41.98 & 0.10 & 99.95 & 0.05 & 50.00 & 37.46 & 0.43 & 0.48 \\
 & Freqnet\cite{tan2024frequencyaware} & 58.43 & 13.46 & 96.00 & 7.50 & 51.75 & 59.83 & 13.53 & 9.49 \\ \cline{2-10} 
 & PoundNet & 97.71 & 79.43 & 48.27 & \textbf{99.95} & 74.11 & 97.85 & 79.40 & \textbf{90.17} \\ \hline
 
\end{tabular}
\end{table}

\begin{table}[]
\small
\caption{Comprehensive comparisons of our method and other ai-generated images detectors on ForenSynths dataset. \zhiwu{$\mathbf{ACC_r}$, $\mathbf{ACC_f}$ represent detection accurracies for real images and fake images, respectively.} For each test subset, the best results are highlighted in boldface and the second best results are underlined.}
\begin{tabular}{c|l|rrrrrrrr}
\hline
& \multicolumn{1}{c|}{\textbf{Method}} & \multicolumn{1}{c}{$\mathbf{AP}$} & \multicolumn{1}{c}{$\mathbf{F1}$} & \multicolumn{1}{c}{$\mathbf{ACC_r}$} & \multicolumn{1}{c}{$\mathbf{ACC_f}$} & \multicolumn{1}{c}{$\mathbf{ACC}$} & \multicolumn{1}{c}{$\mathbf{AUC_{roc}}$} & \multicolumn{1}{c}{$\mathbf{AUC_{f1}}$} & \multicolumn{1}{c}{$\mathbf{AUC_{f2}}$} \\ \hline

\multirow{11}{*}{\rotatebox{90}{StyleGAN}} & CNNDet\cite{wang2020cnn} & {\ul 97.75} & 53.51 & 99.92 & 36.55 & 68.24 & {\ul 97.77} & 53.30 & 42.28 \\
 & FreDect\cite{frank2020leveraging} & 78.26 & {\ul 65.66} & 85.49 & {\ul 55.97} & 70.73 & 71.05 & 59.70 & {\ul 54.68} \\
 & GramNet\cite{liu2020global} & 62.26 & 0.86 & 99.93 & 0.43 & 50.18 & 59.76 & 1.18 & 0.95 \\
 & Fusing\cite{ju2022fusing} & \textbf{98.10} & 54.25 & 99.90 & 37.26 & 68.58 & \textbf{98.11} & 53.45 & 42.81 \\
 & LNP\cite{liu2022detecting} & 73.82 & 13.62 & 99.27 & 7.36 & 53.31 & 74.08 & 14.36 & 9.82 \\
 & SPrompts\cite{wang2022s} & 94.53 & 10.74 & \textbf{99.98} & 5.68 & 52.83 & 94.13 & 12.18 & 8.27 \\
 & UnivFD\cite{ojha2023towards} & 94.72 & 60.94 & 99.72 & 43.95 & {\ul 71.83} & 94.15 & {\ul 59.93} & 49.83 \\
 & LGrad\cite{tan2023learning} & 60.35 & 0.20 & \textbf{99.98} & 0.10 & 50.04 & 63.12 & 0.63 & 0.60 \\
 & NPR\cite{tan2023rethinking} & 51.99 & 0.63 & 99.87 & 0.32 & 50.09 & 47.81 & 1.05 & 0.87 \\
 & Freqnet\cite{tan2024frequencyaware} & 55.26 & 1.58 & 99.37 & 0.80 & 50.08 & 55.95 & 2.16 & 1.58 \\ \cline{2-10} 
 & PoundNet & 96.57 & \textbf{85.16} & 96.56 & \textbf{76.70} & \textbf{86.63} & 96.74 & \textbf{82.35} & \textbf{77.27} \\ \hline
 
\multirow{11}{*}{\rotatebox{90}{StyleGAN2}} & CNNDet\cite{wang2020cnn} & {\ul 96.81} & 41.89 & 99.89 & 26.53 & 63.21 & {\ul 96.90} & 42.06 & 31.82 \\
 & FreDect\cite{frank2020leveraging} & 78.10 & {\ul 59.68} & 89.95 & {\ul 46.81} & {\ul 68.38} & 75.50 & {\ul 53.65} & {\ul 48.17} \\
 & GramNet\cite{liu2020global} & 59.59 & 0.20 & 99.94 & 0.10 & 50.02 & 60.52 & 0.54 & 0.55 \\
 & Fusing\cite{ju2022fusing} & \textbf{97.28} & 29.86 & 99.92 & 17.56 & 58.74 & \textbf{97.50} & 31.11 & 22.99 \\
 & LNP\cite{liu2022detecting} & 73.48 & 7.12 & 99.29 & 3.72 & 51.50 & 75.68 & 8.06 & 5.47 \\
 & SPrompts\cite{wang2022s} & 87.92 & 4.45 & \textbf{99.96} & 2.28 & 51.12 & 88.45 & 5.46 & 3.72 \\
 & UnivFD\cite{ojha2023towards} & 93.42 & 39.67 & 99.61 & 24.84 & 62.22 & 93.52 & 42.13 & 33.17 \\
 & LGrad\cite{tan2023learning} & 61.90 & 0.72 & \textbf{99.96} & 0.36 & 50.16 & 64.33 & 1.19 & 0.96 \\
 & NPR\cite{tan2023rethinking} & 48.34 & 0.00 & 99.77 & 0.00 & 49.89 & 47.72 & 0.34 & 0.42 \\
 & Freqnet\cite{tan2024frequencyaware} & 62.69 & 6.84 & 99.22 & 3.57 & 51.40 & 62.36 & 7.27 & 4.91 \\ \cline{2-10} 
 & PoundNet & 93.63 & \textbf{67.88} & 98.11 & \textbf{52.35} & \textbf{75.23} & 93.71 & \textbf{66.02} & \textbf{56.73} \\ \hline
\multirow{11}{*}{\rotatebox{90}{whichfaceisreal}} & CNNDet\cite{wang2020cnn} & 82.80 & 70.01 & 88.40 & 60.10 & 74.25 & 85.08 & 69.00 & 63.28 \\
 & FreDect\cite{frank2020leveraging} & 51.96 & 15.92 & 89.00 & 9.60 & 49.30 & 55.16 & 18.37 & 14.53 \\
 & GramNet\cite{liu2020global} & 56.72 & 67.27 & 2.90 & \textbf{99.90} & 51.40 & 61.49 & 67.28 & {\ul 83.65} \\
 & Fusing\cite{ju2022fusing} & 93.84 & 12.56 & \textbf{100.00} & 6.70 & 53.35 & {\ul 94.90} & 17.03 & 12.05 \\
 & LNP\cite{liu2022detecting} & 56.19 & 24.61 & 86.70 & 15.90 & 51.30 & 60.81 & 25.24 & 19.56 \\
 & SPrompts\cite{wang2022s} & \textbf{96.40} & \textbf{90.26} & 84.50 & {\ul 95.00} & \textbf{89.75} & \textbf{96.72} & \textbf{87.52} & \textbf{90.47} \\
 & UnivFD\cite{ojha2023towards} & {\ul 94.12} & 61.57 & {\ul 99.50} & 44.70 & 72.10 & 93.85 & 60.97 & 51.30 \\
 & LGrad\cite{tan2023learning} & 56.85 & 21.46 & 91.00 & 13.10 & 52.05 & 60.15 & 22.68 & 17.06 \\
 & NPR\cite{tan2023rethinking} & 48.07 & 31.90 & 70.90 & 24.50 & 47.70 & 51.48 & 32.00 & 27.35 \\
 & Freqnet\cite{tan2024frequencyaware} & 45.36 & 5.89 & 87.90 & 3.40 & 45.65 & 49.17 & 6.61 & 4.87 \\ \cline{2-10} 
 & PoundNet & 88.09 & {\ul 77.96} & 83.70 & 74.30 & {\ul 79.00} & 88.63 & {\ul 73.54} & 71.89 \\ \hline
 
\multirow{11}{*}{\rotatebox{90}{Average}} & CNNDet\cite{wang2020cnn} & 89.05 & 54.04 & 97.68 & 42.67 & 70.18 & 89.53 & 53.59 & 46.33 \\
 & FreDect & 67.48 & 55.64 & 71.79 & {\ul 53.78} & 62.78 & 66.48 & 49.63 & 49.13 \\
 & Freqnet & 53.12 & 8.58 & 94.36 & 6.38 & 50.38 & 53.52 & 9.00 & 7.48 \\
 & Fusing & 91.21 & 42.63 & \textbf{99.61} & 32.07 & 65.84 & 91.18 & 43.07 & 36.22 \\
 & GramNet & 57.52 & 14.30 & 81.58 & 18.61 & 50.10 & 58.34 & 14.57 & 16.67 \\
 & LGrad & 53.12 & 3.90 & 94.22 & 2.90 & 48.57 & 53.84 & 4.46 & 3.85 \\
 & LNP & 64.89 & 15.88 & 93.72 & 12.27 & 53.00 & 66.49 & 16.63 & 13.96 \\
 & NPR & 47.60 & 12.98 & 89.56 & 12.33 & 50.95 & 45.04 & 13.19 & 12.58 \\
 & PoundNet & \textbf{94.09} & \textbf{79.21} & 77.83 & \textbf{82.91} & \textbf{80.37} & \textbf{94.06} & \textbf{77.65} & \textbf{79.45} \\
 & SPrompts & 90.19 & 39.73 & 93.77 & 33.31 & 63.55 & 89.94 & 40.04 & 35.52 \\ 
 \cline{2-10} 
 & UnivFD & {\ul 92.96} & {\ul 60.49} & {\ul 99.04} & 50.35 & {\ul 74.69} & {\ul 92.31} & {\ul 59.73} & {\ul 53.62} \\ \hline

\end{tabular}
\end{table}

\begin{figure}[]
\centering
\centerline{\includegraphics[width=1\linewidth]{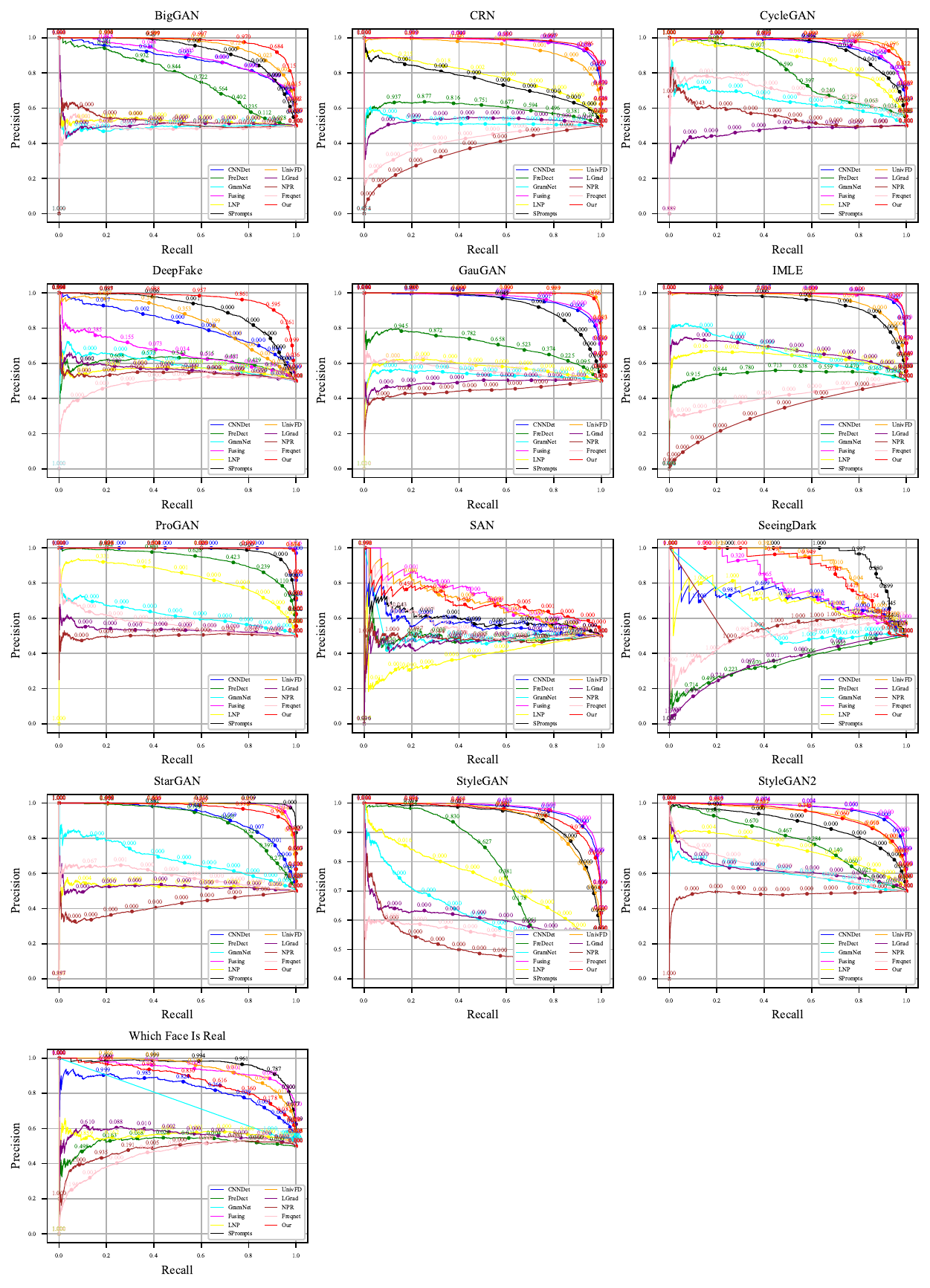}}
\caption{\zhiwu{Precision-Recall Curves of deepfake detection methods on the ForenSynths dataset. The numbers on each curve represent the decision thresholds that define the boundary between positive and negative predictions. The numbers typically fall within a narrow range.}}
\label{fig:}
\end{figure}

\begin{figure}[]
\centering
\centerline{\includegraphics[width=1\linewidth]{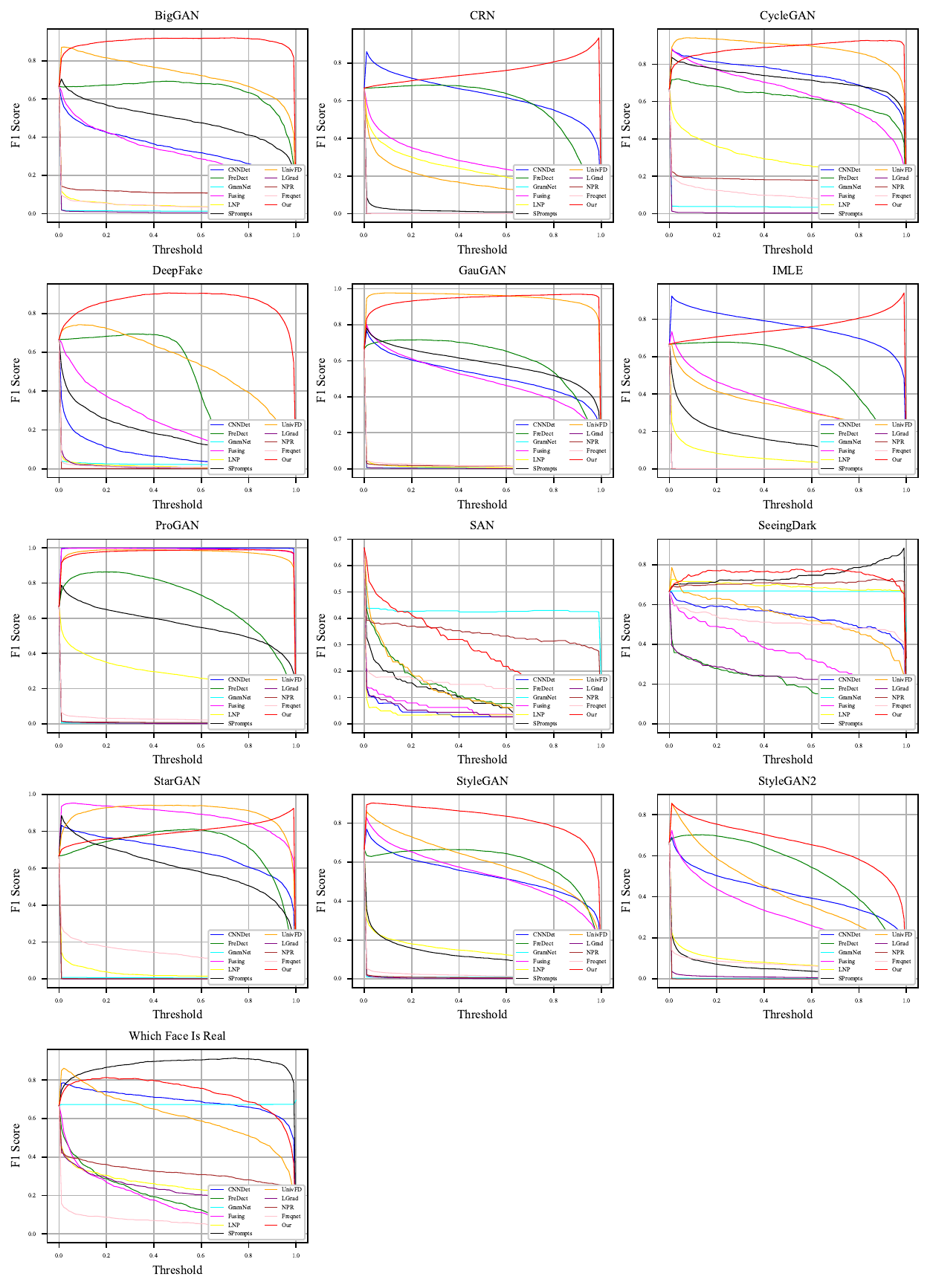}}
\caption{\zhiwu{F1 Curves with different threshold of logits of deepfake detection methods on various deepfakes on the ForenSynths dataset. The horizontal axis denotes the decision thresholds that determine the boundary between positive and negative predictions.}}
\label{fig:}
\end{figure}

\newpage

\begin{table}[]
\small
\caption{Comprehensive comparisons of our method and other ai-generated images detectors on GANGen-Detection dataset. \zhiwu{$\mathbf{ACC_r}$, $\mathbf{ACC_f}$ represent detection accurracies for real images and fake images, respectively.} For each test subset, the best results are highlighted in boldface and the second best results are underlined.}

\begin{tabular}{c|l|rrrrrrrr}
\hline
& \multicolumn{1}{c|}{\textbf{Method}} & \multicolumn{1}{c}{$\mathbf{AP}$} & \multicolumn{1}{c}{$\mathbf{F1}$} & \multicolumn{1}{c}{$\mathbf{ACC_r}$} & \multicolumn{1}{c}{$\mathbf{ACC_f}$} & \multicolumn{1}{c}{$\mathbf{ACC}$} & \multicolumn{1}{c}{$\mathbf{AUC_{roc}}$} & \multicolumn{1}{c}{$\mathbf{AUC_{f1}}$} & \multicolumn{1}{c}{$\mathbf{AUC_{f2}}$} \\ \hline

\multirow{11}{*}{\rotatebox{90}{AttGAN}} & CNNDet\cite{wang2020cnn} & 90.14 & 28.73 & 99.55 & 16.85 & 58.20 & {\ul 90.45} & 29.82 & 21.67 \\
 & FreDect\cite{frank2020leveraging} & 41.31 & 31.33 & 54.10 & 27.10 & 40.60 & 39.78 & 32.02 & 33.62 \\
 & GramNet\cite{liu2020global} & 66.63 & 0.20 & \textbf{100.00} & 0.10 & 50.05 & 68.34 & 0.54 & 0.55 \\
 & Fusing\cite{ju2022fusing} & 88.68 & 30.46 & 99.55 & 18.05 & 58.80 & 88.79 & 32.00 & 23.96 \\
 & LNP\cite{liu2022detecting} & 84.73 & 4.87 & 99.85 & 2.50 & 51.18 & 86.25 & 6.60 & 4.48 \\
 & SPrompts\cite{wang2022s} & 86.00 & 6.01 & \textbf{100.00} & 3.10 & 51.55 & 83.75 & 7.77 & 5.27 \\
 & UnivFD\cite{ojha2023towards} & {\ul 91.37} & {\ul 70.68} & 97.90 & {\ul 55.80} & {\ul 76.85} & 89.30 & {\ul 67.26} & {\ul 59.14} \\
 & LGrad\cite{tan2023learning} & 50.92 & 0.00 & \textbf{100.00} & 0.00 & 50.00 & 52.29 & 0.34 & 0.42 \\
 & NPR\cite{tan2023rethinking} & 40.60 & 0.60 & 99.95 & 0.30 & 50.12 & 33.24 & 0.89 & 0.77 \\
 & Freqnet\cite{tan2024frequencyaware} & 56.29 & 19.22 & 93.25 & 11.35 & 52.30 & 56.13 & 19.73 & 14.27 \\ \cline{2-10} 
 & PoundNet & \textbf{93.82} & \textbf{87.06} & 78.05 & \textbf{94.00} & \textbf{86.02} & \textbf{94.41} & \textbf{85.23} & \textbf{88.77} \\ \hline
 
\multirow{11}{*}{\rotatebox{90}{BEGAN}} & CNNDet\cite{wang2020cnn} & 51.27 & 1.77 & 99.15 & 0.90 & 50.02 & 52.07 & 3.82 & 2.77 \\
 & FreDect\cite{frank2020leveraging} & 92.80 & 16.14 & 99.75 & 8.80 & 54.27 & 94.27 & 29.70 & 25.65 \\
 & GramNet\cite{liu2020global} & 45.73 & 0.20 & 99.50 & 0.10 & 49.80 & 48.00 & 0.52 & 0.54 \\
 & Fusing\cite{ju2022fusing} & {\ul 95.61} & 57.37 & 99.30 & 40.50 & 69.90 & {\ul 95.74} & 55.20 & 46.12 \\
 & LNP\cite{liu2022detecting} & 40.04 & 0.39 & 96.80 & 0.20 & 48.50 & 35.92 & 1.60 & 1.29 \\
 & SPrompts\cite{wang2022s} & \textbf{99.53} & 66.87 & {\ul 99.95} & 50.25 & 75.10 & \textbf{99.54} & 64.71 & 55.20 \\
 & UnivFD\cite{ojha2023towards} & 86.67 & {\ul 76.78} & 84.20 & 72.15 & \textbf{78.17} & 87.37 & {\ul 70.88} & 69.26 \\
 & LGrad\cite{tan2023learning} & 59.37 & 0.00 & \textbf{100.00} & 0.00 & 50.00 & 64.50 & 0.34 & 0.43 \\
 & NPR\cite{tan2023rethinking} & 49.43 & 66.70 & 0.15 & \textbf{100.00} & 50.08 & 48.86 & 66.69 & {\ul 83.34} \\
 & Freqnet\cite{tan2024frequencyaware} & 38.88 & 66.67 & 1.00 & {\ul 99.50} & 50.25 & 33.90 & 66.26 & 82.53 \\ \cline{2-10} 
 & PoundNet & 94.47 & \textbf{80.88} & 53.60 & 99.40 & {\ul 76.50} & 94.84 & \textbf{79.85} & \textbf{89.15} \\ \hline
 
\multirow{11}{*}{\rotatebox{90}{CramerGAN}} & CNNDet\cite{wang2020cnn} & 95.40 & 64.33 & 99.20 & 47.80 & 73.50 & 95.24 & 63.24 & 53.33 \\
 & FreDect\cite{frank2020leveraging} & 52.21 & 1.38 & 99.40 & 0.70 & 50.05 & 53.75 & 5.28 & 4.07 \\
 & GramNet\cite{liu2020global} & 49.42 & 0.20 & {\ul 99.95} & 0.10 & 50.02 & 48.71 & 0.63 & 0.60 \\
 & Fusing\cite{ju2022fusing} & 98.46 & 87.29 & 98.70 & 78.45 & 88.58 & 98.46 & 84.16 & 78.88 \\
 & LNP\cite{liu2022detecting} & 49.88 & 4.64 & 96.80 & 2.45 & 49.62 & 51.03 & 6.38 & 4.56 \\
 & SPrompts\cite{wang2022s} & 94.05 & 31.62 & 99.90 & 18.80 & 59.35 & 94.08 & 32.19 & 23.72 \\
 & UnivFD\cite{ojha2023towards} & {\ul 98.80} & \textbf{94.24} & 91.20 & 96.95 & \textbf{94.08} & {\ul 98.84} & \textbf{89.74} & {\ul 91.38} \\
 & LGrad\cite{tan2023learning} & 54.23 & 0.00 & \textbf{100.00} & 0.00 & 50.00 & 55.51 & 0.38 & 0.45 \\
 & NPR\cite{tan2023rethinking} & 50.62 & 66.69 & 0.10 & \textbf{100.00} & 50.05 & 51.23 & 66.69 & 83.35 \\
 & Freqnet\cite{tan2024frequencyaware} & 48.40 & 67.13 & 5.95 & 98.05 & 52.00 & 50.92 & 66.74 & 82.23 \\ \cline{2-10} 
 & PoundNet & \textbf{98.88} & {\ul 90.31} & 78.90 & {\ul 99.70} & {\ul 89.30} & \textbf{98.92} & {\ul 88.86} & \textbf{94.24} \\ \hline
 
\multirow{11}{*}{\rotatebox{90}{GANimation}} & CNNDet\cite{wang2020cnn} & 47.07 & 1.85 & 98.30 & 0.95 & 49.62 & 46.42 & 3.35 & 2.44 \\
 & FreDect\cite{frank2020leveraging} & 53.52 & 0.89 & 99.20 & 0.45 & 49.83 & 56.29 & 7.32 & 6.19 \\
 & GramNet\cite{liu2020global} & 46.23 & 0.10 & 99.50 & 0.05 & 49.78 & 46.93 & 0.46 & 0.50 \\
 & Fusing\cite{ju2022fusing} & 56.31 & 26.69 & 88.95 & 17.10 & 53.02 & 55.20 & 27.88 & 23.51 \\
 & LNP\cite{liu2022detecting} & 47.84 & 4.45 & 96.70 & 2.35 & 49.53 & 48.11 & 5.88 & 4.20 \\
 & SPrompts\cite{wang2022s} & {\ul 64.01} & 0.00 & \textbf{100.00} & 0.00 & 50.00 & {\ul 63.66} & 0.34 & 0.42 \\
 & UnivFD\cite{ojha2023towards} & 63.33 & 31.86 & 92.60 & 20.35 & {\ul 56.47} & 60.83 & 31.56 & 26.70 \\
 & LGrad\cite{tan2023learning} & 49.64 & 0.00 & \textbf{100.00} & 0.00 & 50.00 & 50.68 & 0.36 & 0.43 \\
 & NPR\cite{tan2023rethinking} & 51.61 & \textbf{66.69} & 0.10 & \textbf{100.00} & 50.05 & 53.05 & \textbf{66.70} & \textbf{83.34} \\
 & Freqnet\cite{tan2024frequencyaware} & 51.50 & {\ul 65.99} & 14.10 & {\ul 91.55} & 52.83 & 53.60 & {\ul 65.53} & {\ul 78.58} \\ \cline{2-10} 
 & PoundNet & \textbf{65.28} & 63.12 & 57.10 & 65.90 & \textbf{61.50} & \textbf{65.67} & 59.42 & 61.28 \\ \hline
 
\end{tabular}
\end{table}

\begin{table}[]
\small
\caption{Comprehensive comparisons of our method and other ai-generated images detectors on GANGen-Detection dataset. \zhiwu{$\mathbf{ACC_r}$, $\mathbf{ACC_f}$ represent detection accurracies for real images and fake images, respectively.} For each test subset, the best results are highlighted in boldface and the second best results are underlined.}

\begin{tabular}{c|l|rrrrrrrr}
\hline
& \multicolumn{1}{c|}{\textbf{Method}} & \multicolumn{1}{c}{$\mathbf{AP}$} & \multicolumn{1}{c}{$\mathbf{F1}$} & \multicolumn{1}{c}{$\mathbf{ACC_r}$} & \multicolumn{1}{c}{$\mathbf{ACC_f}$} & \multicolumn{1}{c}{$\mathbf{ACC}$} & \multicolumn{1}{c}{$\mathbf{AUC_{roc}}$} & \multicolumn{1}{c}{$\mathbf{AUC_{f1}}$} & \multicolumn{1}{c}{$\mathbf{AUC_{f2}}$} \\ \hline

\multirow{11}{*}{\rotatebox{90}{InfoMaxGAN}} & CNNDet\cite{wang2020cnn} & 90.20 & 46.63 & 99.20 & 30.65 & 64.92 & 89.30 & 47.31 & 37.22 \\
 & FreDect\cite{frank2020leveraging} & 49.74 & 0.20 & 99.40 & 0.10 & 49.75 & 52.83 & 3.37 & 2.72 \\
 & GramNet\cite{liu2020global} & 57.27 & 0.70 & {\ul 99.95} & 0.35 & 50.15 & 57.45 & 1.08 & 0.89 \\
 & Fusing\cite{ju2022fusing} & 95.20 & 71.32 & 98.70 & 56.15 & 77.42 & 95.14 & 68.32 & 59.56 \\
 & LNP\cite{liu2022detecting} & 65.06 & 17.59 & 96.80 & 9.95 & 53.37 & 66.43 & 19.34 & 14.05 \\
 & SPrompts\cite{wang2022s} & 92.72 & 31.05 & 99.90 & 18.40 & 59.15 & 92.48 & 31.65 & 23.18 \\
 & UnivFD\cite{ojha2023towards} & {\ul 98.10} & \textbf{92.86} & 91.20 & 94.30 & \textbf{92.75} & {\ul 98.06} & {\ul 88.10} & {\ul 89.03} \\
 & LGrad\cite{tan2023learning} & 45.58 & 0.00 & \textbf{100.00} & 0.00 & 50.00 & 42.99 & 0.37 & 0.44 \\
 & NPR\cite{tan2023rethinking} & 50.22 & 66.69 & 0.10 & \textbf{100.00} & 50.05 & 50.43 & 66.69 & 83.34 \\
 & Freqnet\cite{tan2024frequencyaware} & 51.24 & 65.17 & 5.95 & 93.80 & 49.88 & 50.12 & 64.77 & 79.20 \\ \cline{2-10} 
 & PoundNet & \textbf{98.81} & {\ul 90.33} & 78.90 & {\ul 99.75} & {\ul 89.33} & \textbf{98.87} & \textbf{88.84} & \textbf{94.20} \\ \hline
 
\multirow{11}{*}{\rotatebox{90}{MMDGAN}} & CNNDet\cite{wang2020cnn} & 91.95 & 50.78 & 99.20 & 34.30 & 66.75 & 91.66 & 50.27 & 40.08 \\
 & FreDect\cite{frank2020leveraging} & 48.13 & 0.50 & 99.40 & 0.25 & 49.83 & 48.84 & 3.77 & 2.96 \\
 & GramNet\cite{liu2020global} & 48.77 & 0.00 & {\ul 99.95} & 0.00 & 49.98 & 48.07 & 0.34 & 0.42 \\
 & Fusing\cite{ju2022fusing} & {\ul 97.41} & 81.00 & 98.70 & 68.95 & 83.83 & {\ul 97.41} & 77.87 & 70.91 \\
 & LNP\cite{liu2022detecting} & 50.63 & 5.28 & 96.80 & 2.80 & 49.80 & 51.90 & 7.16 & 5.10 \\
 & SPrompts\cite{wang2022s} & 90.55 & 19.31 & 99.90 & 10.70 & 55.30 & 90.86 & 20.61 & 14.50 \\
 & UnivFD\cite{ojha2023towards} & \textbf{98.65} & \textbf{93.85} & 91.20 & 96.20 & \textbf{93.70} & \textbf{98.67} & \textbf{89.36} & {\ul 90.85} \\
 & LGrad\cite{tan2023learning} & 52.11 & 0.00 & \textbf{100.00} & 0.00 & 50.00 & 53.99 & 0.37 & 0.44 \\
 & NPR\cite{tan2023rethinking} & 50.55 & 66.69 & 0.10 & \textbf{100.00} & 50.05 & 51.08 & 66.69 & 83.34 \\
 & Freqnet\cite{tan2024frequencyaware} & 50.92 & 66.61 & 5.95 & 96.90 & 51.42 & 51.87 & 66.16 & 81.33 \\ \cline{2-10} 
 & PoundNet & 97.06 & {\ul 89.66} & 78.90 & {\ul 98.40} & {\ul 88.65} & 97.28 & {\ul 87.46} & \textbf{92.13} \\ \hline
 
\multirow{11}{*}{\rotatebox{90}{RelGAN}} & CNNDet\cite{wang2020cnn} & 96.29 & 67.09 & 99.65 & 50.65 & 75.15 & 95.84 & 65.24 & 54.85 \\
 & FreDect\cite{frank2020leveraging} & 93.88 & 76.95 & 44.80 & {\ul 97.05} & 70.93 & 92.84 & 76.09 & {\ul 84.79} \\
 & GramNet\cite{liu2020global} & 40.99 & 0.00 & \textbf{100.00} & 0.00 & 50.00 & 36.91 & 0.34 & 0.42 \\
 & Fusing\cite{ju2022fusing} & {\ul 97.60} & 74.88 & 99.50 & 60.15 & 79.83 & {\ul 97.38} & 72.44 & 63.69 \\
 & LNP\cite{liu2022detecting} & 89.28 & 11.92 & 99.85 & 6.35 & 53.10 & 89.99 & 13.48 & 9.23 \\
 & SPrompts\cite{wang2022s} & 90.76 & 11.23 & \textbf{100.00} & 5.95 & 52.98 & 88.74 & 13.46 & 9.28 \\
 & UnivFD\cite{ojha2023towards} & 96.47 & {\ul 83.41} & 98.45 & 72.65 & {\ul 85.55} & 95.83 & {\ul 79.11} & 73.25 \\
 & LGrad\cite{tan2023learning} & 52.77 & 0.00 & \textbf{100.00} & 0.00 & 50.00 & 55.18 & 0.34 & 0.42 \\
 & NPR\cite{tan2023rethinking} & 39.04 & 0.00 & 99.95 & 0.00 & 49.98 & 33.93 & 0.34 & 0.42 \\
 & Freqnet\cite{tan2024frequencyaware} & 48.59 & 7.79 & 92.60 & 4.35 & 48.48 & 50.30 & 8.67 & 6.17 \\ \cline{2-10} 
 & PoundNet & \textbf{97.97} & \textbf{90.41} & 82.30 & \textbf{97.10} & \textbf{89.70} & \textbf{97.83} & \textbf{88.78} & \textbf{92.67} \\ \hline
 
\multirow{11}{*}{\rotatebox{90}{S3GAN}} & CNNDet\cite{wang2020cnn} & 85.41 & 36.91 & 98.60 & 22.95 & 60.77 & 85.87 & 36.97 & 27.71 \\
 & FreDect\cite{frank2020leveraging} & {\ul 97.30} & {\ul 85.16} & 68.25 & \textbf{97.70} & 82.97 & {\ul 97.00} & {\ul 83.50} & {\ul 90.20} \\
 & GramNet\cite{liu2020global} & 51.58 & 2.35 & 99.25 & 1.20 & 50.22 & 50.56 & 2.73 & 1.94 \\
 & Fusing\cite{ju2022fusing} & 87.99 & 35.12 & 99.30 & 21.45 & 60.38 & 87.86 & 35.59 & 26.62 \\
 & LNP\cite{liu2022detecting} & 89.60 & 37.20 & 98.90 & 23.10 & 61.00 & 90.74 & 37.24 & 27.97 \\
 & SPrompts\cite{wang2022s} & 91.92 & 43.72 & \textbf{99.90} & 28.00 & 63.95 & 91.32 & 43.79 & 33.39 \\
 & UnivFD\cite{ojha2023towards} & 97.03 & 84.10 & 98.85 & 73.40 & {\ul 86.12} & 96.53 & 81.65 & 75.39 \\
 & LGrad\cite{tan2023learning} & 56.80 & 0.00 & \textbf{99.90} & 0.00 & 49.95 & 58.02 & 0.47 & 0.51 \\
 & NPR\cite{tan2023rethinking} & 47.53 & 9.05 & 93.45 & 5.05 & 49.25 & 47.01 & 9.23 & 6.46 \\
 & Freqnet\cite{tan2024frequencyaware} & 47.63 & 3.46 & 97.75 & 1.80 & 49.78 & 46.82 & 3.91 & 2.73 \\ \cline{2-10} 
 & PoundNet & \textbf{98.10} & \textbf{92.96} & 89.75 & {\ul 95.75} & \textbf{92.75} & \textbf{98.03} & \textbf{91.25} & \textbf{93.07} \\ \hline

\end{tabular}
\end{table}

\begin{table}[]
\small
\caption{Comprehensive comparisons of our method and other ai-generated images detectors on GANGen-Detection dataset. \zhiwu{$\mathbf{ACC_r}$, $\mathbf{ACC_f}$ represent detection accurracies for real images and fake images, respectively.} For each test subset, the best results are highlighted in boldface and the second best results are underlined.}

\begin{tabular}{c|l|rrrrrrrr}
\hline
& \multicolumn{1}{c|}{\textbf{Method}} & \multicolumn{1}{c}{$\mathbf{AP}$} & \multicolumn{1}{c}{$\mathbf{F1}$} & \multicolumn{1}{c}{$\mathbf{ACC_r}$} & \multicolumn{1}{c}{$\mathbf{ACC_f}$} & \multicolumn{1}{c}{$\mathbf{ACC}$} & \multicolumn{1}{c}{$\mathbf{AUC_{roc}}$} & \multicolumn{1}{c}{$\mathbf{AUC_{f1}}$} & \multicolumn{1}{c}{$\mathbf{AUC_{f2}}$} \\ \hline

\multirow{11}{*}{\rotatebox{90}{SNGAN}} & CNNDet\cite{wang2020cnn} & 87.12 & 37.62 & 99.20 & 23.35 & 61.27 & 86.35 & 37.99 & 28.86 \\
 & FreDect\cite{frank2020leveraging} & 50.44 & 0.59 & 99.40 & 0.30 & 49.85 & 52.42 & 4.30 & 3.37 \\
 & GramNet\cite{liu2020global} & 48.72 & 0.20 & {\ul 99.95} & 0.10 & 50.02 & 48.61 & 0.54 & 0.55 \\
 & Fusing\cite{ju2022fusing} & 90.45 & 50.48 & 98.70 & 34.20 & 66.45 & 90.90 & 49.13 & 39.71 \\
 & LNP\cite{liu2022detecting} & 45.59 & 2.96 & 96.80 & 1.55 & 49.18 & 44.85 & 4.37 & 3.15 \\
 & SPrompts\cite{wang2022s} & 85.55 & 12.37 & 99.90 & 6.60 & 53.25 & 85.60 & 13.73 & 9.44 \\
 & UnivFD\cite{ojha2023towards} & \textbf{94.96} & {\ul 86.78} & 91.20 & 83.40 & \textbf{87.30} & \textbf{94.58} & {\ul 81.08} & 79.47 \\
 & LGrad\cite{tan2023learning} & 50.59 & 0.00 & \textbf{100.00} & 0.00 & 50.00 & 51.19 & 0.34 & 0.42 \\
 & NPR\cite{tan2023rethinking} & 50.05 & 66.69 & 0.10 & \textbf{100.00} & 50.05 & 50.11 & 66.68 & {\ul 83.34} \\
 & Freqnet\cite{tan2024frequencyaware} & 52.41 & 66.45 & 5.95 & {\ul 96.55} & 51.25 & 53.40 & 65.92 & 80.96 \\ \cline{2-10} 
 & PoundNet & {\ul 94.02} & \textbf{87.69} & 78.90 & 94.55 & {\ul 86.72} & {\ul 94.57} & \textbf{84.37} & \textbf{87.68} \\ \hline
 
\multirow{11}{*}{\rotatebox{90}{STGAN}} & CNNDet\cite{wang2020cnn} & 92.11 & 37.04 & 99.90 & 22.75 & 61.32 & 91.72 & 37.66 & 28.44 \\
 & FreDect\cite{frank2020leveraging} & 56.61 & 0.20 & \textbf{100.00} & 0.10 & 50.05 & 58.06 & 2.03 & 1.59 \\
 & GramNet\cite{liu2020global} & 44.63 & 0.20 & 99.70 & 0.10 & 49.90 & 44.37 & 0.53 & 0.54 \\
 & Fusing\cite{ju2022fusing} & \textbf{97.81} & 70.77 & 99.65 & 54.95 & 77.30 & \textbf{97.74} & 68.62 & 59.11 \\
 & LNP\cite{liu2022detecting} & 39.82 & 5.92 & 86.95 & 3.45 & 45.20 & 35.07 & 7.72 & 5.89 \\
 & SPrompts\cite{wang2022s} & 89.43 & 18.01 & {\ul 99.95} & 9.90 & 54.93 & 88.52 & 19.20 & 13.34 \\
 & UnivFD\cite{ojha2023towards} & 93.71 & {\ul 84.94} & 91.50 & 80.10 & \textbf{85.80} & 92.62 & {\ul 80.00} & 77.81 \\
 & LGrad\cite{tan2023learning} & 49.91 & 0.00 & {\ul 99.95} & 0.00 & 49.98 & 51.02 & 0.36 & 0.44 \\
 & NPR\cite{tan2023rethinking} & 51.01 & 66.69 & 0.10 & \textbf{100.00} & 50.05 & 51.96 & 66.69 & {\ul 83.35} \\
 & Freqnet\cite{tan2024frequencyaware} & 50.36 & 66.42 & 9.75 & {\ul 94.60} & 52.18 & 54.06 & 65.98 & 80.12 \\ \cline{2-10} 
 & PoundNet & {\ul 93.82} & \textbf{85.52} & 76.65 & 92.15 & {\ul 84.40} & {\ul 93.44} & \textbf{82.65} & \textbf{86.15} \\ \hline
 
\multirow{11}{*}{\rotatebox{90}{Average}} & CNNDet & 82.69 & 37.27 & 99.19 & 25.11 & 62.15 & 82.49 & 37.57 & 29.74 \\
 & FreDect & 63.59 & 21.33 & 86.37 & 23.25 & 54.81 & 64.61 & 24.74 & 25.52 \\
 & Freqnet & 49.62 & 49.49 & 33.23 & 68.84 & 51.04 & 50.11 & 49.37 & 58.81 \\
 & Fusing & 90.55 & 58.54 & 98.11 & 45.00 & 71.55 & 90.46 & 57.12 & 49.21 \\
 & GramNet & 50.00 & 0.41 & 99.78 & 0.21 & 49.99 & 49.80 & 0.77 & 0.69 \\
 & LGrad & 52.19 & 0.00 & \textbf{99.98} & 0.00 & 49.99 & 53.54 & 0.37 & 0.44 \\
 & LNP & 60.25 & 9.52 & 96.62 & 5.47 & 51.05 & 60.03 & 10.98 & 7.99 \\
 & NPR & 48.07 & 47.65 & 29.41 & 70.53 & 49.97 & 47.09 & 47.73 & 59.10 \\
 & PoundNet & \textbf{93.22} & \textbf{85.79} & 75.31 & \textbf{93.67} & \textbf{84.49} & \textbf{93.39} & \textbf{83.67} & \textbf{87.93} \\
 & SPrompts & 88.45 & 24.02 & {\ul 99.94} & 15.17 & 57.55 & 87.86 & 24.75 & 18.77 \\ \cline{2-10} 
 & UnivFD & {\ul 91.91} & {\ul 79.95} & 92.83 & {\ul 74.53} & {\ul 83.68} & {\ul 91.26} & {\ul 75.87} & {\ul 73.23} \\ \hline

\end{tabular}
\end{table}

\begin{figure}[]
\centering
\centerline{\includegraphics[width=1\linewidth]{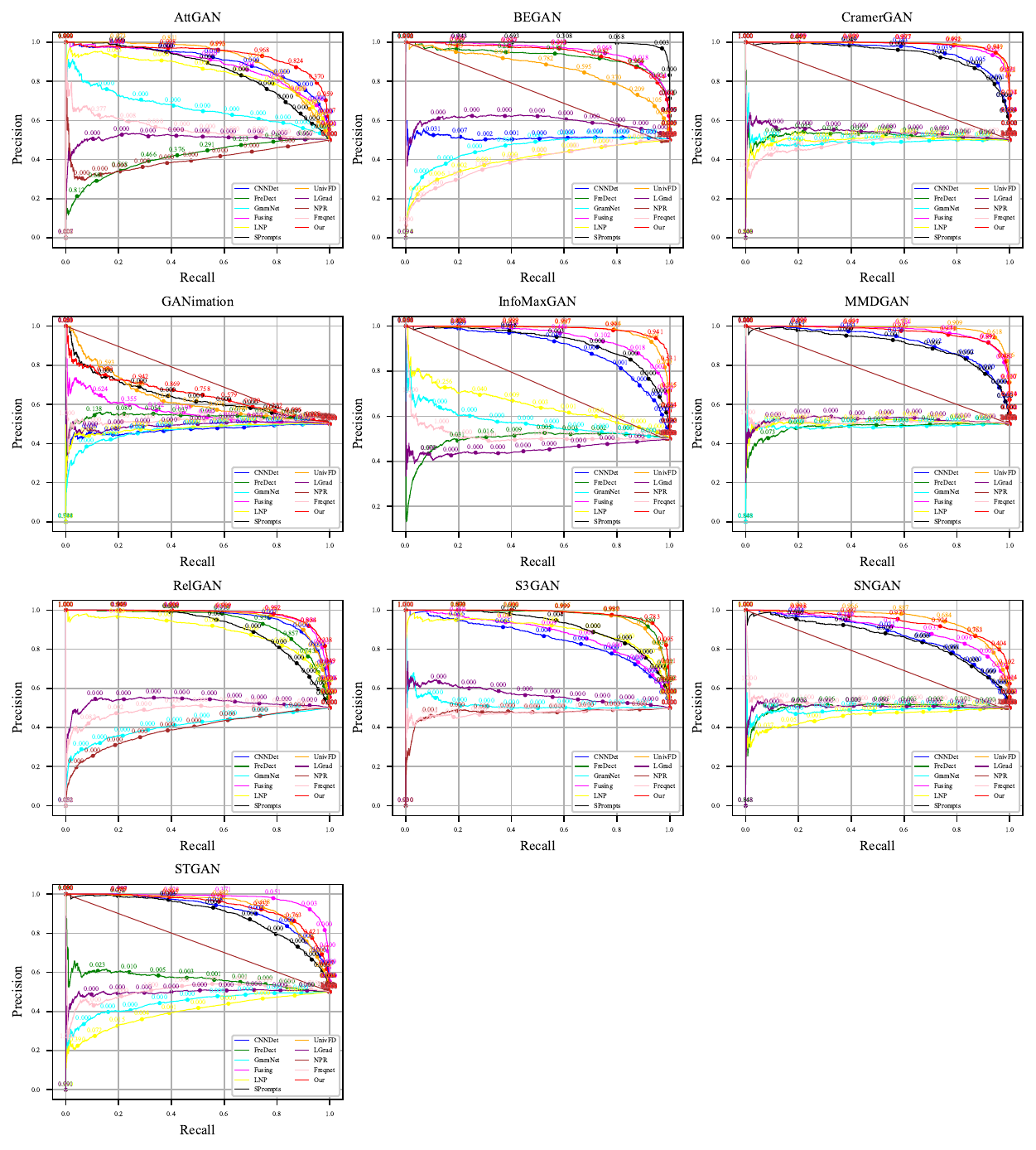}}
\caption{\zhiwu{Precision-Recall Curves of deepfake detection methods on the GANGen dataset. The numbers on each curve represent the decision thresholds that define the boundary between positive and negative predictions. The numbers typically fall within a narrow range.}}
\label{fig:}
\end{figure}

\begin{figure}[]
\centering
\centerline{\includegraphics[width=1\linewidth]{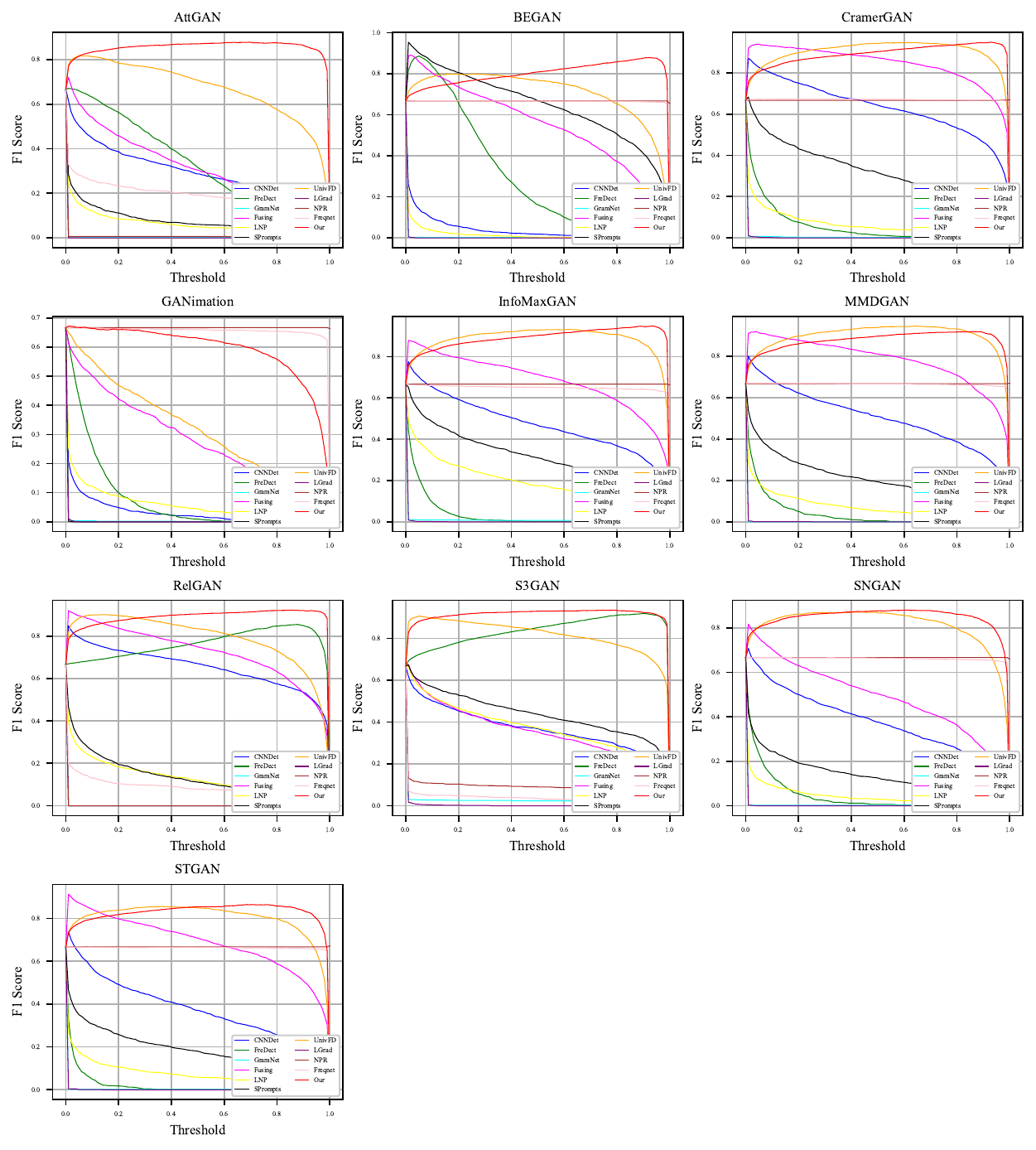}}
\caption{\zhiwu{F1 Curves with different threshold of logits of deepfake detection methods on various deepfakes on the GANGen dataset. The horizontal axis denotes the decision thresholds that determine the boundary between positive and negative predictions.}}
\label{fig:}
\end{figure}

\begin{table}[]
\small
\caption{Comprehensive comparisons of our method and other ai-generated images detectors on GenImage dataset. \zhiwu{$\mathbf{ACC_r}$, $\mathbf{ACC_f}$ represent detection accurracies for real images and fake images, respectively.} For each test subset, the best results are highlighted in boldface and the second best results are underlined.}
\begin{tabular}{c|l|rrrrrrrr}
\hline
& \multicolumn{1}{c|}{\textbf{Method}} & \multicolumn{1}{c}{$\mathbf{AP}$} & \multicolumn{1}{c}{$\mathbf{F1}$} & \multicolumn{1}{c}{$\mathbf{ACC_r}$} & \multicolumn{1}{c}{$\mathbf{ACC_f}$} & \multicolumn{1}{c}{$\mathbf{ACC}$} & \multicolumn{1}{c}{$\mathbf{AUC_{roc}}$} & \multicolumn{1}{c}{$\mathbf{AUC_{f1}}$} & \multicolumn{1}{c}{$\mathbf{AUC_{f2}}$} \\ \hline

\multirow{11}{*}{\rotatebox{90}{WuKong}} & CNNDet\cite{wang2020cnn} & 56.32 & 4.41 & 98.68 & 2.28 & 50.48 & 58.27 & 5.29 & 3.64 \\
 & FreDect\cite{frank2020leveraging} & 65.39 & 20.27 & 97.00 & 11.62 & 54.31 & 65.89 & 21.74 & 16.05 \\
 & GramNet\cite{liu2020global} & {\ul 79.26} & \textbf{59.93} & 90.77 & \textbf{46.73} & \textbf{68.75} & \textbf{82.78} & \textbf{59.62} & \textbf{51.11} \\
 & Fusing\cite{ju2022fusing} & 62.77 & 1.49 & \textbf{99.87} & 0.75 & 50.31 & 66.20 & 2.13 & 1.56 \\
 & LNP\cite{liu2022detecting} & 60.55 & 8.31 & 98.13 & 4.42 & 51.28 & 61.57 & 9.40 & 6.45 \\
 & SPrompts\cite{wang2022s} & \textbf{79.59} & 5.47 & \textbf{99.87} & 2.82 & 51.34 & {\ul 80.96} & 6.92 & 4.70 \\
 & UnivFD\cite{ojha2023towards} & 76.25 & 14.38 & 99.13 & 7.82 & 53.47 & 78.04 & 17.34 & 12.56 \\
 & LGrad\cite{tan2023learning} & 53.69 & 0.40 & 99.40 & 0.20 & 49.80 & 57.10 & 0.98 & 0.83 \\
 & NPR\cite{tan2023rethinking} & 62.42 & {\ul 30.42} & 93.33 & {\ul 19.13} & {\ul 56.23} & 62.80 & {\ul 30.42} & {\ul 22.70} \\
 & Freqnet\cite{tan2024frequencyaware} & 50.08 & 0.95 & 98.82 & 0.48 & 49.65 & 52.11 & 1.43 & 1.11 \\
  \cline{2-10} 
 & PoundNet & 72.91 & 24.54 & 97.38 & 14.35 & 55.87 & 74.31 & 25.91 & 19.20 \\
 \hline
 
\multirow{11}{*}{\rotatebox{90}{BigGAN}} & CNNDet\cite{wang2020cnn} & 88.04 & 33.81 & 99.00 & 20.55 & 59.77 & 89.25 & 34.53 & 25.54 \\
 & FreDect\cite{frank2020leveraging} & 77.10 & 69.06 & 63.95 & {\ul 71.75} & 67.85 & 73.63 & 64.19 & {\ul 65.56} \\
 & GramNet\cite{liu2020global} & 49.14 & 1.28 & 99.35 & 0.65 & 50.00 & 49.23 & 1.63 & 1.24 \\
 & Fusing\cite{ju2022fusing} & 89.37 & 31.67 & 99.55 & 18.90 & 59.23 & 90.01 & 32.36 & 23.83 \\
 & LNP\cite{liu2022detecting} & 52.31 & 3.85 & 98.05 & 2.00 & 50.02 & 53.54 & 4.50 & 3.13 \\
 & SPrompts\cite{wang2022s} & 92.67 & 49.74 & {\ul 99.85} & 33.15 & 66.50 & 91.59 & 48.90 & 38.06 \\
 & UnivFD\cite{ojha2023towards} & {\ul 95.48} & {\ul 74.44} & 98.80 & 60.00 & {\ul 79.40} & {\ul 95.02} & {\ul 73.04} & 64.60 \\
 & LGrad\cite{tan2023learning} & 50.93 & 0.60 & \textbf{99.95} & 0.30 & 50.12 & 51.84 & 0.98 & 0.82 \\
 & NPR\cite{tan2023rethinking} & 52.42 & 10.87 & 96.50 & 5.95 & 51.23 & 50.83 & 11.16 & 7.67 \\
 & Freqnet\cite{tan2024frequencyaware} & 49.14 & 3.73 & 97.35 & 1.95 & 49.65 & 47.66 & 4.41 & 3.07 \\
  \cline{2-10} 
 & PoundNet & \textbf{97.62} & \textbf{91.73} & 89.40 & \textbf{93.70} & \textbf{91.55} & \textbf{97.48} & \textbf{89.77} & \textbf{91.08} \\
 \hline
 
\multirow{11}{*}{\rotatebox{90}{Midjourney}} & CNNDet\cite{wang2020cnn} & 63.72 & 12.06 & 98.75 & 6.50 & 52.62 & 63.00 & 12.92 & 8.83 \\
 & FreDect\cite{frank2020leveraging} & 75.00 & 36.57 & 96.85 & 23.08 & 59.97 & 73.56 & 36.81 & 28.45 \\
 & GramNet\cite{liu2020global} & {\ul 86.81} & \textbf{74.94} & 89.95 & \textbf{65.95} & \textbf{77.95} & {\ul 86.51} & \textbf{74.71} & \textbf{69.14} \\
 & Fusing\cite{ju2022fusing} & 75.68 & 1.72 & \textbf{99.90} & 0.87 & 50.38 & 78.69 & 2.99 & 2.12 \\
 & LNP\cite{liu2022detecting} & 78.41 & 36.64 & 97.83 & 22.92 & 60.38 & 76.23 & 36.58 & 27.46 \\
 & SPrompts\cite{wang2022s} & \textbf{91.26} & 19.47 & {\ul 99.73} & 10.82 & 55.27 & \textbf{91.28} & 21.25 & 15.02 \\
 & UnivFD\cite{ojha2023towards} & 48.81 & 1.51 & 98.97 & 0.77 & 49.87 & 48.43 & 3.15 & 2.31 \\
 & LGrad\cite{tan2023learning} & 76.93 & 13.68 & 99.23 & 7.40 & 53.32 & 77.79 & 14.48 & 9.91 \\
 & NPR\cite{tan2023rethinking} & 75.14 & {\ul 46.25} & 92.75 & {\ul 32.27} & {\ul 62.51} & 76.89 & {\ul 46.09} & {\ul 36.75} \\
 & Freqnet\cite{tan2024frequencyaware} & 73.48 & 11.86 & 98.73 & 6.38 & 52.56 & 76.23 & 12.23 & 8.32 \\
  \cline{2-10} 
 & PoundNet & 63.79 & 14.89 & 97.47 & 8.25 & 52.86 & 64.50 & 16.96 & 12.28 \\
 \hline
 
\multirow{11}{*}{\rotatebox{90}{VQDM}}& CNNDet\cite{wang2020cnn} & 63.81 & 5.38 & 98.68 & 2.80 & 50.74 & 68.52 & 6.35 & 4.35 \\
 & FreDect\cite{frank2020leveraging} & 93.56 & {\ul 76.50} & 97.10 & {\ul 63.73} & {\ul 80.42} & 93.76 & {\ul 69.52} & {\ul 62.90} \\
 & GramNet\cite{liu2020global} & 44.31 & 2.46 & 90.13 & 1.37 & 45.75 & 46.84 & 2.79 & 2.08 \\
 & Fusing\cite{ju2022fusing} & 79.29 & 4.59 & \textbf{99.90} & 2.35 & 51.12 & 80.46 & 5.99 & 4.09 \\
 & LNP\cite{liu2022detecting} & 44.76 & 3.21 & 97.75 & 1.67 & 49.71 & 40.46 & 3.78 & 2.66 \\
 & SPrompts\cite{wang2022s} & 75.05 & 8.54 & {\ul 99.85} & 4.47 & 52.16 & 73.06 & 9.61 & 6.50 \\
 & UnivFD\cite{ojha2023towards} & {\ul 94.28} & 65.83 & 98.92 & 49.60 & 74.26 & {\ul 94.08} & 63.73 & 54.45 \\
 & LGrad\cite{tan2023learning} & 45.52 & 0.36 & 99.42 & 0.18 & 49.80 & 43.32 & 0.80 & 0.71 \\
 & NPR\cite{tan2023rethinking} & 43.27 & 7.76 & 92.30 & 4.35 & 48.33 & 39.94 & 8.16 & 5.75 \\
 & Freqnet\cite{tan2024frequencyaware} & 54.41 & 5.92 & 98.32 & 3.10 & 50.71 & 52.55 & 6.33 & 4.31 \\
  \cline{2-10} 
 & PoundNet & \textbf{95.41} & \textbf{81.00} & 97.33 & \textbf{69.88} & \textbf{83.61} & \textbf{95.15} & \textbf{78.69} & \textbf{72.15} \\
 \hline

\multirow{11}{*}{\rotatebox{90}{ADM}} & CNNDet\cite{wang2020cnn} & 62.54 & 4.66 & 98.65 & 2.42 & 50.53 & 67.06 & 5.24 & 3.61 \\
 & FreDect\cite{frank2020leveraging} & {\ul 89.57} & {\ul 58.30} & 96.67 & {\ul 42.52} & {\ul 69.59} & \textbf{91.78} & {\ul 53.54} & {\ul 46.56} \\
 & GramNet\cite{liu2020global} & 41.47 & 1.39 & 90.53 & 0.77 & 45.65 & 41.36 & 1.72 & 1.35 \\
 & Fusing\cite{ju2022fusing} & 72.44 & 1.52 & \textbf{99.95} & 0.77 & 50.36 & 73.67 & 2.65 & 1.90 \\
 & LNP\cite{liu2022detecting} & 48.18 & 5.45 & 97.68 & 2.87 & 50.28 & 44.45 & 6.13 & 4.23 \\
 & SPrompts\cite{wang2022s} & 59.24 & 0.76 & {\ul 99.78} & 0.38 & 50.08 & 60.61 & 1.40 & 1.09 \\
 & UnivFD\cite{ojha2023towards} & 86.36 & 38.11 & 99.03 & 23.77 & 61.40 & 86.34 & 38.23 & 29.51 \\
 & LGrad\cite{tan2023learning} & 41.22 & 0.40 & 99.47 & 0.20 & 49.83 & 35.73 & 0.80 & 0.71 \\
 & NPR\cite{tan2023rethinking} & 39.38 & 3.49 & 93.07 & 1.90 & 47.48 & 33.97 & 3.87 & 2.78 \\
 & Freqnet\cite{tan2024frequencyaware} & 55.88 & 3.33 & 98.73 & 1.72 & 50.22 & 57.09 & 3.82 & 2.65 \\
 \cline{2-10} 
 & PoundNet & \textbf{90.49} & \textbf{62.92} & 97.25 & \textbf{47.17} & \textbf{72.21} & {\ul 90.56} & \textbf{61.37} & \textbf{52.11} \\
 \hline

\end{tabular}
\end{table}

\begin{table}[]
\small
\caption{Comprehensive comparisons of our method and other ai-generated images detectors on GenImage dataset. \zhiwu{$\mathbf{ACC_r}$, $\mathbf{ACC_f}$ represent detection accurracies for real images and fake images, respectively.} For each test subset, the best results are highlighted in boldface and the second best results are underlined.}
\begin{tabular}{c|l|rrrrrrrr}
\hline
& \multicolumn{1}{c|}{\textbf{Method}} & \multicolumn{1}{c}{$\mathbf{AP}$} & \multicolumn{1}{c}{$\mathbf{F1}$} & \multicolumn{1}{c}{$\mathbf{ACC_r}$} & \multicolumn{1}{c}{$\mathbf{ACC_f}$} & \multicolumn{1}{c}{$\mathbf{ACC}$} & \multicolumn{1}{c}{$\mathbf{AUC_{roc}}$} & \multicolumn{1}{c}{$\mathbf{AUC_{f1}}$} & \multicolumn{1}{c}{$\mathbf{AUC_{f2}}$} \\ \hline

\multirow{11}{*}{\rotatebox{90}{Glide}} & CNNDet\cite{wang2020cnn} & 69.33 & 6.74 & 98.65 & 3.53 & 51.09 & 74.62 & 7.79 & 5.33 \\
 & FreDect\cite{frank2020leveraging} & {\ul 90.12} & {\ul 52.09} & 97.03 & {\ul 36.27} & {\ul 66.65} & {\ul 94.73} & {\ul 49.16} & {\ul 44.40} \\
 & GramNet\cite{liu2020global} & 44.48 & 1.82 & 91.15 & 1.00 & 46.08 & 47.13 & 2.10 & 1.60 \\
 & Fusing\cite{ju2022fusing} & 84.93 & 6.38 & \textbf{99.92} & 3.30 & 51.61 & 85.54 & 8.64 & 5.93 \\
 & LNP\cite{liu2022detecting} & 65.44 & 11.83 & 97.90 & 6.42 & 52.16 & 67.94 & 12.99 & 8.97 \\
 & SPrompts\cite{wang2022s} & 52.98 & 0.46 & {\ul 99.83} & 0.23 & 50.03 & 54.53 & 0.85 & 0.74 \\
 & UnivFD\cite{ojha2023towards} & 87.86 & 37.93 & 99.22 & 23.58 & 61.40 & 88.01 & 38.69 & 30.04 \\
 & LGrad\cite{tan2023learning} & 45.59 & 0.59 & 99.33 & 0.30 & 49.82 & 42.04 & 1.18 & 0.95 \\
 & NPR\cite{tan2023rethinking} & 40.29 & 7.47 & 93.48 & 4.13 & 48.81 & 32.29 & 7.70 & 5.39 \\
 & Freqnet\cite{tan2024frequencyaware} & 51.39 & 0.56 & 98.85 & 0.28 & 49.57 & 53.87 & 0.99 & 0.83 \\
  \cline{2-10} 
 & PoundNet & \textbf{96.06} & \textbf{80.85} & 97.28 & \textbf{69.70} & \textbf{83.49} & \textbf{96.60} & \textbf{77.43} & \textbf{71.01} \\
 \hline
 
\multirow{11}{*}{\rotatebox{90}{SD-v1.4}} & CNNDet\cite{wang2020cnn} & 57.85 & 4.81 & 98.65 & 2.50 & 50.58 & 60.65 & 5.47 & 3.76 \\
 & FreDect\cite{frank2020leveraging} & 69.05 & 23.04 & 96.82 & 13.43 & 55.12 & 70.10 & 24.40 & 18.20 \\
 & GramNet\cite{liu2020global} & {\ul 79.52} & \textbf{61.69} & 89.70 & \textbf{49.20} & \textbf{69.45} & {\ul 81.57} & \textbf{61.52} & \textbf{53.52} \\
 & Fusing\cite{ju2022fusing} & 63.03 & 0.40 & \textbf{99.92} & 0.20 & 50.06 & 66.83 & 1.21 & 0.98 \\
 & LNP\cite{liu2022detecting} & 59.63 & 7.74 & 98.10 & 4.10 & 51.10 & 61.24 & 8.47 & 5.81 \\
 & SPrompts\cite{wang2022s} & \textbf{82.04} & 4.55 & {\ul 99.85} & 2.33 & 51.09 & \textbf{84.08} & 6.06 & 4.14 \\
 & UnivFD\cite{ojha2023towards} & 66.00 & 5.37 & 99.18 & 2.78 & 50.98 & 69.72 & 8.65 & 6.17 \\
 & LGrad\cite{tan2023learning} & 51.14 & 0.36 & 99.42 & 0.18 & 49.80 & 54.02 & 0.82 & 0.73 \\
 & NPR\cite{tan2023rethinking} & 65.21 & {\ul 34.66} & 93.22 & {\ul 22.38} & {\ul 57.80} & 65.48 & {\ul 34.56} & {\ul 26.22} \\
 & Freqnet\cite{tan2024frequencyaware} & 48.19 & 1.02 & 98.83 & 0.52 & 49.68 & 49.00 & 1.42 & 1.11 \\
 \cline{2-10} 
 & PoundNet & 66.30 & 13.32 & 97.43 & 7.32 & 52.38 & 69.32 & 16.17 & 11.71 \\
 \hline

\multirow{11}{*}{\rotatebox{90}{SD-v1.5}}& CNNDet\cite{wang2020cnn} & 58.00 & 4.76 & 98.49 & 2.48 & 50.48 & 60.85 & 5.45 & 3.74 \\
 & FreDect\cite{frank2020leveraging} & 68.15 & 22.36 & 96.84 & 12.99 & 54.91 & 69.37 & 24.00 & 17.93 \\
 & GramNet\cite{liu2020global} & {\ul 79.11} & \textbf{60.93} & 89.90 & \textbf{48.24} & \textbf{69.07} & {\ul 81.21} & \textbf{60.70} & \textbf{52.56} \\
 & Fusing\cite{ju2022fusing} & 63.50 & 0.50 & \textbf{99.91} & 0.25 & 50.08 & 66.92 & 1.22 & 0.98 \\
 & LNP\cite{liu2022detecting} & 58.89 & 8.18 & 97.96 & 4.35 & 51.16 & 60.57 & 8.94 & 6.13 \\
 & SPrompts\cite{wang2022s} & \textbf{81.50} & 4.39 & {\ul 99.80} & 2.25 & 51.02 & \textbf{83.97} & 5.62 & 3.84 \\
 & UnivFD\cite{ojha2023towards} & 65.66 & 5.54 & 99.10 & 2.88 & 50.99 & 69.57 & 8.52 & 6.08 \\
 & LGrad\cite{tan2023learning} & 50.39 & 0.55 & 99.46 & 0.27 & 49.87 & 53.11 & 0.97 & 0.82 \\
 & NPR\cite{tan2023rethinking} & 64.14 & {\ul 33.39} & 92.85 & {\ul 21.48} & {\ul 57.16} & 64.25 & {\ul 33.40} & {\ul 25.26} \\
 & Freqnet\cite{tan2024frequencyaware} & 48.06 & 1.05 & 98.58 & 0.54 & 49.56 & 49.11 & 1.51 & 1.17 \\
  \cline{2-10} 
 & PoundNet & 65.29 & 12.89 & 97.32 & 7.07 & 52.20 & 69.05 & 15.11 & 10.90 \\
 \hline

\multirow{11}{*}{\rotatebox{90}{Average}} & CNNDet & 64.95 & 9.58 & 98.69 & 5.38 & 52.04 & 67.78 & 10.38 & 7.35 \\
 & FreDect & {\ul 78.49} & {\ul 44.77} & 92.78 & {\ul 34.42} & {\ul 63.60} & {\ul 79.10} & {\ul 42.92} & {\ul 37.51} \\
 & Freqnet & 53.83 & 3.55 & 98.53 & 1.87 & 50.20 & 54.70 & 4.02 & 2.82 \\
 & Fusing & 73.88 & 6.03 & \textbf{99.86} & 3.42 & 51.64 & 76.04 & 7.15 & 5.17 \\
 & GramNet & 63.01 & 33.06 & 91.44 & 26.74 & 59.09 & 64.58 & 33.10 & 29.08 \\
 & LGrad & 51.93 & 2.12 & 99.46 & 1.13 & 50.30 & 51.87 & 2.63 & 1.94 \\
 & LNP & 58.52 & 10.65 & 97.93 & 6.09 & 52.01 & 58.25 & 11.35 & 8.11 \\
 & NPR & 55.28 & 21.79 & 93.44 & 13.95 & 53.69 & 53.31 & 21.92 & 16.56 \\
 & PoundNet & \textbf{80.98} & \textbf{47.77} & 96.36 & \textbf{39.68} & \textbf{68.02} & \textbf{82.12} & \textbf{47.68} & \textbf{42.55} \\
 & SPrompts & 76.79 & 11.67 & {\ul 99.82} & 7.06 & 53.44 & 77.51 & 12.58 & 9.26 \\
 & UnivFD & 77.59 & 30.39 & 99.04 & 21.40 & 60.22 & 78.65 & 31.42 & 25.71 \\ \hline

\end{tabular}
\end{table}

\begin{figure}[]
\centering
\centerline{\includegraphics[width=1\linewidth]{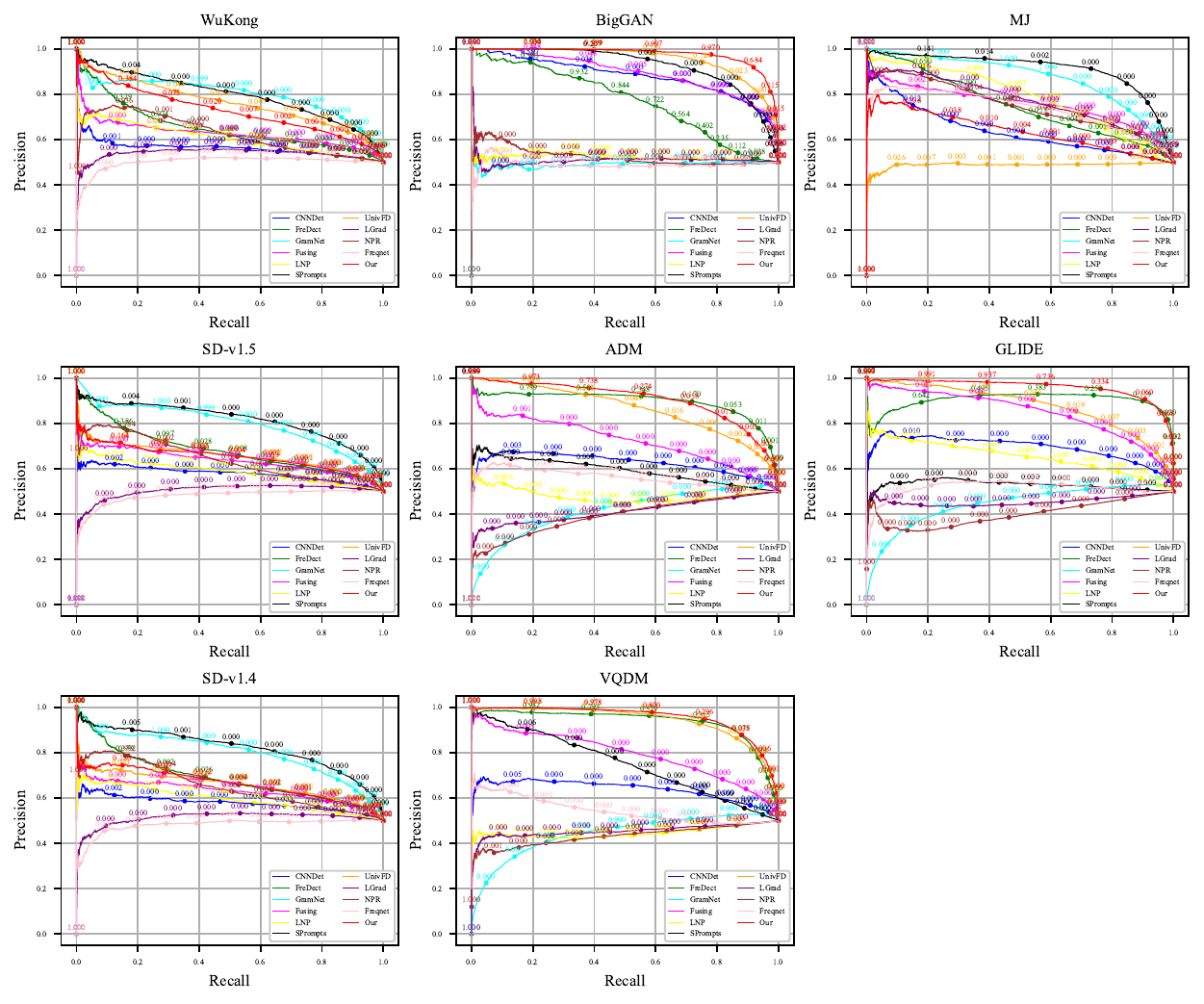}}
\caption{\zhiwu{Precision-Recall Curves of deepfake detection methods on the GenImage dataset. The numbers on each curve represent the decision thresholds that define the boundary between positive and negative predictions. The numbers typically fall within a narrow range.}}
\label{fig:}
\end{figure}

\begin{figure}[]
\centering
\centerline{\includegraphics[width=1\linewidth]{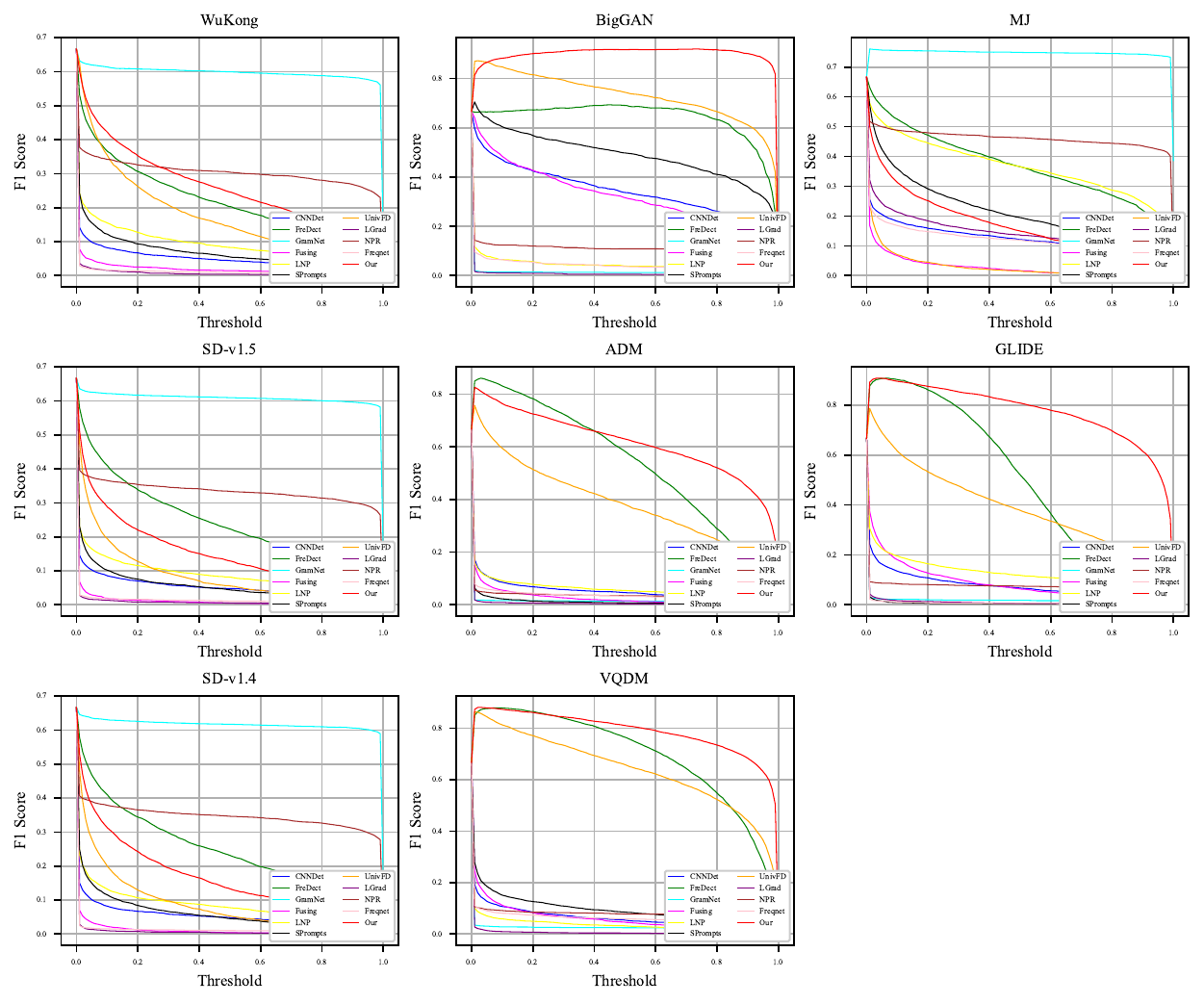}}
\caption{\zhiwu{F1 Curves with different threshold of logits of deepfake detection methods on various deepfakes on the GenImage dataset. The horizontal axis denotes the decision thresholds that determine the boundary between positive and negative predictions.}}
\label{fig:}
\end{figure}

\newpage

\begin{table}[]
\small
\caption{Comprehensive comparisons of our method and other ai-generated images detectors on Ojha dataset.  \zhiwu{$\mathbf{ACC_r}$, $\mathbf{ACC_f}$ represent detection accurracies for real images and fake images, respectively.} For each test subset, the best results are highlighted in boldface and the second best results are underlined.}
\begin{tabular}{c|l|rrrrrrrr}
\hline
& \multicolumn{1}{c|}{\textbf{Method}} & \multicolumn{1}{c}{$\mathbf{AP}$} & \multicolumn{1}{c}{$\mathbf{F1}$} & \multicolumn{1}{c}{$\mathbf{ACC_r}$} & \multicolumn{1}{c}{$\mathbf{ACC_f}$} & \multicolumn{1}{c}{$\mathbf{ACC}$} & \multicolumn{1}{c}{$\mathbf{AUC_{roc}}$} & \multicolumn{1}{c}{$\mathbf{AUC_{f1}}$} & \multicolumn{1}{c}{$\mathbf{AUC_{f2}}$} \\ \hline

\multirow{11}{*}{\rotatebox{90}{DALLE}} & CNNDet\cite{wang2020cnn} & 64.42 & 10.87 & 99.10 & 5.80 & 52.45 & 64.60 & 10.75 & 7.31 \\
 & FreDect\cite{frank2020leveraging} & 76.65 & 67.05 & 73.70 & {\ul 63.70} & 68.70 & 76.17 & 61.69 & {\ul 60.61} \\
 & GramNet\cite{liu2020global} & 47.69 & 1.97 & 99.50 & 1.00 & 50.25 & 44.16 & 2.29 & 1.66 \\
 & Fusing\cite{ju2022fusing} & 73.16 & 14.43 & {\ul 99.70} & 7.80 & 53.75 & 71.49 & 15.73 & 10.82 \\
 & LNP\cite{liu2022detecting} & 41.67 & 5.91 & 94.90 & 3.20 & 49.05 & 35.55 & 6.46 & 4.55 \\
 & SPrompts\cite{wang2022s} & 90.97 & 43.38 & \textbf{100.00} & 27.70 & 63.85 & 88.84 & 42.77 & 32.51 \\
 & UnivFD\cite{ojha2023towards} & {\ul 93.70} & {\ul 67.06} & 99.30 & 50.80 & {\ul 75.05} & {\ul 92.58} & {\ul 64.61} & 55.37 \\
 & LGrad\cite{tan2023learning} & 36.82 & 0.40 & 99.60 & 0.20 & 49.90 & 26.41 & 0.75 & 0.68 \\
 & NPR\cite{tan2023rethinking} & 47.05 & 5.78 & 99.20 & 3.00 & 51.10 & 38.47 & 6.46 & 4.37 \\
 & Freqnet\cite{tan2024frequencyaware} & 49.63 & 1.76 & 98.80 & 0.90 & 49.85 & 48.52 & 2.61 & 1.87 \\
  \cline{2-10} 
 & PoundNet & \textbf{94.56} & \textbf{86.96} & 83.80 & \textbf{89.40} & \textbf{86.60} & \textbf{94.15} & \textbf{84.53} & \textbf{85.71} \\
 
 \hline
 
\multirow{11}{*}{\rotatebox{90}{GLIDE\_100\_10}} & CNNDet\cite{wang2020cnn} & 69.12 & 9.98 & 99.10 & 5.30 & 52.20 & 71.36 & 10.70 & 7.32 \\
 & FreDect\cite{frank2020leveraging} & 55.66 & 40.33 & 73.70 & 31.90 & 52.80 & 66.39 & 39.47 & 39.84 \\
 & GramNet\cite{liu2020global} & 57.08 & 3.71 & 99.50 & 1.90 & 50.70 & 53.04 & 3.78 & 2.61 \\
 & Fusing\cite{ju2022fusing} & 75.67 & 7.12 & {\ul 99.70} & 3.70 & 51.70 & 76.23 & 9.64 & 6.64 \\
 & LNP\cite{liu2022detecting} & 51.97 & 10.63 & 94.90 & 5.90 & 50.40 & 52.25 & 11.87 & 8.42 \\
 & SPrompts\cite{wang2022s} & 60.82 & 1.19 & \textbf{100.00} & 0.60 & 50.30 & 62.25 & 2.15 & 1.57 \\
 & UnivFD\cite{ojha2023towards} & \textbf{95.32} & {\ul 68.45} & 99.30 & {\ul 52.40} & {\ul 75.85} & \textbf{94.87} & {\ul 66.96} & {\ul 58.08} \\
 & LGrad\cite{tan2023learning} & 37.50 & 0.20 & 99.50 & 0.10 & 49.80 & 28.18 & 0.62 & 0.60 \\
 & NPR\cite{tan2023rethinking} & 41.27 & 5.59 & 99.20 & 2.90 & 51.05 & 28.73 & 5.75 & 3.90 \\
 & Freqnet\cite{tan2024frequencyaware} & 43.94 & 0.59 & 98.80 & 0.30 & 49.55 & 41.44 & 0.76 & 0.69 \\
  \cline{2-10} 
 & PoundNet & {\ul 93.67} & \textbf{85.80} & 83.80 & \textbf{87.30} & \textbf{85.55} & {\ul 93.71} & \textbf{83.24} & \textbf{83.95} \\
 \hline
 
\multirow{11}{*}{\rotatebox{90}{GLIDE\_100\_27}}  & CNNDet\cite{wang2020cnn} & 66.61 & 7.63 & 99.10 & 4.00 & 51.55 & 69.87 & 8.17 & 5.56 \\
 & FreDect\cite{frank2020leveraging} & 55.09 & 41.13 & 73.70 & 32.70 & 53.20 & 64.96 & 38.65 & 38.74 \\
 & GramNet\cite{liu2020global} & 58.90 & 3.52 & 99.50 & 1.80 & 50.65 & 54.88 & 4.08 & 2.80 \\
 & Fusing\cite{ju2022fusing} & 72.82 & 7.49 & 99.70 & 3.90 & 51.80 & 73.58 & 8.50 & 5.82 \\
 & LNP\cite{liu2022detecting} & 50.49 & 10.80 & 94.90 & 6.00 & 50.45 & 50.57 & 10.72 & 7.55 \\
 & SPrompts\cite{wang2022s} & 59.02 & 1.19 & \textbf{100.00} & 0.60 & 50.30 & 60.00 & 1.96 & 1.45 \\
 & UnivFD\cite{ojha2023towards} & \textbf{95.52} & {\ul 70.90} & 99.30 & {\ul 55.30} & {\ul 77.30} & \textbf{94.97} & {\ul 67.95} & {\ul 59.24} \\
 & LGrad\cite{tan2023learning} & 37.57 & 0.00 & {\ul 99.80} & 0.00 & 49.90 & 28.02 & 0.49 & 0.52 \\
 & NPR\cite{tan2023rethinking} & 41.67 & 5.59 & 99.20 & 2.90 & 51.05 & 28.57 & 5.85 & 3.97 \\
 & Freqnet\cite{tan2024frequencyaware} & 45.28 & 0.39 & 98.80 & 0.20 & 49.50 & 43.85 & 0.61 & 0.60 \\
  \cline{2-10} 
 & PoundNet & {\ul 93.97} & \textbf{86.08} & 83.80 & \textbf{87.80} & \textbf{85.80} & {\ul 94.12} & \textbf{83.77} & \textbf{84.72} \\
 \hline
 
\multirow{11}{*}{\rotatebox{90}{GLIDE\_50\_27}} & CNNDet\cite{wang2020cnn} & 70.29 & 7.44 & 99.10 & 3.90 & 51.50 & 73.13 & 8.63 & 5.91 \\
 & FreDect\cite{frank2020leveraging} & 55.05 & 41.93 & 73.70 & 33.50 & 53.60 & 65.48 & 38.76 & 39.03 \\
 & GramNet\cite{liu2020global} & 55.05 & 2.17 & 99.50 & 1.10 & 50.30 & 51.69 & 2.38 & 1.72 \\
 & Fusing\cite{ju2022fusing} & 78.06 & 6.74 & {\ul 99.70} & 3.50 & 51.60 & 79.66 & 9.12 & 6.31 \\
 & LNP\cite{liu2022detecting} & 53.01 & 14.15 & 94.90 & 8.00 & 51.45 & 53.14 & 14.10 & 10.02 \\
 & SPrompts\cite{wang2022s} & 63.46 & 1.59 & \textbf{100.00} & 0.80 & 50.40 & 64.61 & 2.81 & 2.00 \\
 & UnivFD\cite{ojha2023towards} & \textbf{95.38} & {\ul 70.40} & 99.30 & {\ul 54.70} & {\ul 77.00} & \textbf{94.90} & {\ul 67.77} & {\ul 58.91} \\
 & LGrad\cite{tan2023learning} & 38.23 & 0.40 & 99.60 & 0.20 & 49.90 & 29.48 & 0.70 & 0.65 \\
 & NPR\cite{tan2023rethinking} & 39.98 & 3.70 & 99.20 & 1.90 & 50.55 & 28.20 & 4.01 & 2.77 \\
 & Freqnet\cite{tan2024frequencyaware} & 41.78 & 0.00 & 98.80 & 0.00 & 49.40 & 37.49 & 0.45 & 0.49 \\
  \cline{2-10} 
 & PoundNet & {\ul 93.58} & \textbf{85.74} & 83.80 & \textbf{87.20} & \textbf{85.50} & {\ul 93.63} & \textbf{83.12} & \textbf{83.79} \\
 \hline
 
\multirow{11}{*}{\rotatebox{90}{Guided}} & CNNDet\cite{wang2020cnn} & 64.91 & 7.96 & 98.70 & 4.20 & 51.45 & 67.82 & 8.73 & 5.96 \\
 & FreDect\cite{frank2020leveraging} & {\ul 91.19} & {\ul 66.92} & 96.80 & {\ul 51.90} & {\ul 74.35} & \textbf{93.88} & {\ul 58.70} & {\ul 52.56} \\
 & GramNet\cite{liu2020global} & 47.34 & 1.31 & 94.00 & 0.70 & 47.35 & 51.73 & 1.79 & 1.38 \\
 & Fusing\cite{ju2022fusing} & 78.04 & 3.91 & \textbf{99.70} & 2.00 & 50.85 & 79.63 & 6.20 & 4.26 \\
 & LNP\cite{liu2022detecting} & 46.99 & 4.60 & 98.00 & 2.40 & 50.20 & 44.73 & 4.92 & 3.40 \\
 & SPrompts\cite{wang2022s} & 66.69 & 1.97 & {\ul 99.50} & 1.00 & 50.25 & 68.03 & 2.98 & 2.11 \\
 & UnivFD\cite{ojha2023towards} & 87.34 & 48.21 & 98.30 & 32.30 & 65.30 & 87.70 & 47.57 & 38.22 \\
 & LGrad\cite{tan2023learning} & 41.80 & 0.00 & 99.20 & 0.00 & 49.60 & 36.88 & 0.38 & 0.45 \\
 & NPR\cite{tan2023rethinking} & 40.30 & 3.09 & 91.80 & 1.70 & 46.75 & 36.64 & 3.40 & 2.48 \\
 & Freqnet\cite{tan2024frequencyaware} & 57.11 & 6.59 & 97.20 & 3.50 & 50.35 & 58.68 & 6.77 & 4.65 \\
  \cline{2-10} 
 & PoundNet & \textbf{93.11} & \textbf{72.83} & 96.80 & \textbf{59.10} & \textbf{77.95} & {\ul 92.94} & \textbf{71.20} & \textbf{63.37} \\ \hline

\end{tabular}
\end{table}

\begin{table}[]
\small
\caption{Comprehensive comparisons of our method and other ai-generated images detectors on Ojha dataset. \zhiwu{$\mathbf{ACC_r}$, $\mathbf{ACC_f}$ represent detection accurracies for real images and fake images, respectively.} For each test subset, the best results are highlighted in boldface and the second best results are underlined.}

\begin{tabular}{c|l|rrrrrrrr}
\hline
& \multicolumn{1}{c|}{\textbf{Method}} & \multicolumn{1}{c}{$\mathbf{AP}$} & \multicolumn{1}{c}{$\mathbf{F1}$} & \multicolumn{1}{c}{$\mathbf{ACC_r}$} & \multicolumn{1}{c}{$\mathbf{ACC_f}$} & \multicolumn{1}{c}{$\mathbf{ACC}$} & \multicolumn{1}{c}{$\mathbf{AUC_{roc}}$} & \multicolumn{1}{c}{$\mathbf{AUC_{f1}}$} & \multicolumn{1}{c}{$\mathbf{AUC_{f2}}$} \\ \hline

\multirow{11}{*}{\rotatebox{90}{LDM\_100}} & CNNDet\cite{wang2020cnn} & 56.95 & 3.70 & 99.10 & 1.90 & 50.50 & 59.85 & 4.47 & 3.09 \\
 & FreDect\cite{frank2020leveraging} & 73.42 & 64.56 & 73.70 & 60.20 & 66.95 & 75.71 & 59.01 & 58.00 \\
 & GramNet\cite{liu2020global} & 53.80 & 1.97 & 99.50 & 1.00 & 50.25 & 51.94 & 2.18 & 1.59 \\
 & Fusing\cite{ju2022fusing} & 68.66 & 5.24 & 99.70 & 2.70 & 51.20 & 68.73 & 7.26 & 4.95 \\
 & LNP\cite{liu2022detecting} & 41.13 & 4.65 & 94.90 & 2.50 & 48.70 & 35.50 & 5.38 & 3.82 \\
 & SPrompts\cite{wang2022s} & 84.40 & 25.48 & \textbf{100.00} & 14.60 & 57.30 & 82.13 & 25.75 & 18.25 \\
 & UnivFD\cite{ojha2023towards} & \textbf{98.44} & {\ul 86.79} & 99.30 & {\ul 77.20} & \textbf{88.25} & \textbf{98.27} & {\ul 83.51} & {\ul 77.89} \\
 & LGrad\cite{tan2023learning} & 36.41 & 0.20 & {\ul 99.90} & 0.10 & 50.00 & 25.43 & 0.70 & 0.65 \\
 & NPR\cite{tan2023rethinking} & 48.21 & 4.65 & 99.20 & 2.40 & 50.80 & 42.73 & 5.03 & 3.43 \\
 & Freqnet\cite{tan2024frequencyaware} & 56.56 & 3.30 & 98.80 & 1.70 & 50.25 & 55.83 & 4.18 & 2.89 \\
  \cline{2-10} 
 & PoundNet & {\ul 95.01} & \textbf{87.40} & 83.80 & \textbf{90.20} & {\ul 87.00} & {\ul 94.81} & \textbf{85.21} & \textbf{86.69} \\ \hline
 
\multirow{11}{*}{\rotatebox{90}{LDM\_200}} & CNNDet\cite{wang2020cnn} & 56.07 & 3.31 & 99.10 & 1.70 & 50.40 & 58.88 & 3.97 & 2.76 \\
 & FreDect\cite{frank2020leveraging} & 71.53 & 63.02 & 73.70 & 58.10 & 65.90 & 74.43 & 56.94 & 55.81 \\
 & GramNet\cite{liu2020global} & 53.00 & 1.78 & 99.50 & 0.90 & 50.20 & 51.78 & 2.11 & 1.54 \\
 & Fusing\cite{ju2022fusing} & 70.20 & 6.56 & 99.70 & 3.40 & 51.55 & 70.58 & 8.37 & 5.71 \\
 & LNP\cite{liu2022detecting} & 42.44 & 4.28 & 94.90 & 2.30 & 48.60 & 38.65 & 5.28 & 3.76 \\
 & SPrompts\cite{wang2022s} & 83.76 & 23.01 & \textbf{100.00} & 13.00 & 56.50 & 81.50 & 23.59 & 16.59 \\
 & UnivFD\cite{ojha2023towards} & \textbf{98.46} & \textbf{86.54} & 99.30 & {\ul 76.80} & \textbf{88.05} & \textbf{98.29} & {\ul 83.44} & {\ul 77.76} \\
 & LGrad\cite{tan2023learning} & 36.41 & 0.00 & {\ul 99.80} & 0.00 & 49.90 & 24.92 & 0.45 & 0.49 \\
 & NPR\cite{tan2023rethinking} & 49.80 & 4.84 & 99.20 & 2.50 & 50.85 & 45.13 & 5.29 & 3.60 \\
 & Freqnet\cite{tan2024frequencyaware} & 58.41 & 5.94 & 98.80 & 3.10 & 50.95 & 56.24 & 6.72 & 4.56 \\
  \cline{2-10} 
 & PoundNet & {\ul 94.18} & {\ul 86.47} & 83.80 & \textbf{88.50} & {\ul 86.15} & {\ul 93.72} & \textbf{84.18} & \textbf{85.20} \\
 \hline
 
\multirow{11}{*}{\rotatebox{90}{LDM\_200\_CFG}} & CNNDet\cite{wang2020cnn} & 58.87 & 3.51 & 99.10 & 1.80 & 50.45 & 60.85 & 4.59 & 3.19 \\
 & FreDect\cite{frank2020leveraging} & 66.50 & {\ul 54.99} & 73.70 & {\ul 47.90} & 60.80 & 68.75 & {\ul 50.74} & {\ul 48.80} \\
 & GramNet\cite{liu2020global} & 50.25 & 2.17 & 99.50 & 1.10 & 50.30 & 46.47 & 2.54 & 1.82 \\
 & Fusing\cite{ju2022fusing} & 72.23 & 10.22 & {\ul 99.70} & 5.40 & 52.55 & 71.43 & 11.41 & 7.81 \\
 & LNP\cite{liu2022detecting} & 43.93 & 7.69 & 94.90 & 4.20 & 49.55 & 38.62 & 8.35 & 5.88 \\
 & SPrompts\cite{wang2022s} & 66.30 & 4.50 & \textbf{100.00} & 2.30 & 51.15 & 63.95 & 6.20 & 4.21 \\
 & UnivFD\cite{ojha2023towards} & \textbf{86.85} & 41.29 & 99.30 & 26.20 & {\ul 62.75} & \textbf{85.66} & 42.48 & 33.40 \\
 & LGrad\cite{tan2023learning} & 39.84 & 0.40 & 99.50 & 0.20 & 49.85 & 33.42 & 0.89 & 0.77 \\
 & NPR\cite{tan2023rethinking} & 48.25 & 6.34 & 99.20 & 3.30 & 51.25 & 41.59 & 6.48 & 4.38 \\
 & Freqnet\cite{tan2024frequencyaware} & 53.86 & 3.11 & 98.80 & 1.60 & 50.20 & 51.48 & 3.63 & 2.53 \\
  \cline{2-10} 
 & PoundNet & {\ul 77.75} & \textbf{63.05} & 83.80 & \textbf{53.50} & \textbf{68.65} & {\ul 77.18} & \textbf{60.44} & \textbf{55.29} \\
 \hline
 
\multirow{11}{*}{\rotatebox{90}{Average}} & CNNDet & 63.41 & 6.80 & 99.05 & 3.58 & 51.31 & 65.79 & 7.50 & 5.14 \\
 & FreDect & 68.14 & 54.99 & 76.59 & 47.49 & 62.04 & 73.22 & 50.49 & 49.17 \\
 & Freqnet & 50.82 & 2.71 & 98.60 & 1.41 & 50.01 & 49.19 & 3.22 & 2.28 \\
 & Fusing & 73.61 & 7.71 & {\ul 99.70} & 4.05 & 51.88 & 73.92 & 9.53 & 6.54 \\
 & GramNet & 52.89 & 2.32 & 98.81 & 1.19 & 50.00 & 50.71 & 2.65 & 1.89 \\
 & LGrad & 38.07 & 0.20 & 99.61 & 0.10 & 49.86 & 29.09 & 0.62 & 0.60 \\
 & LNP & 46.45 & 7.84 & 95.29 & 4.31 & 49.80 & 43.63 & 8.39 & 5.92 \\
 & NPR & 44.57 & 4.95 & 98.28 & 2.58 & 50.42 & 36.26 & 5.28 & 3.61 \\
 & PoundNet & {\ul 91.98} & \textbf{81.79} & 85.42 & \textbf{80.38} & \textbf{82.90} & {\ul 91.78} & \textbf{79.46} & \textbf{78.59} \\
 & SPrompts & 71.93 & 12.79 & \textbf{99.94} & 7.58 & 53.76 & 71.41 & 13.53 & 9.84 \\
 & UnivFD & \textbf{93.87} & {\ul 67.45} & 99.17 & {\ul 53.21} & {\ul 76.19} & \textbf{93.40} & {\ul 65.54} & {\ul 57.36}  \\
 \hline

\end{tabular}
\end{table}

\begin{figure}[]
\centering
\centerline{\includegraphics[width=1\linewidth]{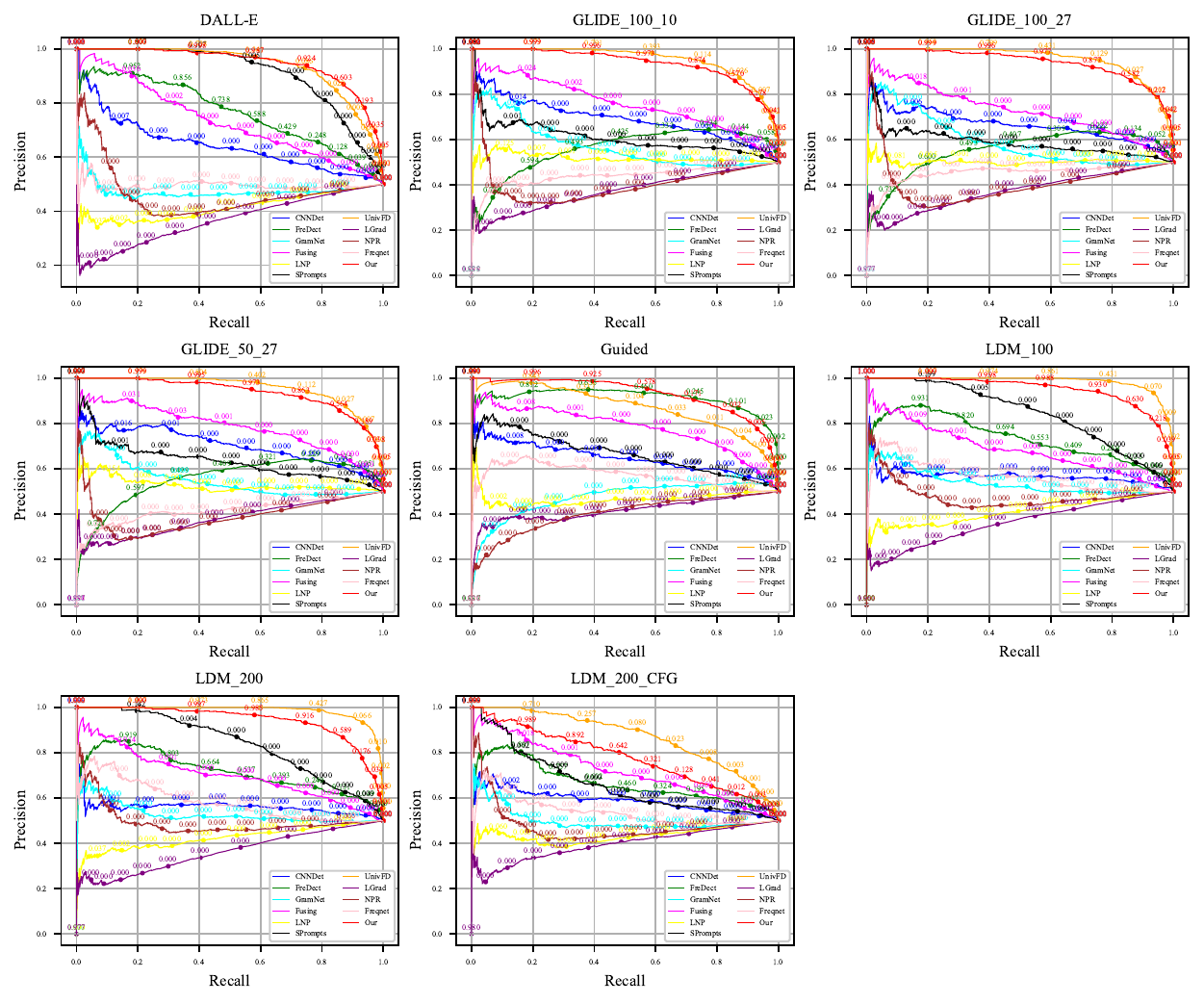}}
\caption{\zhiwu{Precision-Recall Curves of deepfake detection methods on the Ojha dataset. The numbers on each curve represent the decision thresholds that define the boundary between positive and negative predictions. The numbers typically fall within a narrow range.}}
\label{fig:}
\end{figure}

\begin{figure}[]
\centering
\centerline{\includegraphics[width=1\linewidth]{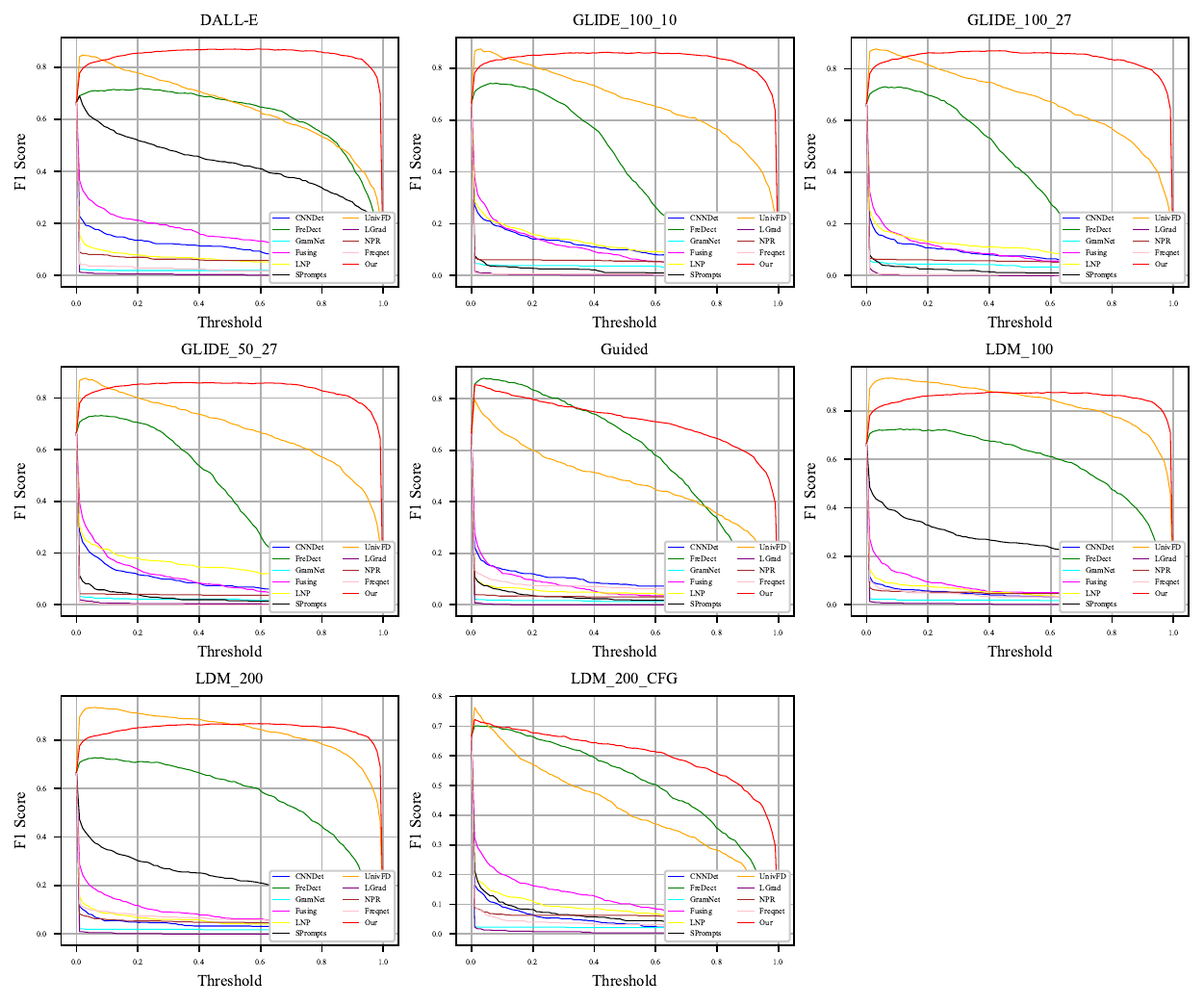}}
\caption{\zhiwu{F1 Curves with different threshold of logits of deepfake detection methods on various deepfakes on the Ojha dataset. The horizontal axis denotes the decision thresholds that determine the boundary between positive and negative predictions.}}
\label{fig:}
\end{figure}

\newpage

\begin{table}[]
\small
\caption{Comprehensive comparisons of our method and other ai-generated images detectors on three classical deepfake detection datasets. \zhiwu{$\mathbf{ACC_r}$, $\mathbf{ACC_f}$ represent detection accurracies for real images and fake images, respectively.} For each test subset, the best results are highlighted in boldface and the second best results are underlined.}
\centering
\begin{tabular}{c|c|rrrrrrrr}
\hline
& \multicolumn{1}{c|}{\textbf{Method}} & \multicolumn{1}{c}{$\mathbf{AP}$} & \multicolumn{1}{c}{$\mathbf{F1}$} & \multicolumn{1}{c}{$\mathbf{ACC_r}$} & \multicolumn{1}{c}{$\mathbf{ACC_f}$} & \multicolumn{1}{c}{$\mathbf{ACC}$} & \multicolumn{1}{c}{$\mathbf{AUC_{roc}}$} & \multicolumn{1}{c}{$\mathbf{AUC_{f1}}$} & \multicolumn{1}{c}{$\mathbf{AUC_{f2}}$} \\ \hline

\multirow{11}{*}{\rotatebox{90}{Celeb-DF-v1}} & CNNDet\cite{wang2020cnn} & 64.40 & 0.14 & 99.84 & 0.07 & 33.94 & 50.08 & 0.72 & 0.66 \\
 & FreDect\cite{frank2020leveraging} & 72.54 & {\ul 53.93} & 69.86 & {\ul 42.64} & {\ul 51.88} & 60.38 & {\ul 46.57} & {\ul 46.04} \\
 & GramNet\cite{liu2020global} & 67.44 & 0.04 & \textbf{99.93} & 0.02 & 33.94 & 52.80 & 0.44 & 0.48 \\
 & Fusing\cite{ju2022fusing} & 61.87 & 0.21 & 99.53 & 0.10 & 33.86 & 46.05 & 1.32 & 1.05 \\
 & LNP\cite{liu2022detecting} & 64.16 & 1.66 & 98.81 & 0.84 & 34.11 & 48.24 & 2.17 & 1.58 \\
 & SPrompts\cite{wang2022s} & 66.38 & 1.75 & 99.39 & 0.88 & 34.33 & 49.87 & 2.53 & 1.81 \\
 & UnivFD\cite{ojha2023towards} & \textbf{83.73} & 16.64 & 99.09 & 9.12 & 39.66 & \textbf{74.19} & 20.54 & 15.24 \\
 & LGrad\cite{tan2023learning} & 65.98 & 0.14 & {\ul 99.85} & 0.07 & 33.95 & 51.64 & 0.60 & 0.58 \\
 & NPR\cite{tan2023rethinking} & 62.13 & 7.28 & 93.90 & 3.90 & 34.45 & 45.86 & 7.59 & 5.18 \\
 & Freqnet\cite{tan2024frequencyaware} & 64.60 & 18.85 & 86.35 & 11.14 & 36.67 & 50.47 & 19.13 & 13.74 \\ \cline{2-10} 
 & PoundNet & {\ul 83.69} & \textbf{71.28} & 69.68 & \textbf{64.01} & \textbf{65.93} & {\ul 73.77} & \textbf{65.69} & \textbf{62.72} \\ \hline
 
\multirow{11}{*}{\rotatebox{90}{Celeb-DF-v2}} & CNNDet\cite{wang2020cnn} & 87.60 & 0.19 & {\ul 99.84} & 0.09 & 13.57 & 54.81 & 0.92 & 0.77 \\
 & FreDect\cite{frank2020leveraging} & 88.10 & {\ul 55.18} & 65.86 & {\ul 40.13} & {\ul 43.61} & 55.77 & {\ul 51.06} & {\ul 47.61} \\
 & GramNet\cite{liu2020global} & 88.43 & 0.08 & \textbf{99.95} & 0.04 & 13.53 & 56.21 & 0.55 & 0.54 \\
 & Fusing\cite{ju2022fusing} & 86.66 & 0.56 & 99.62 & 0.28 & 13.70 & 51.97 & 2.23 & 1.63 \\
 & LNP\cite{liu2022detecting} & 86.76 & 3.30 & 98.45 & 1.68 & 14.75 & 50.88 & 3.96 & 2.71 \\
 & SPrompts\cite{wang2022s} & 87.30 & 1.94 & 99.50 & 0.98 & 14.29 & 50.95 & 2.82 & 1.98 \\
 & UnivFD\cite{ojha2023towards} & {\ul 92.49} & 10.28 & 99.30 & 5.42 & 18.10 & {\ul 67.08} & 14.96 & 10.75 \\
 & LGrad\cite{tan2023learning} & 87.33 & 0.56 & 99.75 & 0.28 & 13.71 & 52.92 & 1.16 & 0.92 \\
 & NPR\cite{tan2023rethinking} & 86.15 & 7.48 & 95.08 & 3.91 & 16.23 & 50.81 & 7.88 & 5.28 \\
 & Freqnet\cite{tan2024frequencyaware} & 86.11 & 17.32 & 89.37 & 9.64 & 20.41 & 51.14 & 17.77 & 12.24 \\ \cline{2-10} 
 & PoundNet & \textbf{93.16} & \textbf{76.95} & 62.44 & \textbf{66.20} & \textbf{65.69} & \textbf{70.19} & \textbf{71.35} & \textbf{65.83} \\ \hline
 
\multirow{11}{*}{\rotatebox{90}{UADFV}} & CNNDet\cite{wang2020cnn} & 61.65 & 0.13 & \textbf{100.00} & 0.07 & 50.42 & 63.42 & 0.65 & 0.62 \\
 & FreDect\cite{frank2020leveraging} & 62.42 & {\ul 52.11} & 75.97 & {\ul 43.83} & 60.03 & 65.15 & {\ul 44.37} & {\ul 44.53} \\
 & GramNet\cite{liu2020global} & 39.31 & 0.00 & 98.77 & 0.00 & 49.77 & 36.05 & 0.33 & 0.42 \\
 & Fusing\cite{ju2022fusing} & 53.18 & 0.65 & {\ul 99.87} & 0.33 & 50.49 & 54.43 & 1.69 & 1.30 \\
 & LNP\cite{liu2022detecting} & 44.87 & 10.18 & 97.22 & 5.51 & 51.73 & 35.74 & 10.34 & 7.09 \\
 & SPrompts\cite{wang2022s} & 46.20 & 11.72 & 88.44 & 6.96 & 48.01 & 48.37 & 12.34 & 9.41 \\
 & UnivFD\cite{ojha2023towards} & \textbf{91.50} & 43.08 & 99.61 & 27.56 & {\ul 63.87} & \textbf{91.27} & 44.23 & 36.37 \\
 & LGrad\cite{tan2023learning} & 44.47 & 0.13 & 98.97 & 0.07 & 49.90 & 43.54 & 0.87 & 0.76 \\
 & NPR\cite{tan2023rethinking} & 43.40 & 10.78 & 94.12 & 6.04 & 50.42 & 35.53 & 11.11 & 7.75 \\
 & Freqnet\cite{tan2024frequencyaware} & 44.55 & 11.44 & 82.36 & 7.15 & 45.05 & 45.22 & 11.89 & 8.95 \\ \cline{2-10} 
 & PoundNet & {\ul 85.92} & \textbf{79.17} & 71.06 & \textbf{84.78} & \textbf{77.86} & {\ul 85.35} & \textbf{71.41} & \textbf{74.32} \\ \hline
\end{tabular}
\end{table}

\end{document}